\documentclass[10pt,twocolumn,letterpaper]{article}

\PassOptionsToPackage{table}{xcolor}
\usepackage[pagenumbers]{wacv} 

\newcommand{\red}[1]{{\color{red}#1}}
\newcommand{\green}[1]{{\color{green}#1}}

\usepackage{eccvabbrv}
\usepackage{multirow}
\usepackage{array}
\usepackage{gensymb}
\usepackage{algorithmic}

\definecolor{wacvblue}{rgb}{0.21,0.49,0.74}
\usepackage[pagebackref,breaklinks,colorlinks,allcolors=wacvblue]{hyperref}

\def\wacvPaperID{3027}
\def\confName{WACV}
\def\confYear{2027}

\title{GroundControl: Satellite-Conditioned Multi-View Diffusion for Reconstruction-Free Novel View Synthesis}

\author{
Akshay Krishnan$^{1,2}$\thanks{Work done as interns at Niantic Spatial} \and
Matias Turkulainen$^{1,3}$\footnotemark[1] \and
Jamie Wynn$^{1}$ \and
Zawar Qureshi$^{1}$ \and
Mohamed Sayed$^{1}$ \and
Guillermo Garcia-Hernando$^{1}$ \and
Sara Vicente$^{1}$ \and
James Hays$^{2}$ \and
Michael Firman$^{1}$ \and
Daniyar Turmukhambetov$^{1}$ \and
{\normalsize
$^{1}$Niantic Spatial \quad
$^{2}$Georgia Institute of Technology \quad
$^{3}$Aalto University \quad
}
}

\begin{document}
\maketitle

\newcommand{\inputsuper}{\text{inp}}
\newcommand{\targetsuper}{\text{tgt}}
\newcommand{\inputimages}{\ensuremath{\mathcal{X}^\inputsuper}\xspace}
\newcommand{\inputcameras}{\ensuremath{\mathcal{C}^\inputsuper}\xspace}
\newcommand{\targetimages}{\ensuremath{\mathcal{X}^\targetsuper}\xspace}
\newcommand{\targetcameras}{\ensuremath{\mathcal{C}^\targetsuper}\xspace}
\newcommand{\satelliteimage}{\ensuremath{S}\xspace}
\newcommand{\ourmodel}{GroundControl\xspace}

\begin{abstract}

Multi-view or video diffusion models have recently emerged as state-of-the-art approaches for novel-view synthesis from sparse image collections. While they can produce high quality images, they tend to hallucinate unrealistic spatial layouts when synthesizing views outside the input views (scene extrapolation). We propose to address this challenge for outdoor scenes by using satellite images as an additional information source to condition the multi-view generative model. We introduce a new satellite-conditioned multi-view diffusion model that combines conditioning from a satellite image and street-level input images when generating novel-views. Our cross-attention based conditioning allows us to retain pretrained multi-view diffusion priors to generate seamless, realistic, novel views from a handful of source views, while also adhering to the scene layout as seen from satellite images. Our model does not use 3D scene reconstruction, and only relies on posed street and satellite views. Our experiments on outdoor datasets show that our model is better than both satellite-conditioned or street-view conditioned baselines on novel-view synthesis, especially on  challenging cases where target views have very little overlap with source views.  Project page: \url{https://nianticspatial.github.io/groundcontrol}

\end{abstract}

\begin{figure}
\includegraphics[width=\linewidth,trim={0 330 82 0},clip]{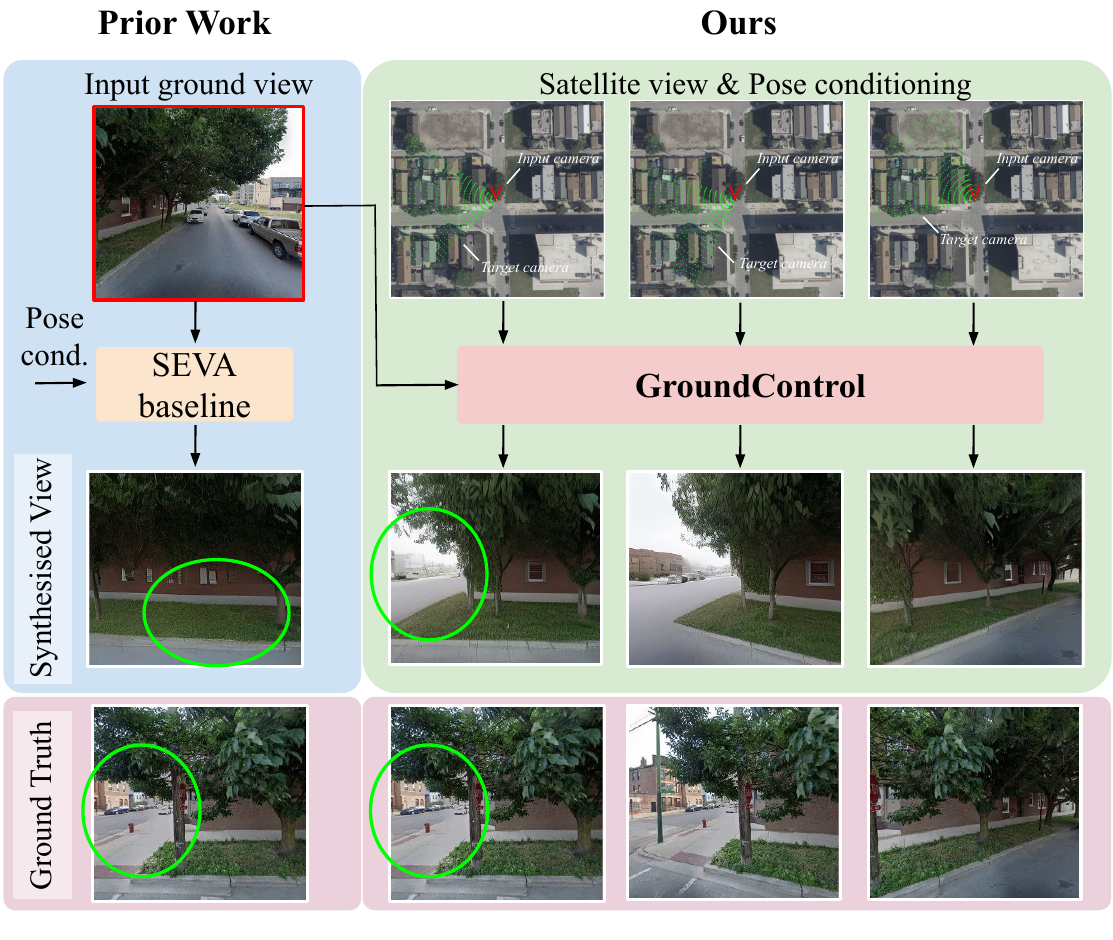}
\vspace{-2em}
\caption{\textbf{GroundControl} builds on recent novel-view synthesis (NVS) diffusion models, and enables more accurate synthesis of novel views by additionally conditioning on a top-down satellite image.
When compared to the NVS diffusion model SEVA \cite{zhou2025stable} (left), our model has better fidelity to the scene's spatial layout.
}
\vspace{-1em}
\label{fig:teaser}
\end{figure}

\vspace{-1em}

\section{Introduction}

Imagine looking at a photo of a street, \eg Fig~\ref{fig:teaser}.
Given just that photo, could you realistically picture what is on either side of the street? 
Current generative models for novel-view synthesis (\eg video models \cite{hoppe2022diffusion,ho2022video,hu2023videocontrolnet,chen2024videocrafter2} or camera-controlled diffusion models \cite{zhou2025stable,guizilini2025zero}) can extrapolate scenes from sparse input views, but they hallucinate outside the input views, generating buildings and roads that may not exist. But what if you also had a satellite image? 
The satellite view provides a good spatial layout prior, although it gives you a limited sense of appearance, \eg the satellite view may not see the facades of buildings and the appearance of smaller objects. 
The street image and the satellite image are both useful, but incomplete. In this work, we fuse the best of both: we introduce a generative model that uses the satellite image as a strong spatial prior to guide ground-level multi-view novel-view synthesis, allowing us to generate new, photorealistic views that adhere to the satellite layout and to the content of street-level photos.

While there is extensive work on generating ground-level images or panoramas from satellite images alone (Table \ref{tab:related_work}),  \cite{xu2024geospecific,zhao2024aerialgo, li2021sat2vid, xu2025groundscape}, doing this without any ground-level information is a highly challenging task. Many such works rely on \emph{oblique} aerial or satellite imagery, where the facade of buildings (and hence street-level textures and geometry) are captured \cite{xu2025groundscape,xu2024geospecific,gao2024skyeyes,Li2024CrossViewDiff}. They struggle with the more common orthographic satellite images which give very limited information about ground-level aesthetics that can only be inferred from street-level source views.  
Furthermore, in applications where sparse ground-level images are available, it is unclear how these methods can be leveraged to complete such partially observed ground-scenes. 
It is therefore appealing to combine satellite images for layout-guidance together with sparse street-level views to generate street-views with accurate textures and appearance.

Text-to-image diffusion models learn a rich image generative prior from large datasets which has been leveraged in many controlled-generation settings, by conditioning on various modalities \cite{mou2024t2i,zhang2023controlnet,petrov2025shapewords,bernal2025precisecam}. However, the combination of multi-view and satellite images poses unique challenges to these conditioning methods due to stark differences in the appearance and viewpoints between them. To this end, we introduce a novel satellite-conditioned multi-view diffusion model for outdoor scenes. By finetuning only our newly added lightweight ray-guided cross-attention layers, we retain the rich multi-view generative priors of the diffusion model. Our approach is reconstruction-free, as in does not assume access to satellite height maps or street-level 3D scene reconstructions or pixel-accurate 6DoF street view-satellite alignment. It only requires relative camera poses and coarse 3DoF (X, Y, heading) satellite alignment. We demonstrate our model's effectiveness across a range of different datasets, from domains including self-driving car data, street-level panoramas, and sparse collections of images from hand held cameras.

\noindent In summary, our key contributions are:

\begin{itemize}[itemsep=0pt, parsep=0pt, topsep=0pt, partopsep=0pt]
    \item We introduce a satellite-conditioned multi-view diffusion model \emph{GroundControl} for novel view synthesis of street scenes. To the best of our knowledge, this is the first satellite-guided multi-view generation approach that does not use height-maps or street-level reconstruction.
    \item We adopt a ray-guided cross-attention method for conditioning state-of-the-art camera controlled diffusion models on 2D satellite image features. Our approach is efficient as it only trains the new cross-attention parameters we add, and on a relatively small dataset. 
    \item Our experiments on diverse datasets show that GroundControl can generate novel views that adhere to the spatial layout of the satellite view, while retaining the multi-view consistency and generative prior of the diffusion model.

\end{itemize}

\section{Related Work}

\begin{table}[ht]
    \centering
    \vspace{-1em}

    \colorlet{goodcol}{green!15}
    \colorlet{badcol}{red!15}
    \newcommand{\goodcell}{\cellcolor{goodcol}}
    \newcommand{\badcell}{\cellcolor{badcol}}

    \setlength{\tabcolsep}{1.5pt}
    \scriptsize
    \begin{tabular}{lcccccc}
            & \shortstack{Street\\cond.} &
            \shortstack{Ortho\\sat.} &
            \shortstack{MV\\cons.} &
            \shortstack{Public\\code} &
            \shortstack{Intermediate\\representation} &
            \shortstack{Output\\modality} \\
            \toprule
\cite{xu2024geospecific}	&	\badcell{N}	&	\badcell{N}	&	\badcell{N}	&	\badcell{N}	&	Mesh	&	Panorama\\
\cite{xu2025groundscape}	&	\badcell{N}	&	\badcell{N}	&	\goodcell{Y}	&	\badcell{N}	&	Mesh	&	Pose-controlled	\\
\cite{gao2024skyeyes}	&	\badcell{N}	&	\badcell{N}	&	\goodcell{Y}	&	\badcell{N}	&	Splat	&	Pose-controlled	\\
\cite{Li2024CrossViewDiff}	&	\badcell{N}	&	\badcell{N}	&	\badcell{N}	&	\badcell{N}	&	Voxels	&	Panorama\\
\cite{shi2022geometry}	&	\badcell{N}	&	\goodcell{Y}	&	\badcell{N}	&	\badcell{N}	&	Warp	&	Panorama\\
\cite{qian2023sat2density}	&	\badcell{N}	&	\goodcell{Y}	&	\badcell{N}	&	\goodcell{Y}	&	Volume	&	Panorama\\
\cite{lu2020geometry, shi2022geometry}	&	\badcell{N}	&	\goodcell{Y}	&	\badcell{N}	&	\goodcell{Y}	&	Heightmap	&	Panorama\\
\cite{li2021sat2vid}	&	\badcell{N}	&	\goodcell{Y}	&	\goodcell{Y}	&	\badcell{N}	&	Voxels	&	Panorama\\
\cite{li2024sat2scene}	&	\badcell{N}	&	\goodcell{Y}	&	\goodcell{Y}	&	\goodcell{Y}	&	Deep point cloud	&	Image	\\
\cite{regmi2018cross,wu2022cross,yang2024retrieval,bajbaa2025satellite}	&	\badcell{N}	&	\goodcell{Y}	&	\badcell{N}	&	\goodcell{Y}	&	Direct (GAN)	&	Panorama\\
\cite{toker2021coming}	&	\badcell{N}	&	\goodcell{Y}	&	\badcell{N}	&	\goodcell{Y}	&	Polar warp	&	Panorama\\
\cite{ze2025controllable}	&	\badcell{N}	&	\goodcell{Y}	&	\badcell{N}	&	\badcell{N}	&	Ray-to-height map	&	Panorama\\
\cite{Qian2025SeeingThroughSatellite,ze2025satdreamer360}	&	\badcell{N}	&	\goodcell{Y}	&	\goodcell{Y}	&	\badcell{N}	&	Triplane	&	Panorama\\
\cite{deng2024streetscapes}	&	\goodcell{Y}	&	\badcell{N*}	&	\goodcell{Y}	&	\badcell{N}	&	Heightmap	&	Pose-controlled	\\
\midrule
\textbf{Ours}	&	\goodcell{Y}	&	\goodcell{Y}	&	\goodcell{Y}	&	\goodcell{Y$\dagger$}	&	Ray-guidance	&	Pose-controlled	\\
\bottomrule
\end{tabular}%
\vspace{-1em}
    \caption{Compared with other related works, \ourmodel enables street-level conditioning and orthorectified satellite views.
    We control views via pose conditioning, and outputs are multi-view consistent.
    Abbreviations: Street cond.~= street-level conditioning; Ortho sat.~= orthorectified satellite; MV cons.~= multi-view consistency.
    * \cite{deng2024streetscapes} accepts map input but not satellite images; $\dagger$ we will release code and models upon acceptance.
    }
    \label{tab:related_work}
\vspace{-1em}

\end{table}

\subsection{Satellite-to-Street Generation}

\noindent \textbf{Street view generation from oblique satellite and aerial views.}
These works typically use \emph{oblique} multi-view satellite imagery to generate a coarse scene geometry, to which visible texture (color) is applied from satellite images.
This coarse reconstruction is then rendered from street-level viewpoints and subsequently refined by a dedicated network.
For example, \cite{xu2025groundscape,xu2024geospecific} use a coarse textured mesh;
\cite{gao2024skyeyes} use a 3D Gaussian Splat \cite{kerbl3Dgaussians};
\cite{Li2024CrossViewDiff} use a voxel-based representation;
and AerialGo \cite{zhao2024aerialgo} use a 3D point representation
similar to \cite{gao2024skyeyes}.
Skyfall-GS \cite{lee2025SkyfallGS} also generates splats from multi-view satellite images, but unlike \cite{gao2024skyeyes} they improve the Gaussian splats themselves.
Unlike these approaches, our method has no restrictions on the types of satellite views it can use as conditioning signal.

\noindent \textbf{Street view generation from orthorectified satellite views.} 
Methods that use orthorectified (``top-down'') satellite inputs are more general, they can generate appearance of invisible building facades (without oblique imagery).
Many of these methods still rely on intermediate representations: Sat2Density \cite{qian2023sat2density} estimates volume density, Sat2Scene \cite{li2024sat2scene} estimates a noisy 3D point cloud and denoises it via diffusion, Sat2Vid \cite{li2021sat2vid} estimates a semantically labeled point cloud, \cite{toker2021coming} uses a polar warp, \cite{Li2024CrossViewDiff} uses street-level voxelized geometry, and other methods estimate depth \cite{lu2020geometry, shi2022geometry}. More recent works Sat2Density++ \cite{Qian2025SeeingThroughSatellite} and SatDreamer360~\cite{ze2025satdreamer360} use ray-guided cross-attention with intermediate triplane representations. These intermediate representations lack data for large-scale supervision, and can create a quality bottleneck as they introduce artifacts downstream. Other works directly map satellite views to ground-level images using GANs \cite{regmi2018cross,wu2022cross,yang2024retrieval,bajbaa2025satellite,goodfellow2020generative,zhu2017unpaired,ren2021cascaded}. StreetScapes \cite{deng2024streetscapes} synthesizes street-level videos from semantic maps and height maps (without true satellite images). Other work generates fully synthetic worlds, sometimes via generated top-down views \cite{lin2023infinicity}.

Our method is the first to directly condition a multi-view generation model on satellite images without any intermediate representations or supervision. It is closely related to \cite{ze2025satdreamer360, ze2025controllable}. SatDreamer360~\cite{ze2025satdreamer360} uses an intermediate triplane representation. It also requires panorama sequences, while ours is trained on multi-view images that are more available. \cite{ze2025controllable} uses a street-to-satellite transformation similar to ours, but their ``attention'' model is actually a height-distribution prediction, not street-to-satellite attention. It also does not allow conditioning on street-view images (only text input). 

\noindent \textbf{Other tasks with satellite and street-level data.}
Other works combine satellite and ground-level views for related tasks. For example, \cite{workman2022revisiting} uses ground-level images to improve satellite segmentation, and Horizon-GS~\cite{jiang2025horizon} combines satellite and street-level inputs for splatting.
Ye \etal \cite{ye2024cross} address the inverse setting by synthesizing satellite views from street images.
Feng \etal \cite{feng2025gps} use GPS as a conditioning signal for image generation, but primarily to guide high-level semantics rather than accurate scene structure.

\subsection{Novel-view Synthesis and Multi-View Diffusion}

Novel view synthesis (NVS) from one or more posed input images has a long history.
Classical methods used image-based rendering with geometry proxies to warp input pixels  \cite{chen2023view,Shih_2020_CVPR,Hedman2018DeepBlending}.
With deep learning, DeepStereo \cite{flynn2016deepstereo} used a network to select pixels from posed input views, while \cite{zhou2016view} learned to warp input-view pixels.
Another line of work directly renders explicit 3D scene representations \cite{kerbl3Dgaussians,aliev2020neural}.

More recently, conditional diffusion models have unlocked direct generation of realistic target views \cite{liu2023zero1to3}.
Works like CAT3D \cite{gao2024cat3d}, EscherNet \cite{kong2024eschernet}, MultiDiff \cite{Muller_2024_CVPR}, PolyOculus \cite{Yu2024PolyOculusNVS} and \cite{watson2022novel,sajjadi2022scene} \emph{implicitly} learn multi-view consistency using ray-based positional encodings \cite{mildenhall2021nerf}, which are now common in novel-view synthesis.
SEVA \cite{zhou2025stable} and MVGD \cite{guizilini2025zero}  further improved temporal consistency by training on fixed-length sequences \cite{zhou2025stable} and performing multi-view diffusion jointly rather than sequentially \cite{guizilini2025zero}. WildCAT3D~\cite{alper2025wildcat3d} further extended this to in-the-wild images.
A key benefit of these approaches is easier scene extrapolation, but they can hallucinate incorrect layouts and semantic artifacts, even when outputs appear plausible.
An alternative line of work generates video from images without camera control \cite{hoppe2022diffusion,ho2022video,hu2023videocontrolnet,chen2024videocrafter2}; we consider camera control to be important for interactive real-world interfaces.
Our work builds on the SEVA \cite{zhou2025stable} backbone approach for novel view synthesis, and our innovations enable the injection of additional priors from satellite views.

\section{Method}
We propose an efficient approach to condition a pretrained camera-controlled multi-view diffusion model on satellite imagery. We extend the widely accepted UNet-based diffusion model architecture for satellite view conditioning by adding ray-guided cross attention parameters between the diffusion embeddings and satellite embeddings from a pretrained satellite encoder. This allows us to leverage both pretrained multi-view diffusion priors as well as pretrained satellite encoders, reducing the need for large numbers of sequence-and-satellite training data. We first provide a brief background on multi-view diffusion models in Section \ref{sec:method-background}, then explain the satellite conditional generation approach in Section \ref{sec:method-sat-cond-prediction} and the ray-guided cross-attention in Section \ref{sec:method-ray-guided-ca}. Our training details are discussed in Section \ref{sec:method-training}. 

\newcommand{\imwidth}{0.14\textwidth}
\begin{figure*}[t]
    \centering
    \vspace{-0.5em}
    \includegraphics[width=1.0\textwidth,trim={0 1320 10 20},clip]{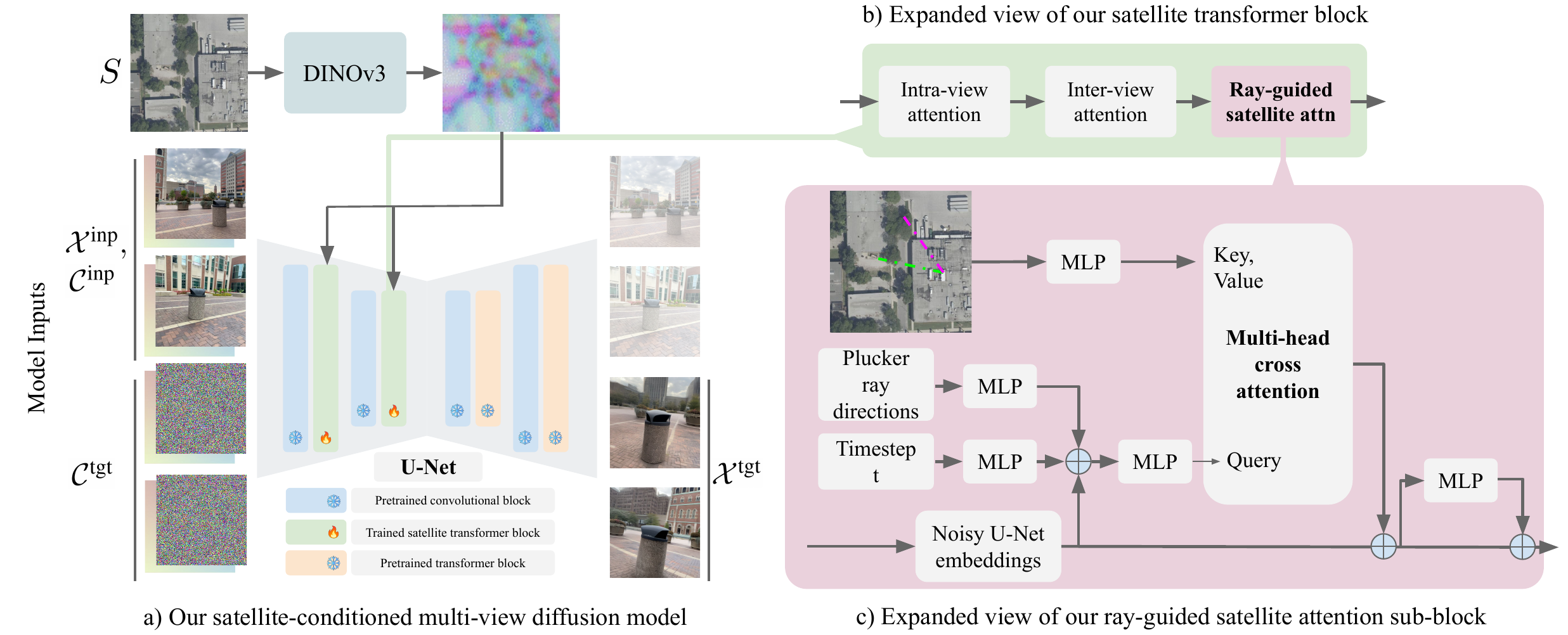}
    \caption{
        \textbf{\ourmodel} incorporates satellite images as an efficient conditioning signal for pose-controllable multi-view diffusion models.
        (a)~Our inputs include a DINOv3 encoded satellite image, known ground-view image(s) \inputimages with associated camera parameters \inputcameras, and target camera parameters \targetcameras corresponding to the views to be synthesized.
        \ourmodel then generates output images \targetimages for each target camera.
        (b)~A key contribution is the integration of satellite information into the SEVA architecture \cite{zhou2025stable}.
        (c)~Our ray-guided attention sub-block fuses satellite features with noisy latent embeddings via multi-head cross-attention.
    }
    \label{fig:method}
    \vspace{-1em}
\end{figure*}

\subsection{Multi-view Diffusion Models}
\label{sec:method-background}

Text-conditioned image diffusion models generate an image $\mathbf{x}$ from text conditions $\mathcal{T}$ by iteratively denoising an initial noise sample $\epsilon$ in $T$ timesteps. They are probabilistic models that sample from the distribution of images conditioned on text $\mathcal{P}(\mathbf{x} | \mathcal{T})$. 
Camera-controlled multi-view diffusion models \cite{zhou2025stable,guizilini2025zero} extend this generative prior to $N$ views, with added conditioning on camera poses and intrinsic parameters. 
Given a set of input images $\inputimages$ and their camera parameters  $\inputcameras$, they generate a set of target views $\targetimages$ for the queried target cameras $\targetcameras$. Probabilistically, they sample from the distribution  $\mathcal{P}(\targetimages | \inputimages, \inputcameras, \targetcameras)$. 

Diffusion models \cite{ho2020denoising} are trained to denoise time-dependent noisy data samples $x_t$. In a variance-exploding diffusion process, the noisy samples $x_t$ are obtained from clean samples $x_0$ by $x_t = x_0 + \sigma_t \epsilon$ where timestep $t$ ranges from [0, T], $\epsilon \sim \mathcal{N} (0, I)$ is noise sampled from a unit Gaussian distribution, and $\sigma_t$ depends on the choice of noise schedule. The model can be trained with different prediction targets; our work uses $\epsilon$-prediction \cite{ho2020denoising}. 

Following standard practice, we use DDIM sampling \cite{song2020denoising} with classifier-free-guidance (CFG) during inference \cite{ho2022classifier}. 
Starting from a pure Gaussian noise sample $x_T \sim \mathcal{N} (0, \sigma_T^2 + 1)$ at $t = T$, the latent diffusion model (LDM) predicts an $\epsilon_t$ at each timestep $t$, from which we compute the predicted $\tilde{x}_0$ and $\tilde{x}_{t-1} = \tilde{x}_0 + \sigma_{t-1} \epsilon_t $ using Tweedie's formula. CFG allows us to condition the $\epsilon_t$ prediction on guidance signals (\eg text), so each $\epsilon_t = \epsilon_{t(u)} + \lambda_\text{cfg} (\epsilon_{t(c)} -  \epsilon_{t(u)})$ where $\epsilon_{t(c)}$ and $\epsilon_{t(u)}$ are conditional and unconditional predictions respectively. A higher $\lambda_\text{cfg}$ value biases the generation closer to the condition. 

In the case of LDMs \cite{rombach2022high}, clean samples $\mathbf{x_0}$ are lower-resolution latents encoded from the image using a pretrained VAE's encoder \cite{kingma2013auto}. For multi-view LDMs, a clean sample comprises multiple target images \targetimages. 
The model is conditioned on additional input views \inputimages, along with camera parameters of both input and target views $\{\inputcameras, \targetcameras\}$. 

\subsection{\ourmodel{}: Satellite-conditioned Multi-view Diffusion}
\label{sec:method-sat-cond-prediction}

We introduce a satellite-conditioned multi-view diffusion model \ourmodel{}, that samples from the conditional distribution $\mathcal{P}(\targetimages \mid \inputimages, \inputcameras, \targetcameras, \satelliteimage)$.
Here, \satelliteimage is the satellite image captured at roughly the same location as $\{\inputcameras, \targetcameras\}$. Effectively Conditioning on satellite images \satelliteimage is non-trivial.
The design choices in \ourmodel{} follow two main principles. First, we retain the multi-view generative prior $\mathcal{P}(\targetimages \mid \inputimages, \inputcameras, \targetcameras)$ of the pretrained model while increasing the likelihood of samples that adhere to the satellite image condition \satelliteimage. Second, we use specialized parameters for satellite images because their semantics differ substantially from those of street-level images.

Hence, \ourmodel{} is finetuned from a multi-view diffusion model \cite{zhou2025stable}, which provides the rich generative multi-view prior from large-scale pretraining on posed images.
Our key contribution is to add residual cross-attention layers to incorporate the satellite conditioning signal into the diffusion model. 
In particular, the residual cross-attention layers contain a zero-initialized linear output layer which ensures that the multi-view diffusion prior is retained at the start of training. Motivated by our second design principle, we encode satellite images with DINOv3~\cite{simeoni2025dinov3}. Our street-level multi-view input images are encoded by the diffusion model itself, alongside the noisy target images.
This allows us to condition the multi-view diffusion model on generalizable satellite embeddings, much like the CLIP text embeddings used in text-to-image diffusion models \cite{ramesh2022hierarchical}. 

Conditioning on a satellite image assumes that the multi-view images are geolocalized. 
In particular, we assume known metric relative poses between views and approximate latitude, longitude and planar orientation of one of the views. 
This allows us to localize each camera with a 2D translation and 1D orientation in the satellite image. 

\subsection{Ray-guided Cross Attention}
\label{sec:method-ray-guided-ca}

An effective way to condition a latent diffusion model is to add additional embeddings (or residuals) from the conditioning signal to specific blocks of the model, directly in the perspective of the generated image. 
When conditioning embeddings are not image-level signals (\eg text, object poses \cite{neuralassets_2024}, etc.), they are incorporated using cross-attention layers where the image embeddings cross-attend to the conditioning embeddings. 
Since satellite images are very different from the street-level image perspective, a viable conditioning method would be to introduce layers for the image tokens to cross-attend to the full satellite image. For a multi-view diffusion model, each image could cross-attend to a transformed satellite image which is centered at and oriented along the corresponding camera's axis. 

\begin{figure}
    \centering
    \vspace{-1em}
    \includegraphics[width=.95\linewidth]{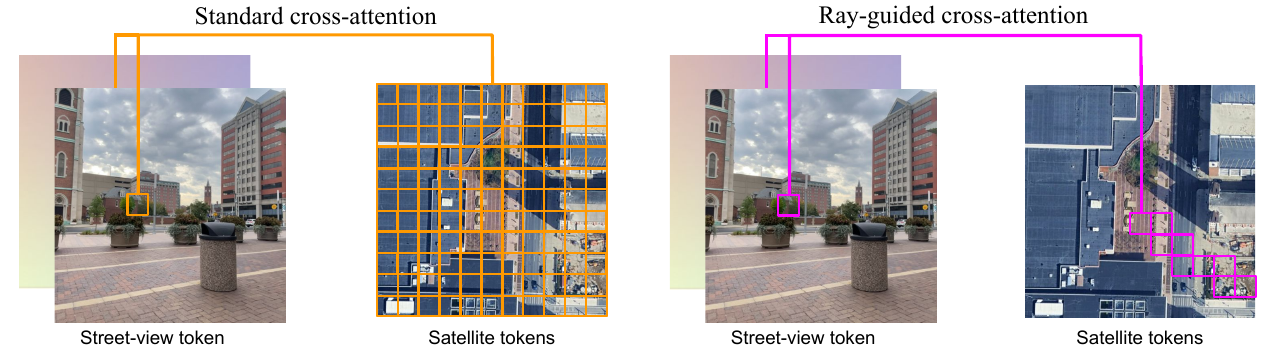}
    \vspace{-0.5em}
    \caption{Our \textbf{ray-guided cross-attention} attends to tokens of the satellite image that are along the 3D ray of the street-view token.}
    \vspace{-1em}
    \label{fig:cross-attention}
\end{figure}

We find that the multi-view diffusion model struggles to learn the conditioning signal from the full satellite image, likely as it is challenging for noisy image tokens to attend to within-view components of the satellite image. 
We address this issue by introducing ray-guided cross-attention layers, inspired by similar works \cite{ze2025controllable,ze2025satdreamer360}. 
Our ray-guided cross-attention (RGCA) layers restricts the cross-attention to be along the 1D ray-direction in the satellite view, allowing it to learn a conditioning signal more efficiently (Figure~\ref{fig:cross-attention}). 
The queries in RGCA are the intermediate UNet features, along with an added embedding computed from Plucker Ray coordinates and the diffusion timestep. We find that the added Plucker and timestep embeddings are important, since the UNet features are noisy in the early diffusion stages. The architecture is shown in Figure \ref{fig:method} (c). 
The final MLP layer that produces the RGCA embeddings is zero-initialized so as to retain the pretrained LDM's latent prior. 
We give full details of RGCA in the supplement.

\subsection{Training}
\label{sec:method-training}

\paragraph{Loss.}
Our model is trained with an L2-loss on the predicted $\epsilon$ noise. We apply this loss only to target views.
\vspace{-1.5em}

\paragraph{Datasets.} We use multiple training datasets that contain geotagged multi-view / panoramic images. These include self-driving datasets with GPS measurements, and geotagged street-view panoramas. In particular, we use nuScenes~\cite{nuscenes}, Argoverse-v2~\cite{Argoverse2}, Metropolis~\cite{metropolis}, and  VIGOR~\cite{zhu2021vigor}. For each training sample, we sample crops from different views of multi-camera rigs in \cite{nuscenes, Argoverse2, metropolis} and from the same panorama in VIGOR. Additionally, we identify pairs of panoramas that are within 20m in VIGOR, and sample views from both panoramas, introducing a camera baseline to reduce scene scale ambiguity. To retain the ability of the multi-view diffusion model to be conditioned on non-driving camera trajectories, we additionally train on training set of DL3DV-10K~\cite{ling2024dl3dv} dataset (excluding Benchmark scenes) without any satellite conditioning. 
\vspace{-1.5em}

\paragraph{Satellite images.} For our training datasets, we compile satellite imagery from various sources (details are provided in the supplementary material). For each multi-view image sample, we crop the satellite images with the center as the center of all cameras, and oriented along north. We used crops of 128m$\times$128m in size, with 4 pixels per meter.  
\vspace{-1.5em}

\paragraph{Implementation details.}
We add our satellite-conditioning blocks to the SEVA model and train only the satellite-conditioning blocks, keeping the SEVA multi-view diffusion weights frozen. We train on multi-view examples with 21 views, with number of input views sampled randomly  from 1 to 8. We train with the Adam optimizer on $16\times$A100 GPUs. We use a learning rate of 3e-5 and a batch size of 192. For our ray-guided cross-attention, we sample 20 points along each ray starting from a minimum distance of 0.5m up to a maximum distance of 62m in the satellite image. We use 1000 diffusion timesteps during training and 50 DDIM steps \cite{song2020denoising} at inference.

\subsection{Inference}

During inference, the model is conditioned on both input views and satellite images. 

\noindent \textbf{Dual classifier-free guidance.}
Our model allows the strength of conditioning on the input image and the satellite view to be controlled independently by applying separate classifier-free guidance to each of them:
\vspace{-5mm}

\begin{equation}
    \hat{\epsilon}_t(i,s) = \epsilon_t(u) + \lambda_i (\epsilon_{t(i)} -  \epsilon_{t(u)}) + \lambda_s (\epsilon_{t(i,s)} -  \epsilon_{t(i)}).
    \label{eq:dual-cfg}
\end{equation}

Here $\epsilon_{t(i,s)}$ is the model's output conditioned on both input and satellite views; $\epsilon_{t(i)}$ is conditioned only on input images; and $\epsilon_t(u)$ is the model's unconditional output.
To enable dual CFG, we drop the input view conditioning at training time with probability 0.3, and drop both the satellite and input view conditioning with probability 0.15.

\begin{table*}
\vspace{-1em}
\newcommand{\rowindent}{\hspace{6pt}}
\setlength{\tabcolsep}{3pt}
\centering
\resizebox{\textwidth}{!}{%
\begin{tabular}{l ccccccc ccccccc}
    & \multicolumn{7}{c}{\textbf{nuScenes} \cite{nuscenes}}
    & \multicolumn{7}{c}{\textbf{Metropolis} \cite{metropolis}} \\
    \cmidrule(lr){2-8} \cmidrule(lr){9-15}
    & PSNR $\uparrow$
    & SSIM $\uparrow$
    & LPIPS $\downarrow$
    & DINO $\downarrow$
    & SEG $\uparrow$
    & IoU$_{\mathrm{road}}$ $\uparrow$
    & IoU$_{\mathrm{bdg}}$ $\uparrow$
    & PSNR $\uparrow$
    & SSIM $\uparrow$
    & LPIPS $\downarrow$
    & DINO $\downarrow$
    & SEG $\uparrow$
    & IoU$_{\mathrm{road}}$ $\uparrow$
    & IoU$_{\mathrm{bdg}}$ $\uparrow$ \\
    \midrule

    \textbf{Street-View only} \\
    \rowindent SEVA
    & 11.334 & 0.308 & \textbf{0.616} & \textbf{6.140} & 0.492
    & 0.425 & 0.182
    & 10.884 & 0.315 & 0.586 & 5.654 & 0.551
    & 0.581 & 0.291 \\

    \midrule
    \textbf{Satellite only} \\
    \rowindent Sat2Density \cite{qian2023sat2density}
    & 10.679 & 0.301 & 0.714 & 6.657 & 0.507
    & -- & --
    & 9.074 & 0.346 & 0.743 & 6.775 & 0.537
    & -- & -- \\

    \midrule
    \textbf{Street-view + Satellite} \\
    \rowindent Ours
    & \textbf{11.880} & \textbf{0.331} & 0.619 & 6.145 & \textbf{0.498}
    & \textbf{0.437} & \textbf{0.200}
    & \textbf{12.809} & \textbf{0.401} & \textbf{0.547}
    & \textbf{5.247} & \textbf{0.605}
    & \textbf{0.630} & \textbf{0.372} \\

    \bottomrule
\end{tabular}
}
\vspace{-1em}
\caption{Comparison with baseline methods on test datasets that
\textbf{match our training domain}, \ie driving datasets. Our model improves
over the baseline systems. Semantic IoU includes vehicles in both the road
and building masks.}
\label{tab:satellite_baselines}
\end{table*}

\begin{table}
\centering
\renewcommand*{\arraystretch}{1}
\newcommand{\rowindent}{\hspace{6pt}}
\setlength{\tabcolsep}{2pt}
    \resizebox{1.0\columnwidth}{!}{
    \begin{tabular}{l ccccc | ccc}
     & \multicolumn{7}{c}{\textbf{DL3DV} \cite{ling2024dl3dv}} \\
    \cmidrule(lr){2-9} 
    & PSNR $\uparrow$	&	SSIM $\uparrow$	&	LPIPS $\downarrow$	&	DINO $\downarrow$	&	SEG $\uparrow$ & TSED $\downarrow$ & \%Consistent $\uparrow$ & \#Corrs. $\uparrow$ \\
    \midrule
    Sat2Density~\cite{qian2023sat2density} & 10.204 & 0.232 & 0.768 & 7.116 & 0.451 & N/A & 0.0 & 0 \\
    ViewCrafter*~\cite{yu2024viewcrafter} &  13.350 &  0.260 & 0.547 & 5.230 &  0.541 & 3845.57 & 2.02 & \textbf{332.62} \\
    SEVA~\cite{zhou2025stable} &	14.912	&	0.360	&	\textbf{0.407}	&	\textbf{3.534}	&	\textbf{0.674}	& 22.75 & 62.99 & 287 \\
    Ours & \textbf{15.170}  &   \textbf{0.372}   &   0.414 & 3.653 & \textbf{0.674}  & \textbf{18.35} & \textbf{65.56} & \textbf{290}\\
    \midrule
    \end{tabular}
}
\vspace{-1em}
\caption{Comparison with baseline methods on DL3DV, a dataset which is \textbf{out of distribution with respect to our  training data}, as it has large-viewpoint NVS with sparse inputs.
We use the geolocalized dataset split from \cite{turkulainen2026crossviewsplatter}. We use a total of 39 sequences with $3$ input views and $37$ test views per sequence. *ViewCrafter is incompatible with our input-target view split since it only generates a continuous trajectory, and therefore uses a different continuous input-target configuration.}
\label{tab:nvs_baselines}
\end{table}

\renewcommand{\imwidth}{0.135\textwidth}
\begin{figure*}
    \centering
    \scalebox{0.90}{
    \setlength{\tabcolsep}{2pt}
    \begin{tabular}{cccccccc}
\raisebox{8pt}{\rotatebox{90}{Input Satellite}} &
\includegraphics[width=\imwidth]{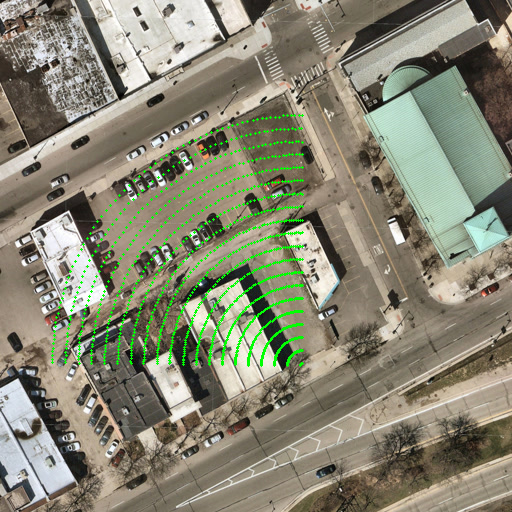} &
\includegraphics[width=\imwidth]{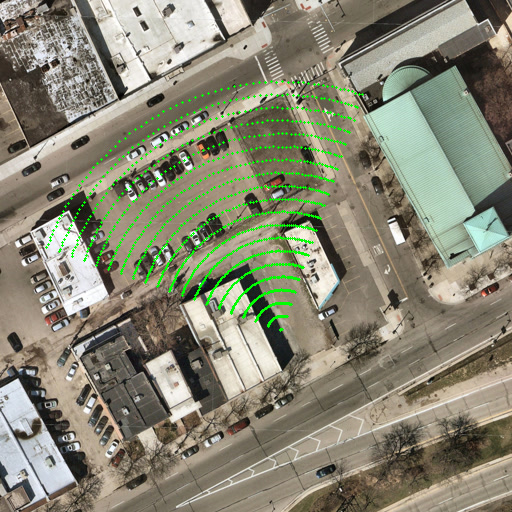} &
\includegraphics[width=\imwidth]{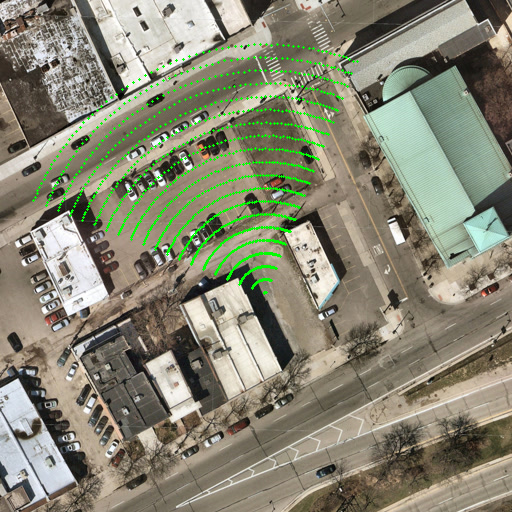} &
\includegraphics[width=\imwidth]{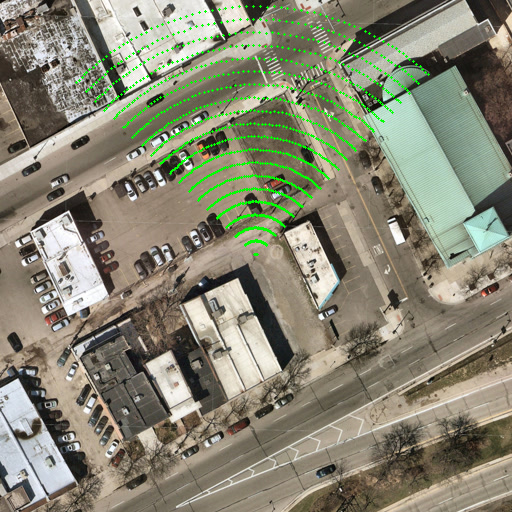} &
\includegraphics[width=\imwidth]{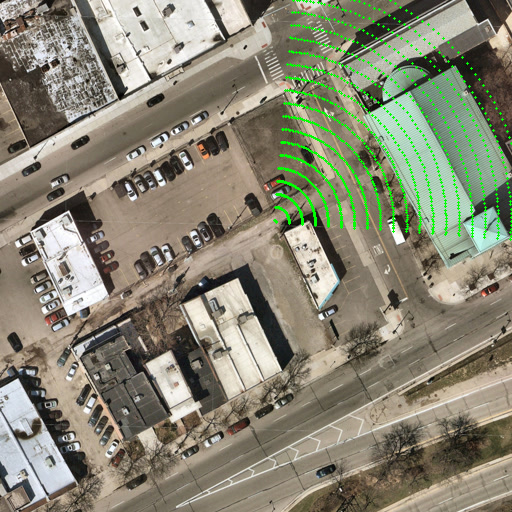} &
\includegraphics[width=\imwidth]{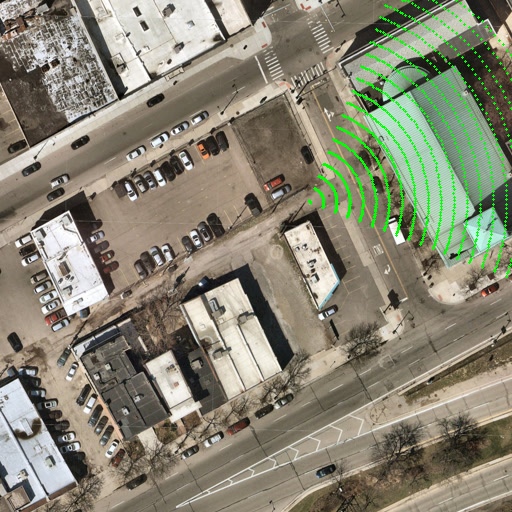} &
\includegraphics[width=\imwidth]{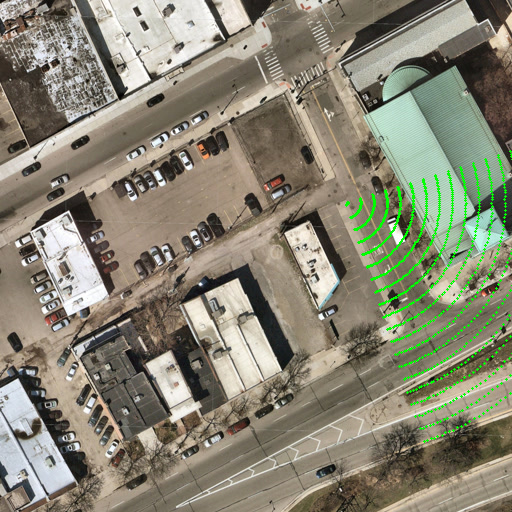}
 \\
\raisebox{14pt}{\rotatebox{90}{Input views}} &
\includegraphics[width=\imwidth]{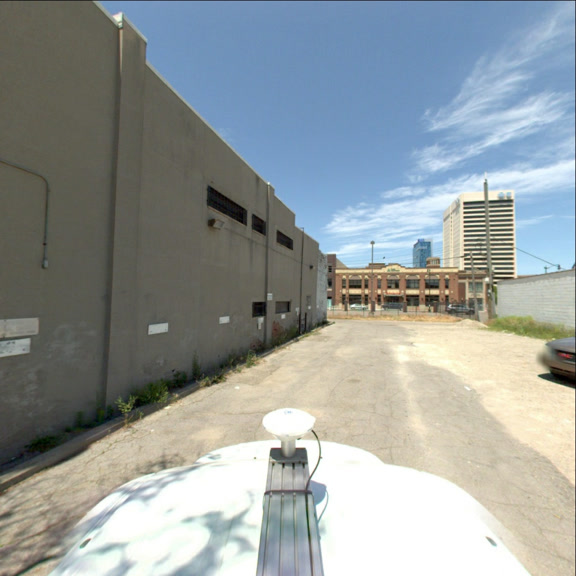} &
\includegraphics[width=\imwidth]{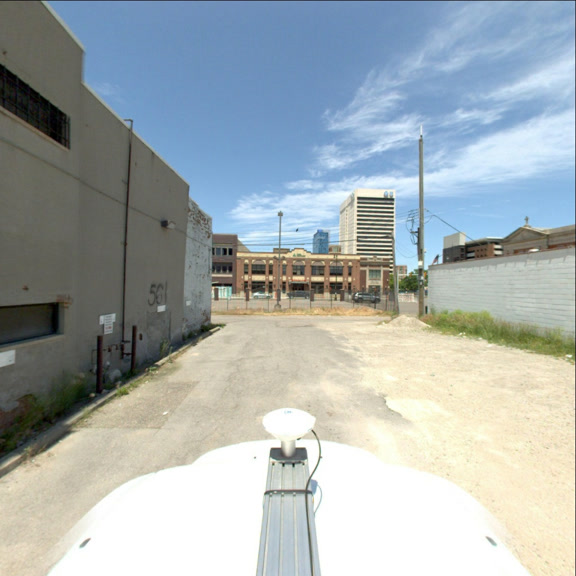} &
\includegraphics[width=\imwidth]{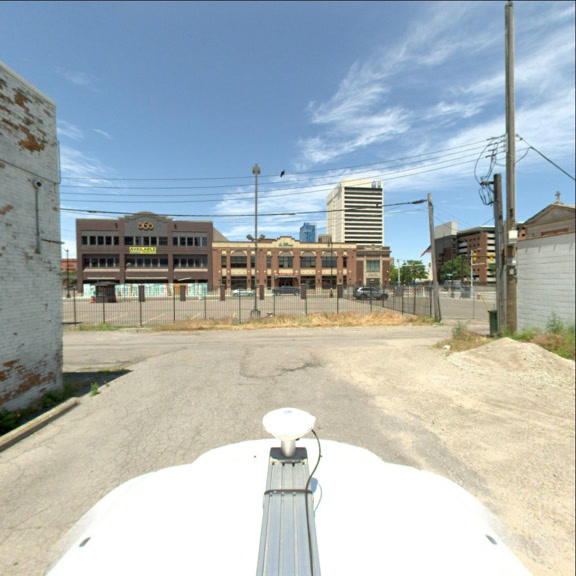} &
\includegraphics[width=\imwidth]{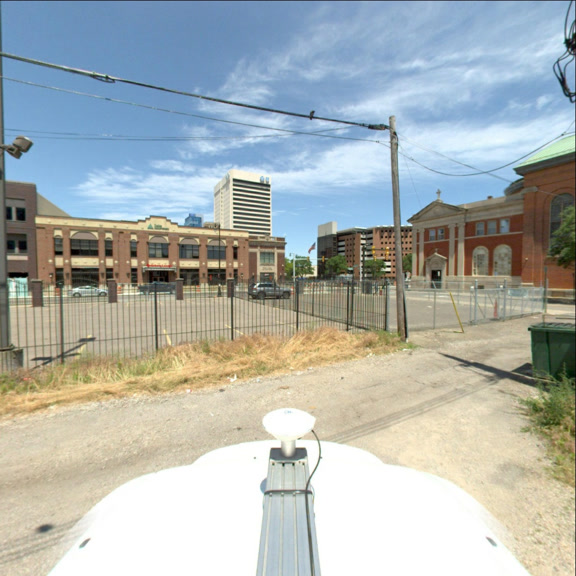} &
\includegraphics[width=\imwidth]{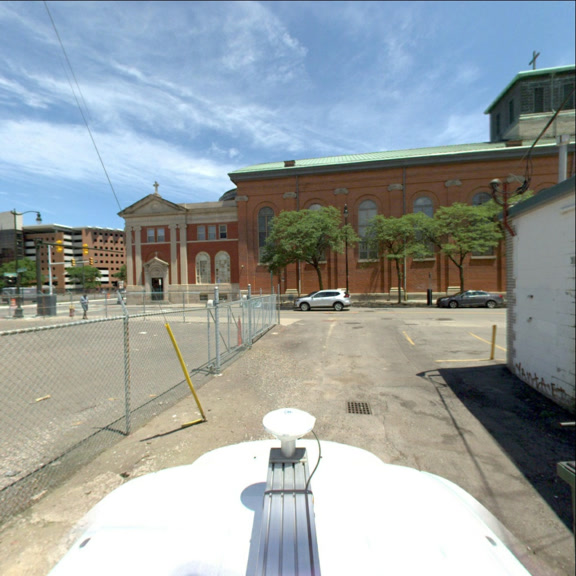} &
\includegraphics[width=\imwidth]{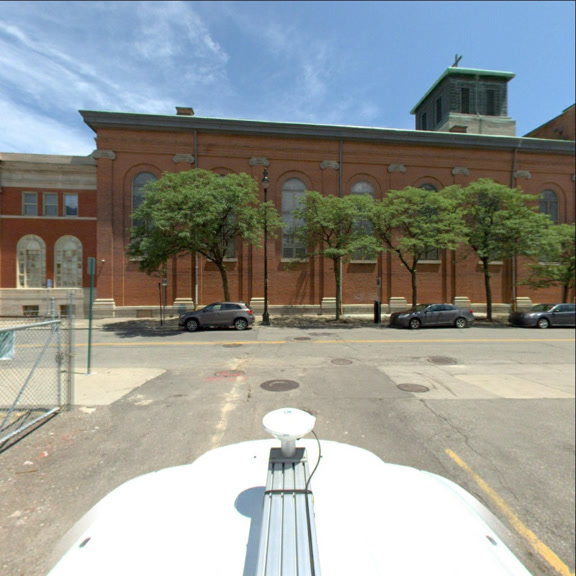} &
\includegraphics[width=\imwidth]{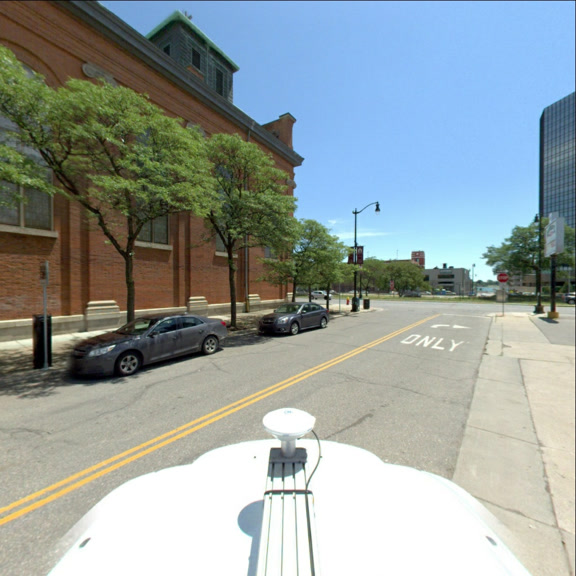}
 \\

\raisebox{4pt}{\rotatebox{90}{Target Satellite}} &
\includegraphics[width=\imwidth]{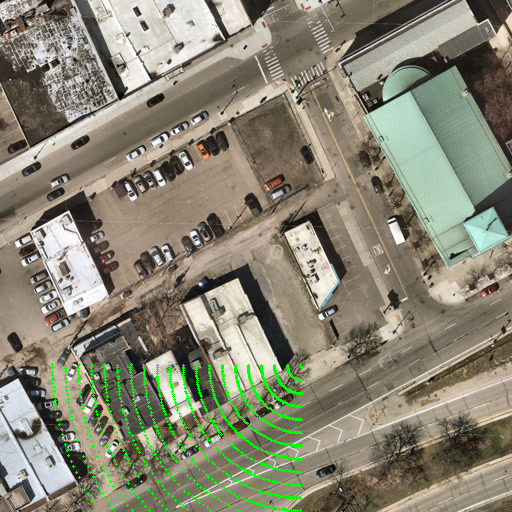} &
\includegraphics[width=\imwidth]{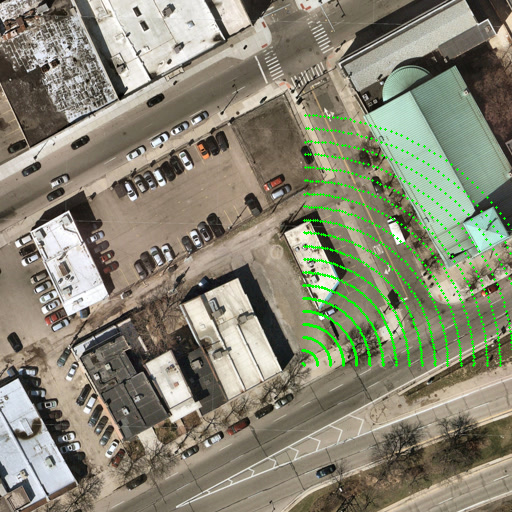} &
\includegraphics[width=\imwidth]{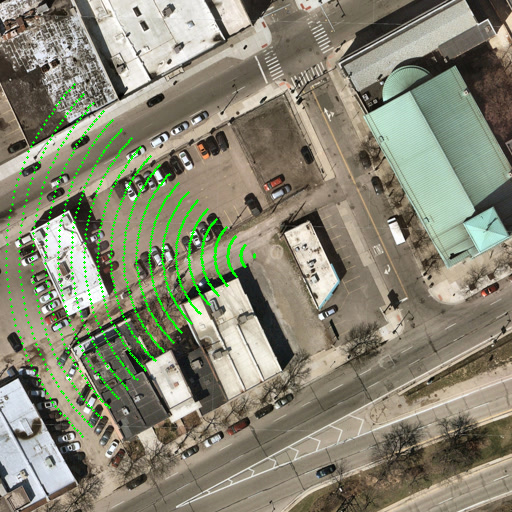} &
\includegraphics[width=\imwidth]{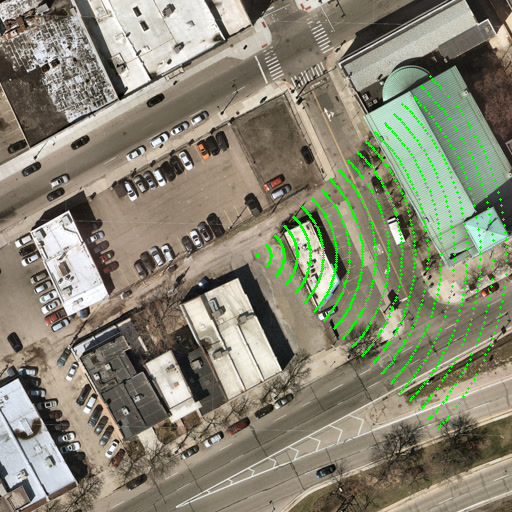} &
\includegraphics[width=\imwidth]{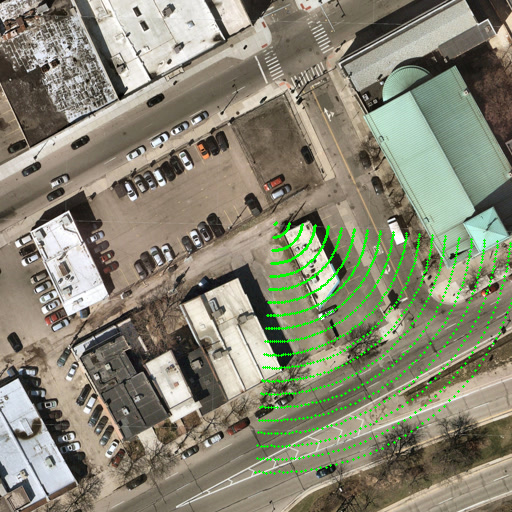} &
\includegraphics[width=\imwidth]{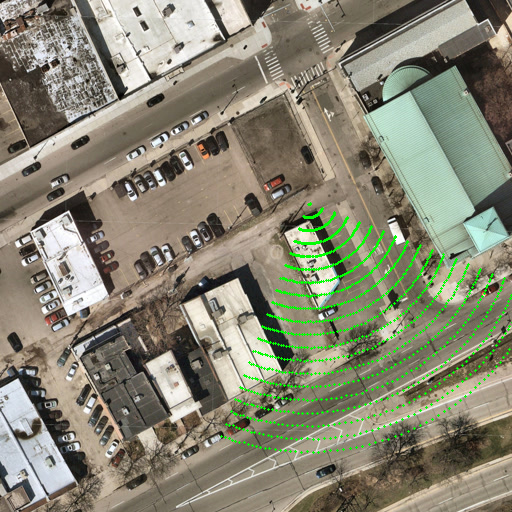} &
\includegraphics[width=\imwidth]{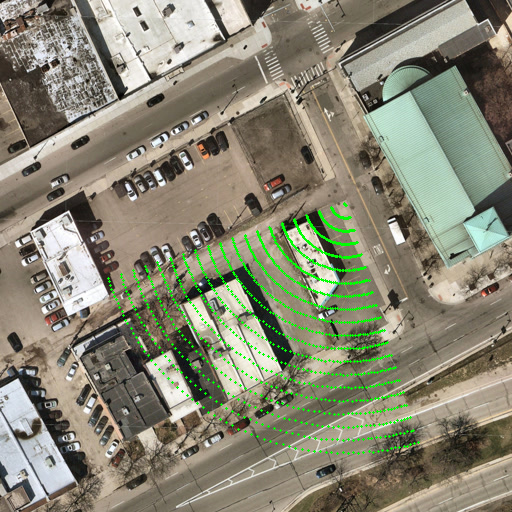}
\\
\raisebox{16pt}{\rotatebox{90}{Ground truth}} &
\includegraphics[width=\imwidth]{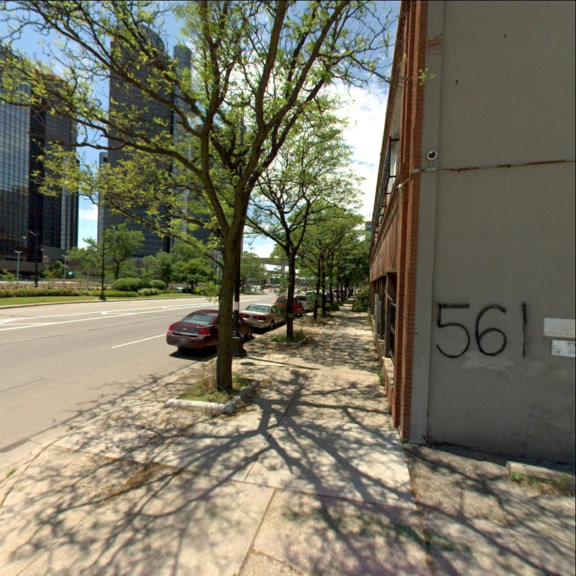} &
\includegraphics[width=\imwidth]{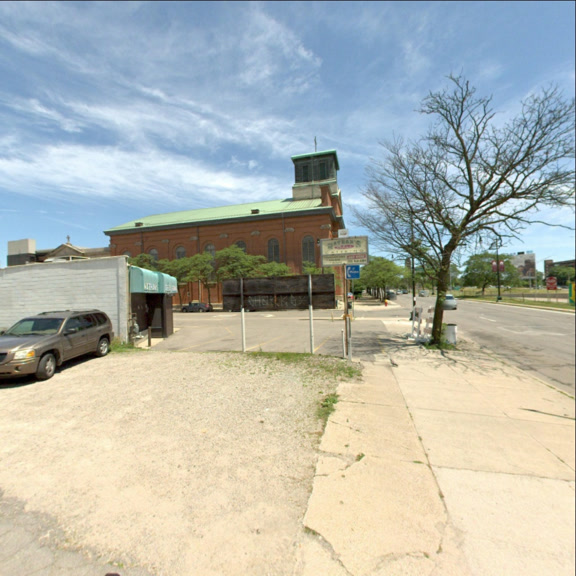} &
\includegraphics[width=\imwidth]{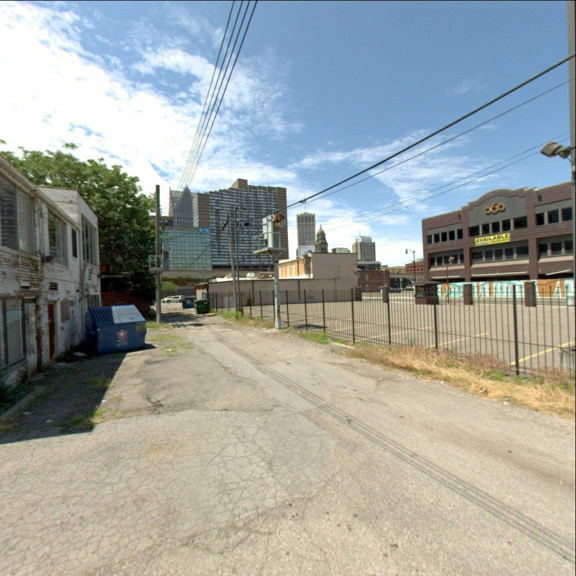} &
\includegraphics[width=\imwidth]{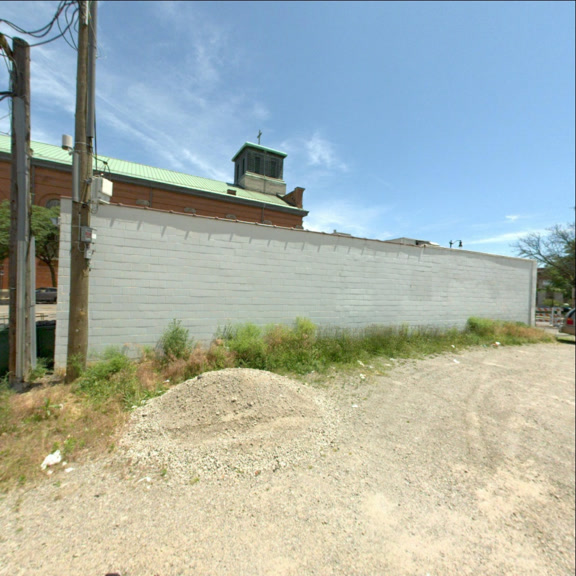} &
\includegraphics[width=\imwidth]{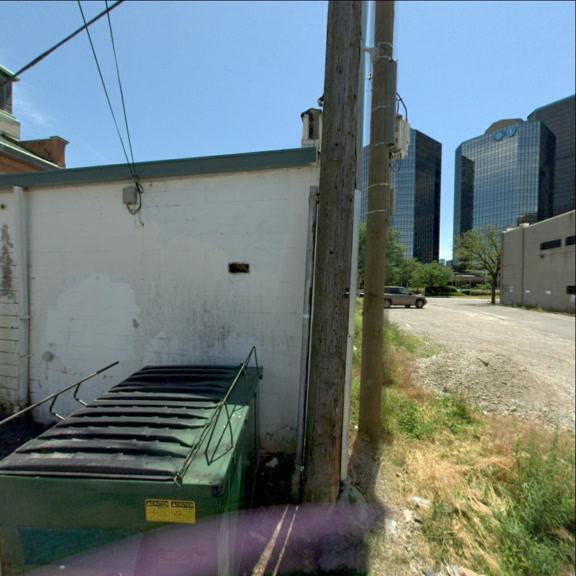} &
\includegraphics[width=\imwidth]{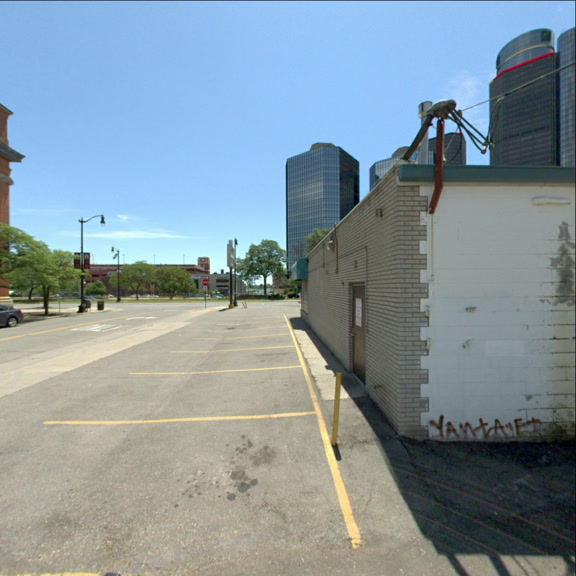} &
\includegraphics[width=\imwidth]{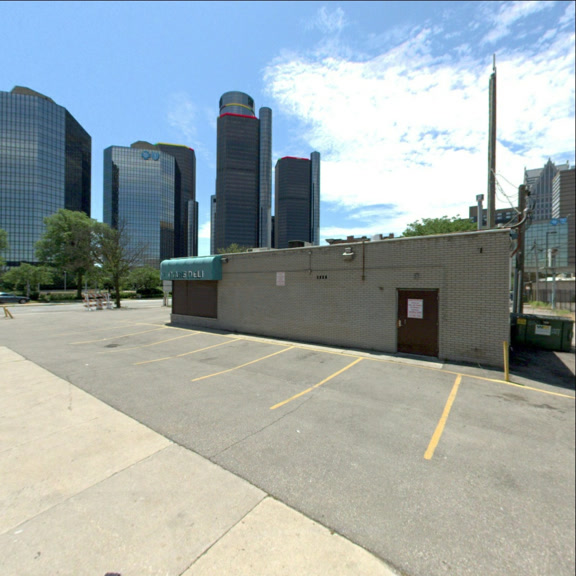}
\\
\raisebox{15pt}{\rotatebox{90}{Ours}} &
\includegraphics[width=\imwidth]{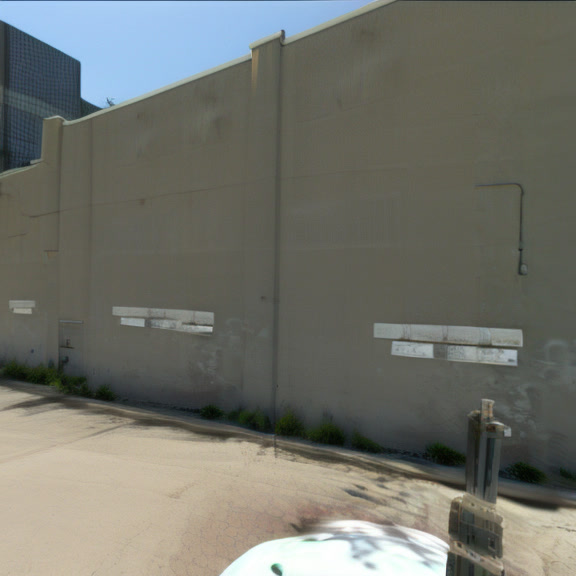} &
\includegraphics[width=\imwidth]{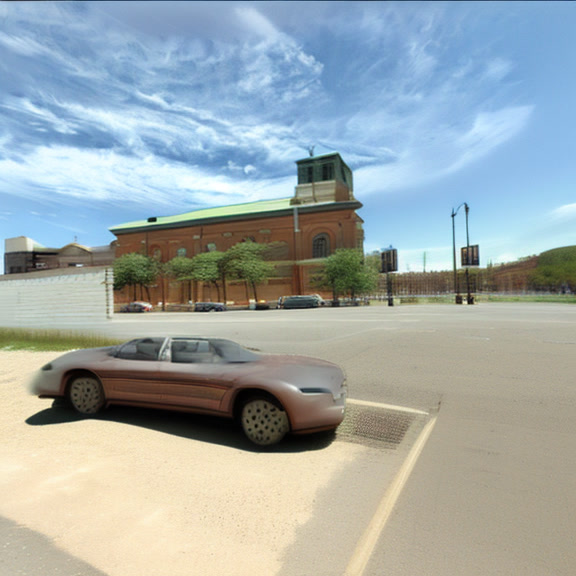} &
\includegraphics[width=\imwidth]{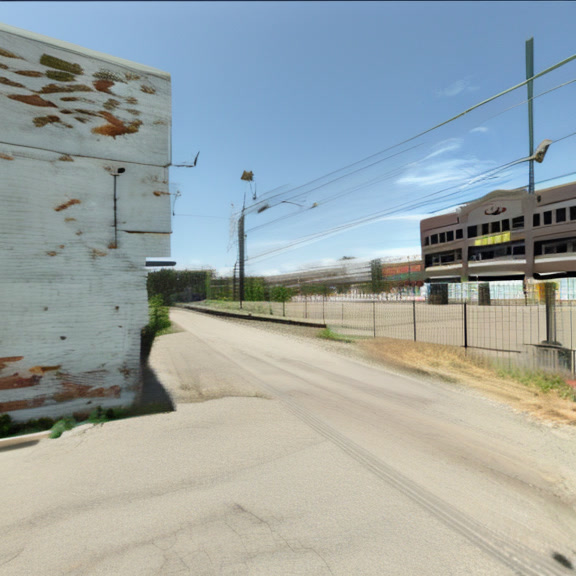} &
\includegraphics[width=\imwidth]{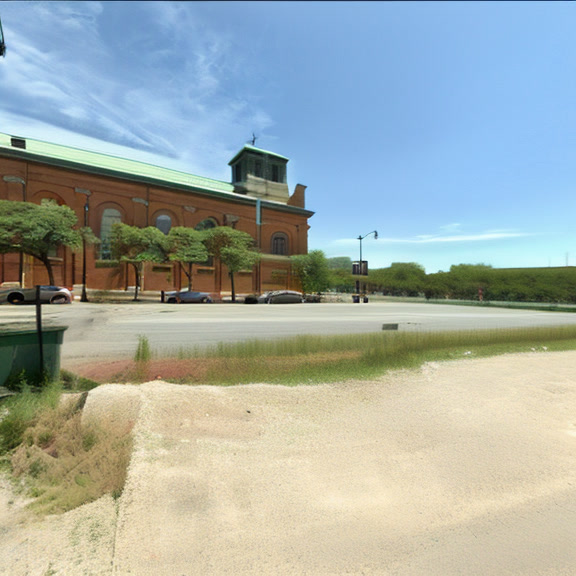} &
\includegraphics[width=\imwidth]{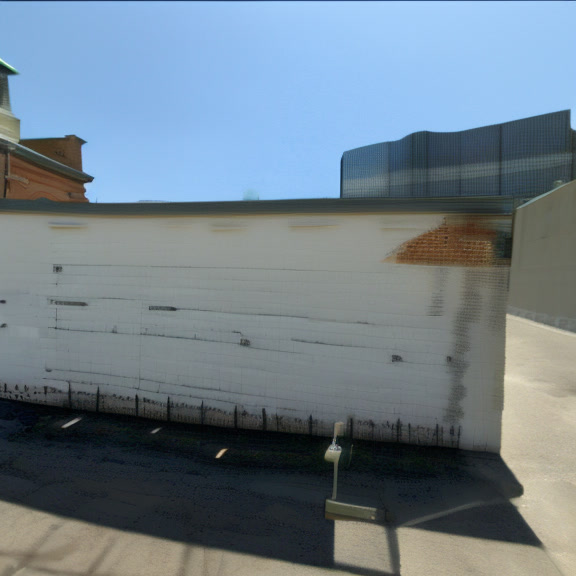} &
\includegraphics[width=\imwidth]{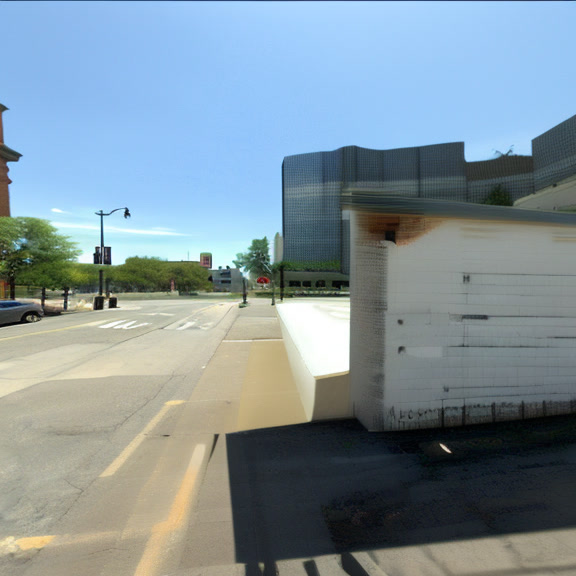} &
\includegraphics[width=\imwidth]{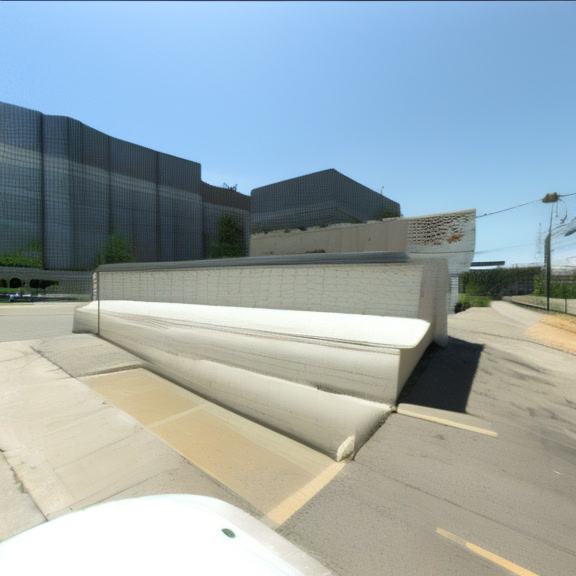}
 \\
 \raisebox{13pt}{\rotatebox{90}{SEVA}} &
\includegraphics[width=\imwidth]{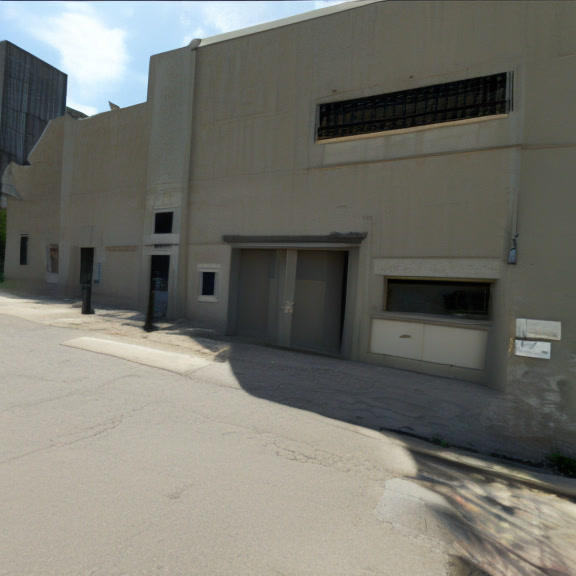} &
\includegraphics[width=\imwidth]{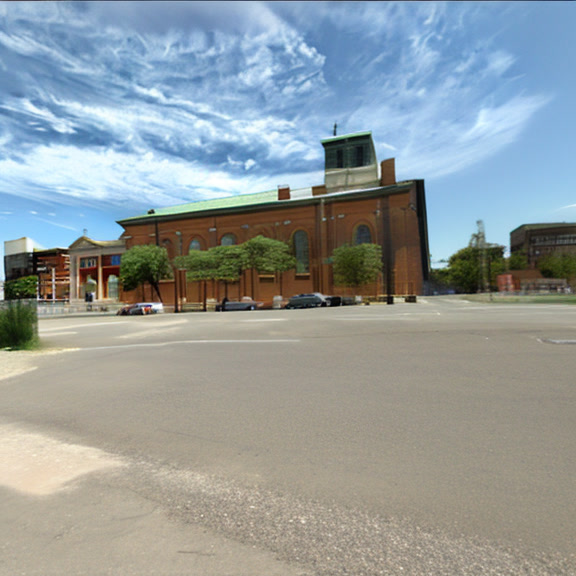} &
\includegraphics[width=\imwidth]{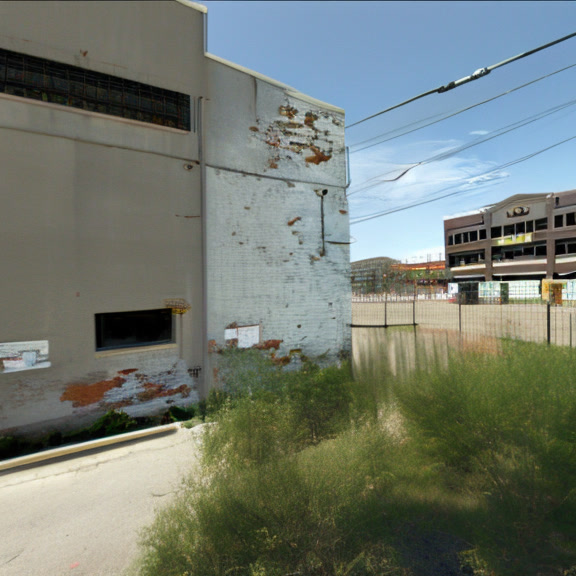} &
\includegraphics[width=\imwidth]{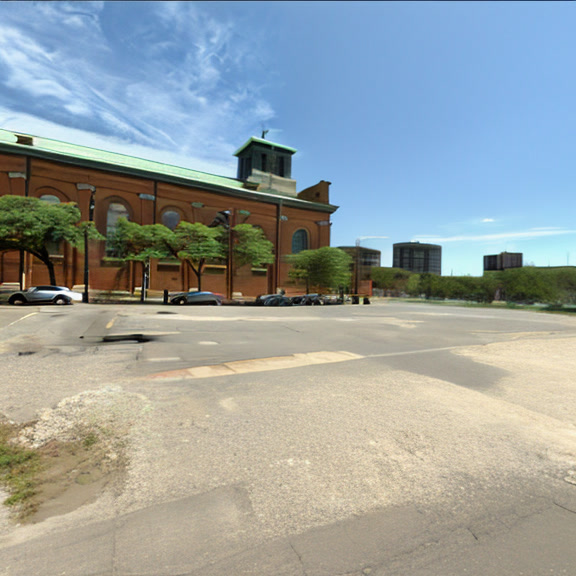} &
\includegraphics[width=\imwidth]{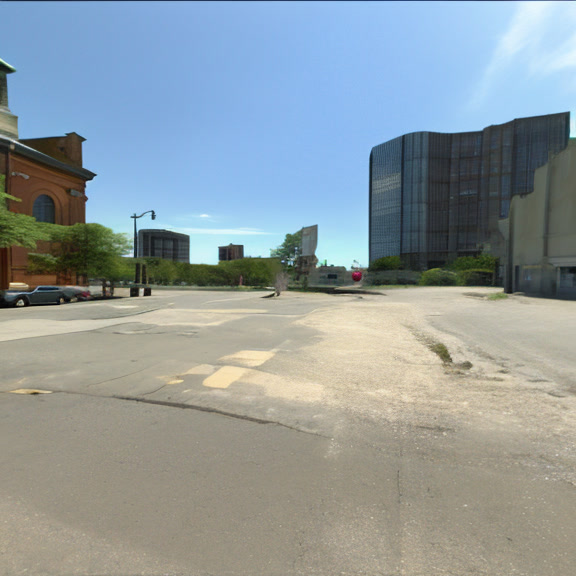} &
\includegraphics[width=\imwidth]{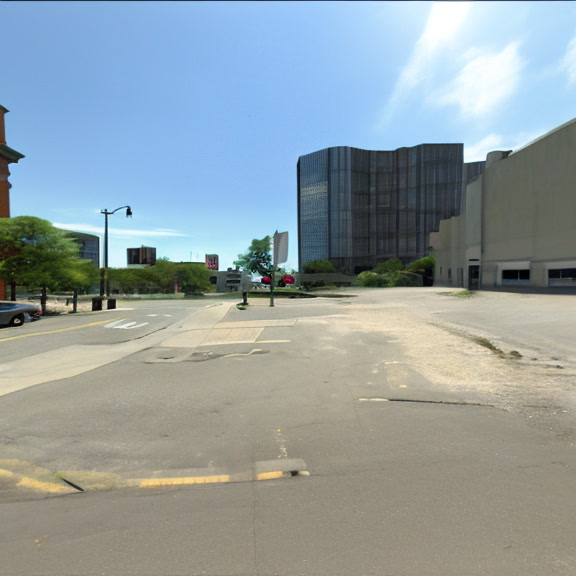} &
\includegraphics[width=\imwidth]{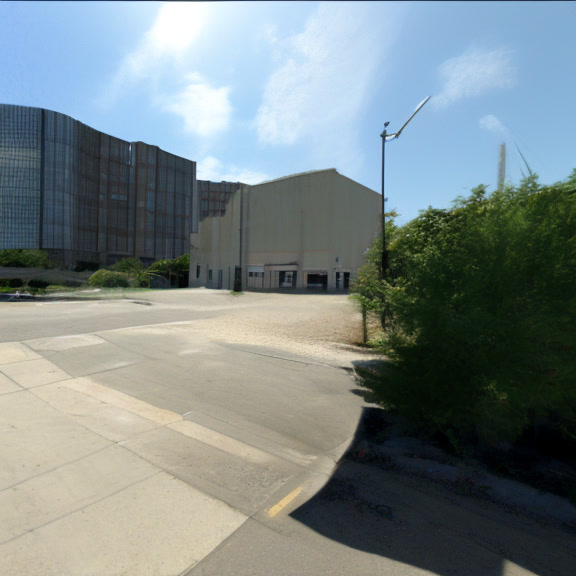}
\\

    \end{tabular}
    }
    \definecolor{softred}{HTML}{F7CCCC}
    \definecolor{softgreen}{HTML}{D6F2D6}
    \vspace{-1em}
    \caption{\textbf{Qualitative results on in-distribution dataset: Metropolis}
    Top 4 rows show satellite images along with corresponding input and target views. Final two rows show synthesis of target views using \ourmodel{} and SEVA. \ourmodel{} is significantly better than SEVA at adhering to the layout of the scene.
    }
    \label{fig:results_in_distribution2}
\end{figure*}

\renewcommand{\imwidth}{0.15\textwidth}
\begin{figure*}
    \centering
    \scalebox{0.9}{
    \begin{tabular}{@{}c@{\hskip 2pt}c@{\hskip 2pt}c|c@{\hskip 2pt}c@{\hskip 2pt}c@{\hskip 2pt}c@{}}
        &
        \scriptsize\shortstack{Satellite view ($S$)} &
        \scriptsize\shortstack{Example input} &
        \scriptsize\shortstack{Ground truth} &
        \scriptsize\shortstack{Ours}&
        \scriptsize\shortstack{SEVA} &
        \scriptsize\shortstack{Sat2Density (cropped)$\dagger$} 
        \\

\raisebox{8pt}{\rotatebox{90}{\scriptsize Metropolis}} &
\includegraphics[width=\imwidth]{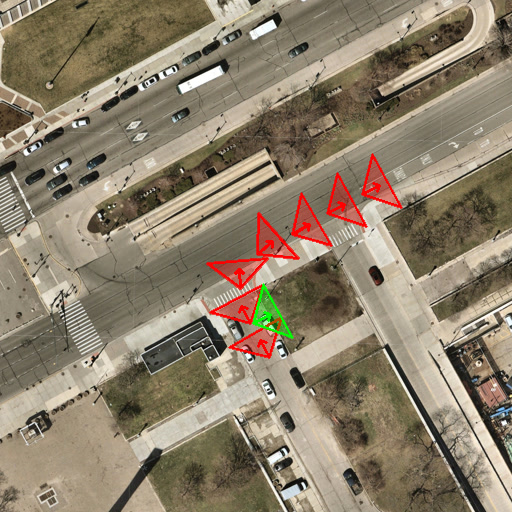} &
\includegraphics[width=\imwidth]{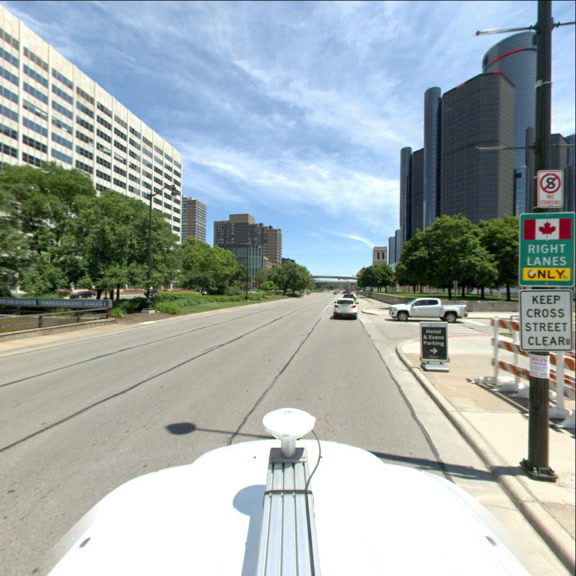} &
\includegraphics[width=\imwidth]{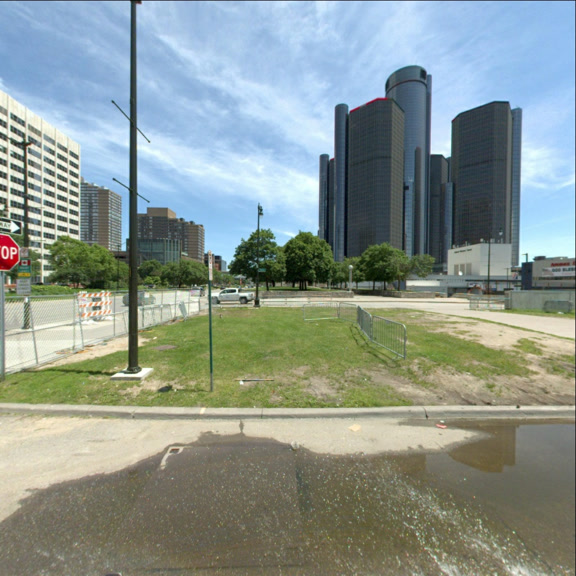} &
\includegraphics[width=\imwidth]{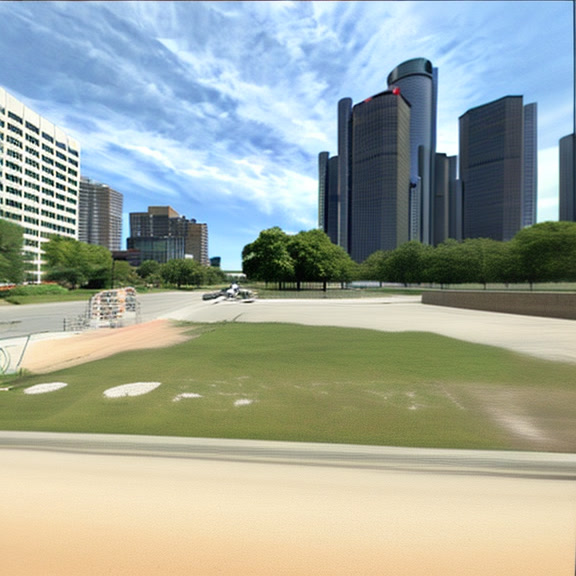} &
\includegraphics[width=\imwidth]{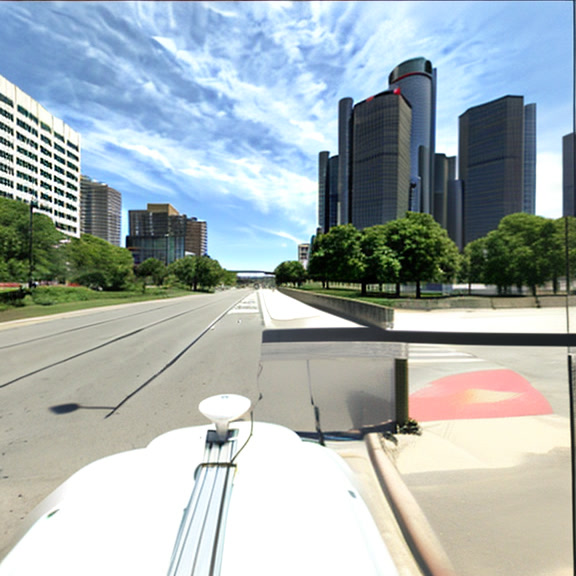} &
\includegraphics[width=\imwidth]{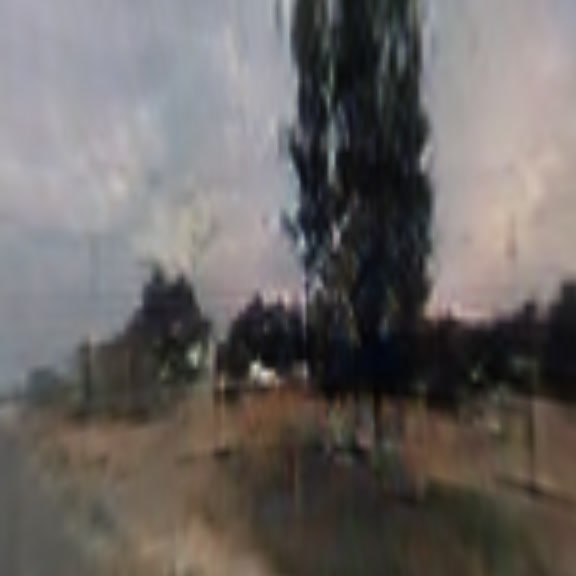} \\

\raisebox{8pt}{\rotatebox{90}{\scriptsize Metropolis}} &
\includegraphics[width=\imwidth]{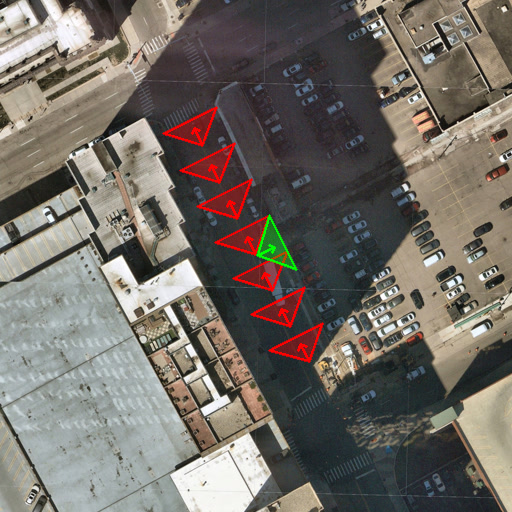} &
\includegraphics[width=\imwidth]{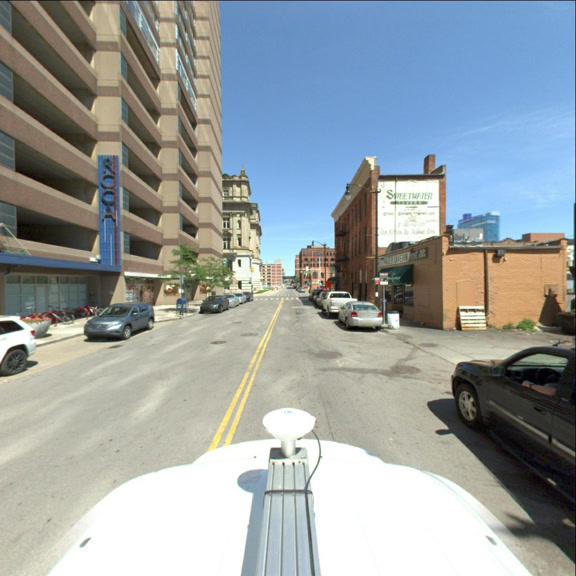} &
\includegraphics[width=\imwidth]{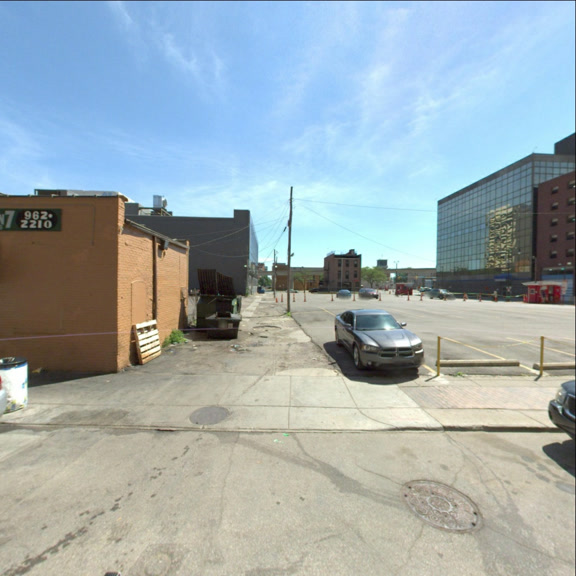} &
\includegraphics[width=\imwidth]{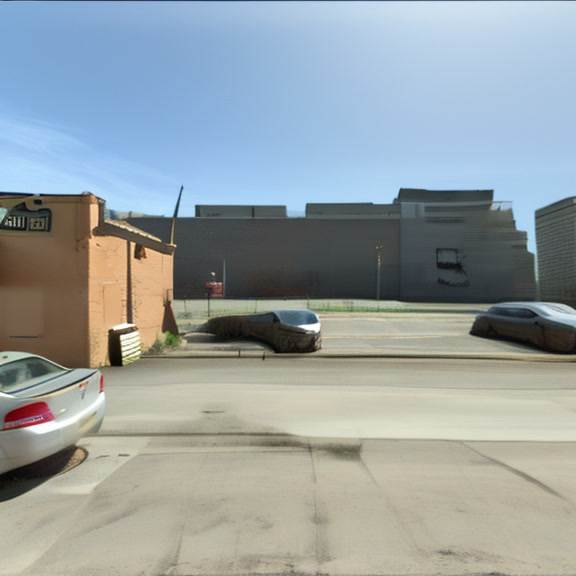} &
\includegraphics[width=\imwidth]{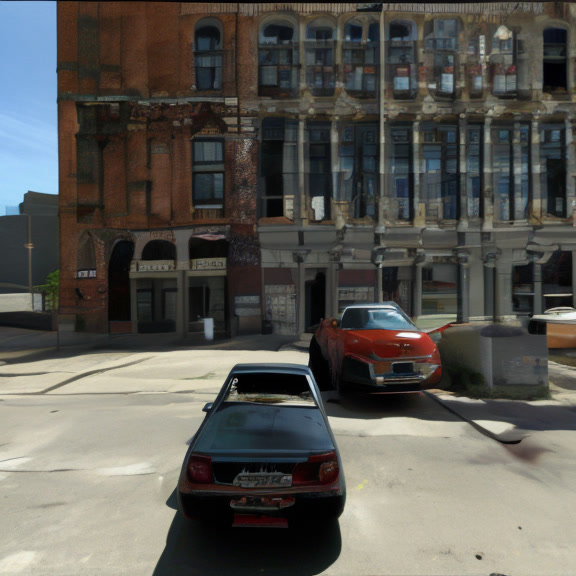} &
\includegraphics[width=\imwidth]{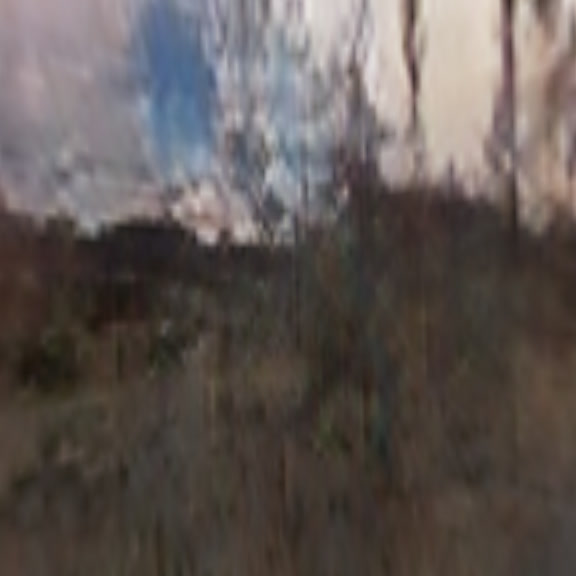} \\

\raisebox{10pt}{\rotatebox{90}{\scriptsize nuScenes}} &
\includegraphics[width=\imwidth]{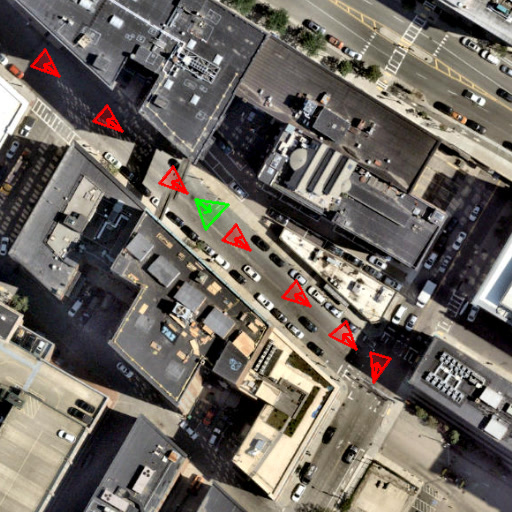} &
\includegraphics[width=\imwidth]{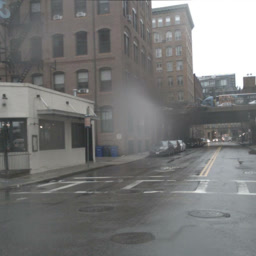} &
\includegraphics[width=\imwidth]{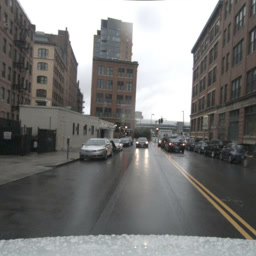} &
\includegraphics[width=\imwidth]{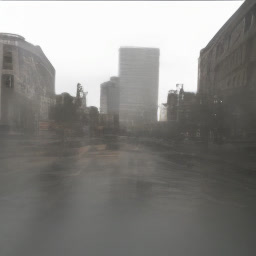} &
\includegraphics[width=\imwidth]{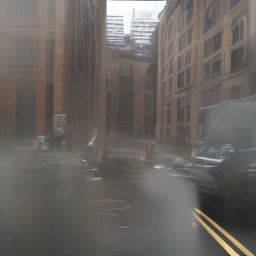} &
\includegraphics[width=\imwidth]{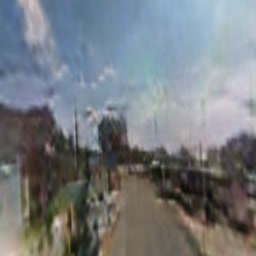} \\

\raisebox{10pt}{\rotatebox{90}{\scriptsize nuScenes}} &
\includegraphics[width=\imwidth]{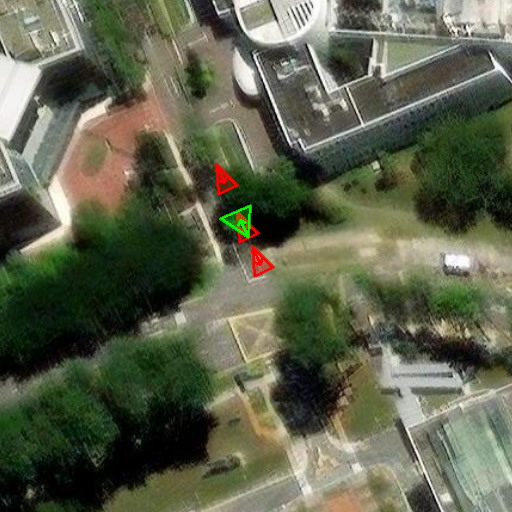} &
\includegraphics[width=\imwidth]{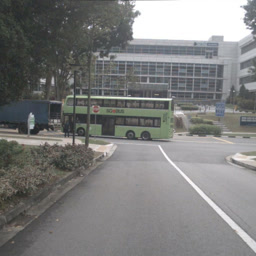} &
\includegraphics[width=\imwidth]{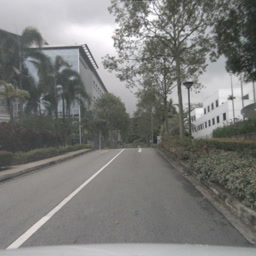} &
\includegraphics[width=\imwidth]{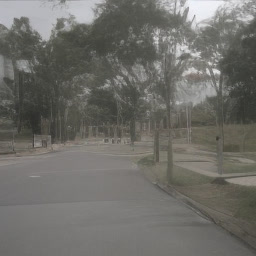} &
\includegraphics[width=\imwidth]{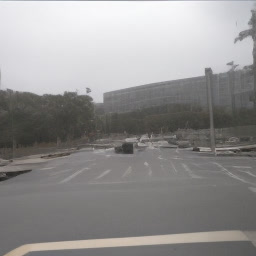} &
\includegraphics[width=\imwidth]{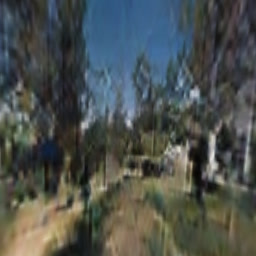}
    \end{tabular}   
    }
    \definecolor{softred}{HTML}{F7CCCC}
    \definecolor{softgreen}{HTML}{D6F2D6}
    \vspace{-1em}
    \caption{\textbf{Qualitative results on in-distribution datasets:} Metropolis and nuScenes.
    Each row shows, from left: The satellite view $S$; an example ground-level `seed' input view \inputimages; a ground truth target view \targetimages; then predictions from \ourmodel, \cite{zhou2025stable} and \cite{qian2023sat2density}.   
    Satellite images have input camera frustums plotted in {\color{red}red}, and the target frustum in {\color{green}{green}}.
    $\dagger$~We crop Sat2Density results to make pinhole images from their panoramic outputs; see supplementary for their full panoramic output. 
    }
    \label{fig:results_in_distribution}
    
\end{figure*}

\renewcommand{\imwidth}{0.13\textwidth}
\begin{figure*}
    \centering
    \begin{tabular}{@{}c@{\hskip 2pt}c@{\hskip 2pt}c|c@{\hskip 2pt}c@{\hskip 2pt}c@{\hskip 2pt}c@{}}
        &
        \scriptsize\shortstack{Satellite view ($S$)} &
        \scriptsize\shortstack{Example input} &
        \scriptsize\shortstack{Ground truth} &
        \scriptsize\shortstack{Ours}&
        \scriptsize\shortstack{SEVA} &
        \scriptsize\shortstack{Sat2Density (cropped)$\dagger$} 
        \\

        \raisebox{10pt}{\rotatebox{90}{\scriptsize \centering DL3DV}} &
         \includegraphics[width=\imwidth,height=\imwidth]{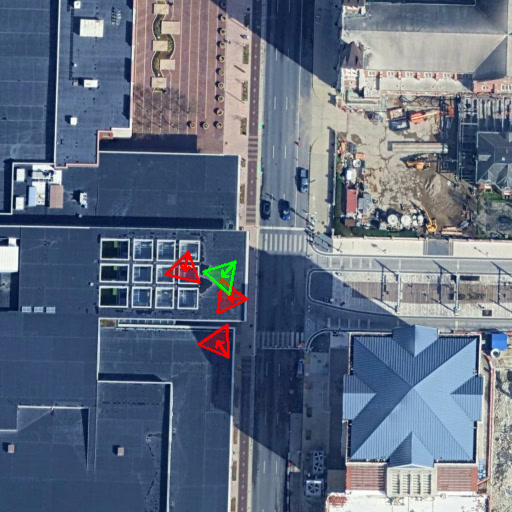} &  
         \includegraphics[width=\imwidth,height=\imwidth]{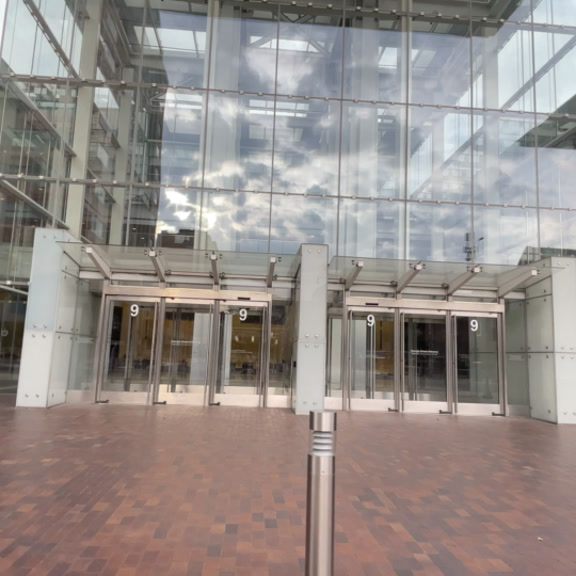} &  
         \includegraphics[width=\imwidth,height=\imwidth]{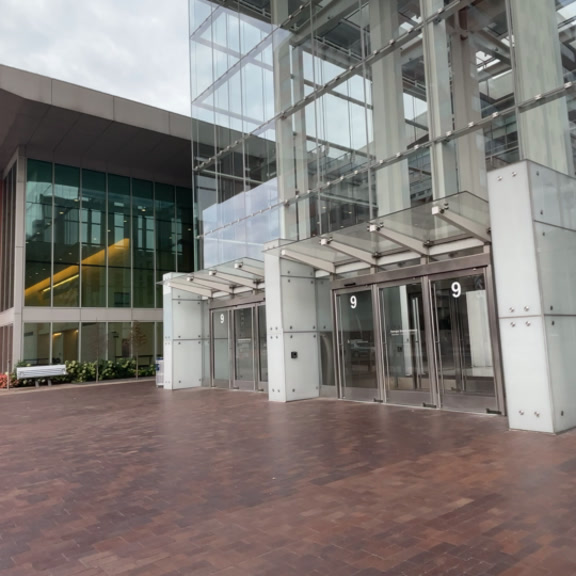} &  
         \includegraphics[width=\imwidth,height=\imwidth]{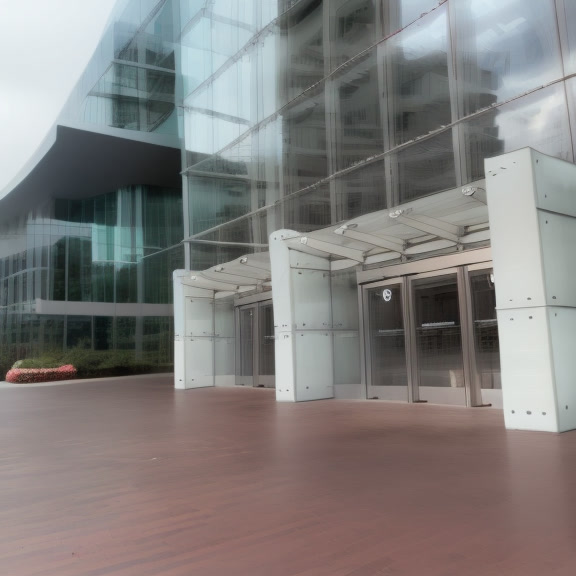} &  
         \includegraphics[width=\imwidth,height=\imwidth]{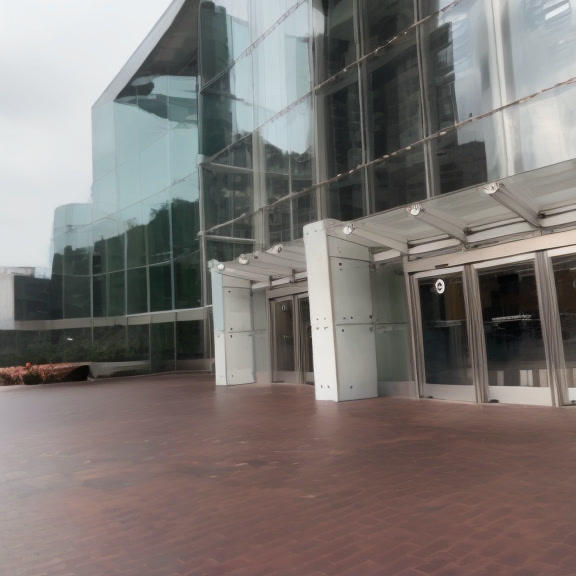} & 
         \includegraphics[width=\imwidth,height=\imwidth]{
         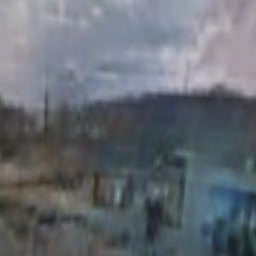} 
         \\

         \raisebox{10pt}{\rotatebox{90}{\scriptsize \centering DL3DV}} &
         \includegraphics[width=\imwidth,height=\imwidth]{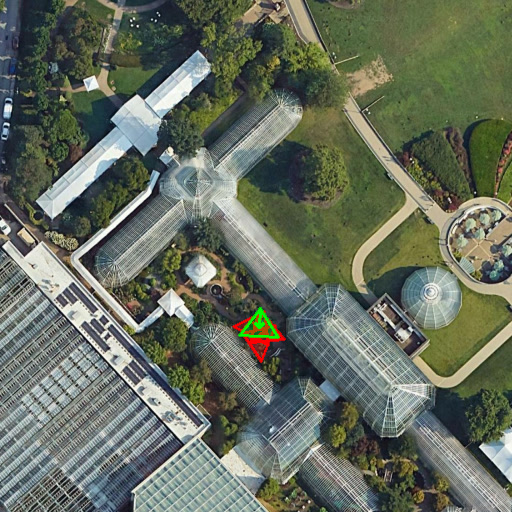} &  
         \includegraphics[width=\imwidth,height=\imwidth]{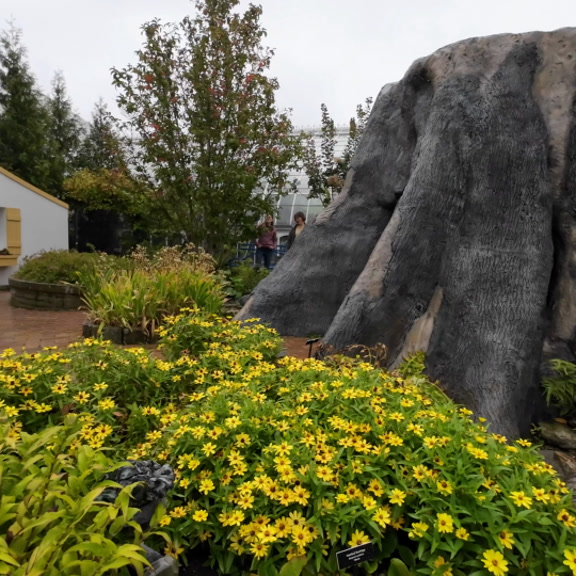} &  
         \includegraphics[width=\imwidth,height=\imwidth]{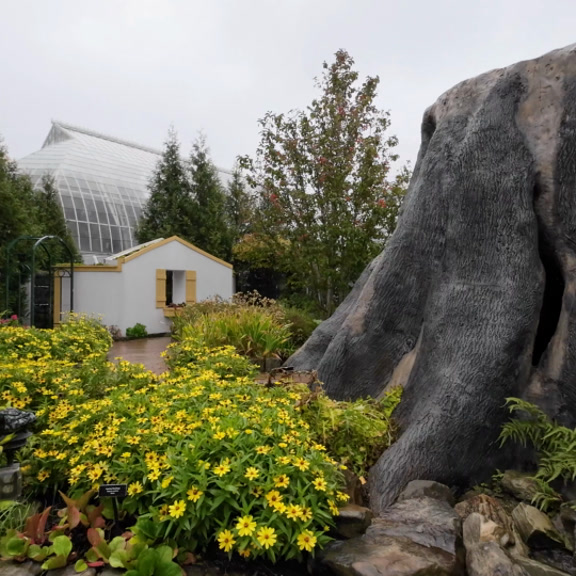} &  
         \includegraphics[width=\imwidth,height=\imwidth]{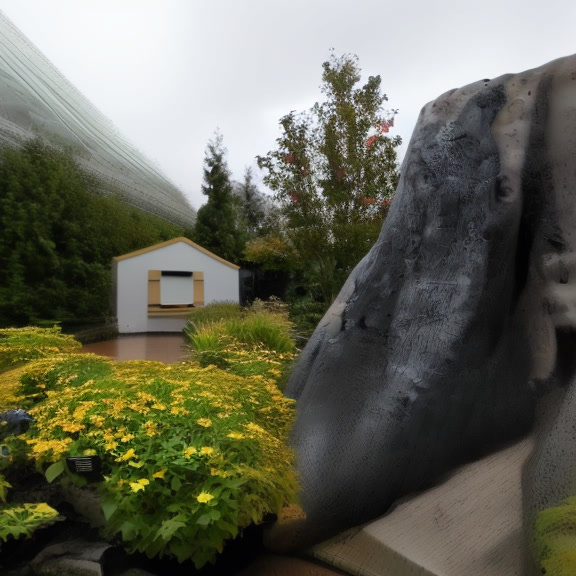} &  
         \includegraphics[width=\imwidth,height=\imwidth]{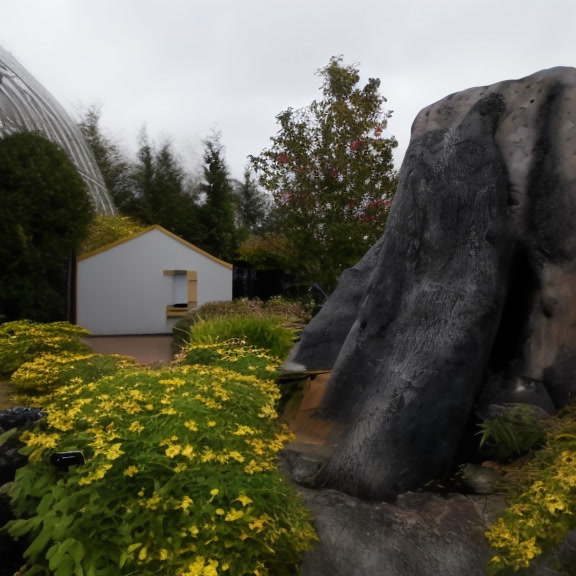} &  
         \includegraphics[width=\imwidth,height=\imwidth]
         {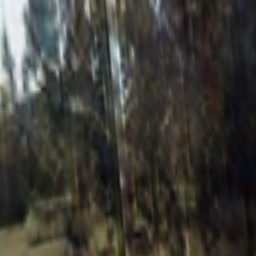} 
         \\
         
    \end{tabular}
    \vspace{-1em}
    \caption{\textbf{Qualitative results on the DL3DV benchmark:} Despite being far from the distribution of street scenes, we find that \ourmodel{} learns to leverage the satellite image when generating regions without sufficient overlap from sparse input views (\eg buildings in the background). $\dagger$~We resample Sat2Density panoramas to make pinhole images; see supplementary for their full panoramic output. {\color{red}Red} triangles in satellite view are inputs, {\color{green}green} is the shown target view.
    }
    \label{fig:results_out_of_distribution}
\end{figure*}

\section{Experiments}

Our experiments evaluate the quality of our satellite-conditioned novel view generations on outdoor datasets. 
\vspace{-1.5em}

\paragraph{Baselines.} While several works address satellite-conditioned street-view generation or multi-view/video generation in isolation, our work is unique in combining both these areas. We compare to baselines from both families for the multi-view novel-view synthesis (NVS) task. For multi-view diffusion models, we compare to the state-of-the-art SEVA \cite{zhou2025stable}. As our model adopts their backbone, we find SEVA to be the closest point of comparison, as pretraining determines NVS quality to a great extent.
Among models that only use satellite-conditioning, we compare to Sat2Density \cite{qian2023sat2density}, which is one of the very few publicly-available models, see Table~\ref{tab:related_work}.
\vspace{-1.5em}

\paragraph{Datasets.} We evaluate our model on validation / test splits of Metropolis \cite{metropolis} (36 scenes) and nuScenes \cite{nuscenes} (150 scenes). nuScenes is a self-driving dataset with a 6-camera rig: front, front-left, front-right, back, back-left,  back-right. We use $7$ front camera views as input, collected in a sequence at equal time intervals. The $35$ views from the other five cameras are used as targets. Metropolis contains a 3-camera rig facing forward, left, and right. We use the forward-facing camera as input and left and right facing cameras as target. This configuration allows us to evaluate target views with very little overlap with input views. These datasets are in-distribution with our training data.

As an out-of-distribution generalization experiment, we  evaluate on outdoor scenes from the well-established DL3DV benchmark~\cite{ling2024dl3dv}.
The original benchmark is not geo-registered; we use the geo-alignment of \cite{turkulainen2026crossviewsplatter}. We use the same input-target splits from  SEVA~\cite{zhou2025stable}: target frames are equally spaced along the trajectory (every 8th frame) and input frames are chosen by applying K-medoids clustering to the camera poses of the remaining frames. Despite this, the input-target overlap in DL3DV is substantially higher than that of our in-distribution evaluation datasets, leaving less room for improvement using satellite images. 
\vspace{-1.5em}

\paragraph{Metrics.}
Following standard practice in novel-view synthesis ~\cite{zhou2025stable} we evaluate \textbf{generation quality for the target views} (PSNR, SSIM, LPIPS). Since we focus on satellite-guided controllability in unseen regions, we  evaluate the adherence of the generated images to the satellite layout. We compute the semantic masks for roads and buildings using \cite{carion2025sam3segmentconcepts}, and report their \textbf{semantic IoU}. This serves as a good metric for layout alignment, since obtaining vectorized layout annotations for our data or generations is non-trivial.  We also report the \textbf{semantic similarity} of predictions using features from foundational semantic models as an additional layout adherence metric. In particular, we use L2 error of DINO~\cite{caron2021emerging} features and cosine similarity of Segment Anything~\cite{kirillov2023segany} (SEG in tables) encoder features as proposed in~\cite{ze2025controllable}. We evaluate the \textbf{multi-view consistency} of our generated images using SIFT correspondences \cite{schoenberger2016sfm}  on overlapping views. We report TSED (Thresholded Symmetric Epipolar Distance) \cite{Yu2023PhotoconsistentNVS, zhou2025stable} for correspondences in these views, \% of consistently generated views (with at least 10 correspondences under 8 $\text{px}^2$ error, and the median number of correspondences. We report the metrics on a single randomly sampled generation for each example. For a particular example, we use the same random noise sample for SEVA and our model. 

\vspace{-1em}
\paragraph{In-distribution results.}
Table~\ref{tab:satellite_baselines} shows that  Sat2Density struggles to recover street-level images from satellite imagery and an appearance code. The street-view-only SEVA baseline produces perceptually plausible target views from its learned multi-view prior, but has weaker semantic IoU, indicating a weaker semantic layout-adherence. \ourmodel{} improves image quality and layout-adherence on both datasets. On nuScenes, our gains are primarily in PSNR, SSIM, and semantic IoU, while LPIPS and DINO remain comparable to SEVA, thus slightly improving layout adherence while preserving perceptual quality. The gains are substantially larger on Metropolis, where SEVA is much worse than on nuScenes. \ourmodel{} improves across all metrics (\eg, 10.88$\rightarrow$12.81 PSNR,  0.291$\rightarrow$0.372 building IoU). This suggests that satellite condition is more useful when the street-view prior alone provides weaker constraints on unseen scene content.

\vspace{-1.5em}
\paragraph{Out-of-distribution results.}
On DL3DV (Table~\ref{tab:nvs_baselines}), \ourmodel{} generalizes well despite being trained only on driving datasets. It improves PSNR and SSIM, while remaining comparable to SEVA in LPIPS, SEG, and DINO. Satellite conditioning also improves multi-view consistency, reducing TSED and increasing the fraction of consistent views. Since DL3DV has much higher overlap between input and target views than Metropolis or nuScenes, there is less room for satellite conditioning to help. Sat2Density however performs substantially worse, highlighting the difficulty of generating these views from satellite imagery alone. Overall, these results suggest that satellite conditioning can improve generation quality and multi-view consistency even outside our training distribution.

\subsection{Ablations}

We ablate some of the key design choices of our method on the Metropolis dataset in Table \ref{tab:ablations}. We provide more qualitative comparisons and ablation details in the supplement. 

\noindent \textbf{Without ray-guided cross-attention (RGCA).} An alternative to using ray-guidance during cross-attention is for each street-image token to attend to all satellite tokens. We find that given our dataset size, this baseline struggles to leverage the satellite image for the novel-view synthesis, as evidenced by the lower performance in Table \ref{tab:ablations}. 
\vspace{-1.5em}

\begin{table}
\centering
\setlength{\tabcolsep}{3pt}
\renewcommand*{\arraystretch}{1}
\resizebox{\linewidth}{!}{%
\begin{tabular}{l ccc }
    \cmidrule(lr){2-4}
     & PSNR $\uparrow$  &   DINO $\downarrow$   &   SEG  $\uparrow$ \\
    \midrule
    Satellite view as input view  & 11.312 & 5.862 & 0.558 \\
    W/o ray guided cross-attention (RGCA) & 11.675 & 5.583 & 0.573 \\
    W/o Ray and timestep embed.\ in RGCA query & 11.372 & 5.309 & 0.579 \\
    UNet decoder conditioning  & 11.291 & 5.400 & 0.572 \\
    W/o separate CFG & 12.706 & \textbf{5.144} & \textbf{0.608}  \\
    Ours & \textbf{12.809} & 5.247 & 0.605 \\
    \midrule
\end{tabular}%
}
\vspace{-1em}
\caption{Ablation study on the Metropolis dataset.}
\vspace{-1.5em}
\label{tab:ablations}
\end{table}

\paragraph{UNet decoder conditioning.} We find that adding the satellite cross-attention layers in the UNet encoder performs better than in the decoder.  
\vspace{-1.5em}

\paragraph{RGCA query structure:} We add a raymap and diffusion timestep embedding to the queries in all RGCA layers (Fig.~\ref{fig:method}). We find that this is important, as queries are otherwise noisy, especially early in the diffusion process.
\vspace{-1.5em}

\paragraph{Satellite view as input view.} We consider a conditioning design that leverages the same multi-view diffusion model for the satellite images. We provide the satellite image as another input view, with a learned embedding for the plucker ray input that SEVA requires. This is comparable to methods in related work \cite{vuong2025aerialmegadepth}. It re-uses the pretrained diffusion model to encode satellite images, and the information between satellite and ground views transfers through the cross-view attention layers in SEVA. We finetune the self- and cross-view layers. As reported in Table \ref{tab:ablations}, we find that this does not condition the generation as expected. This can be explained by the lack of specialized learned modules to process satellite images, as they are significantly different from ground view images captured by consumer cameras. 
\vspace{-1.5em}
 
\paragraph{Without dual CFG.} We replace our dual CFG scheme with a more standard single-parameter CFG, using $\epsilon_t(i,s) = \epsilon_t(u) + \lambda (\epsilon_{t(i,s)} - \epsilon_{t(u)})$ instead of eq.~\ref{eq:dual-cfg}, using $\lambda=4$. We note that using different classifier-free guidance strengths for input views versus satellite images can slightly improve generation quality (PSNR).

\subsection{Qualitative Results}

Figure~\ref{fig:results_in_distribution} shows how our model compares with baselines \cite{zhou2025stable} and \cite{qian2023sat2density} on driving datasets \cite{nuscenes} and \cite{metropolis}.
Each row shows a different scene from the two datasets.
The first image shows the satellite view $S$; the second is an example input view.
The final four columns show the same target viewpoint: with the ground truth target \targetimages, our generation,  
 SEVA \cite{zhou2025stable}, and the Sat2Density \cite{qian2023sat2density} baselines respectively.
We crop Sat2Density outputs from full panoramas to match the FOV of our target cameras.
We observe that our method captures the structure of unseen parts of the scene better than the baselines, giving more accurate placements of trees, buildings and street furniture.
In the supplementary material we show visualize the attention maps for the satellite images.

\section{Conclusion}

We have introduced \ourmodel, which enables effective satellite conditioning of multi-view diffusion models.
Our architectural innovations ensure the satellite image is incorporated efficiently and effectively.
Our extensive evaluations on public benchmark datasets validate the utility of our contributions and set a new state-of-the-art for free-viewpoint satellite-conditioned novel-view synthesis.
\vspace{-0.5em}

\small
\paragraph{Acknowledgements} We thank Filippo Aleotti, Gabriel Brostow, and Alan Paul for insightful discussions and support.

{
    \small
    \bibliographystyle{ieeenat_fullname}
    \bibliography{main}
}

\clearpage
\appendix
\setcounter{figure}{0}
\setcounter{table}{0}
\renewcommand{\thefigure}{\Roman{figure}}
\renewcommand{\thetable}{\Roman{table}}
\maketitlesupplementary
\newcommand{\satpanel}[1]{%
  \begin{minipage}[b]{.6\textwidth}
    \includegraphics[width=\linewidth]{figures/supplementary/#1.jpg}\vspace{2pt}
    \includegraphics[width=\linewidth]{figures/supplementary/#1_region.jpg}
  \end{minipage}\hfill
  \begin{minipage}[b]{.3\textwidth}
    \includegraphics[width=\linewidth]{figures/supplementary/#1_gt.jpg}
  \end{minipage}%
}

This supplementary material provides additional experiments and analysis, more qualitative results, along with more details on datasets, the model, and more baselines. In Section \ref{sec:exp_details}, we provide additional experiments that evaluate our robustness to satellite alignment, quantify performance across viewpoint overlaps, and provide additional baselines. We provide additional details on our training data in Section \ref{sec:supp_training_data} and the ray-guided cross-attention architecture in Section \ref{sec:supp_model_details}. We provide more qualitative results in \ref{sec:supp_qual_results}, highlighting our multi-view consistency, challenging cases, comparisons on our ablations, and a discussion on the impact of dual CFG. We also include a discussion of our limitations and directions for future work. 

\section{Additional Experimental Details}
\label{sec:exp_details}

\subsection{Performance across different input-target view overlap levels}
\label{sec:psnr_vs_overlap}

\begin{figure*}
    \centering
    \includegraphics[width=\textwidth]{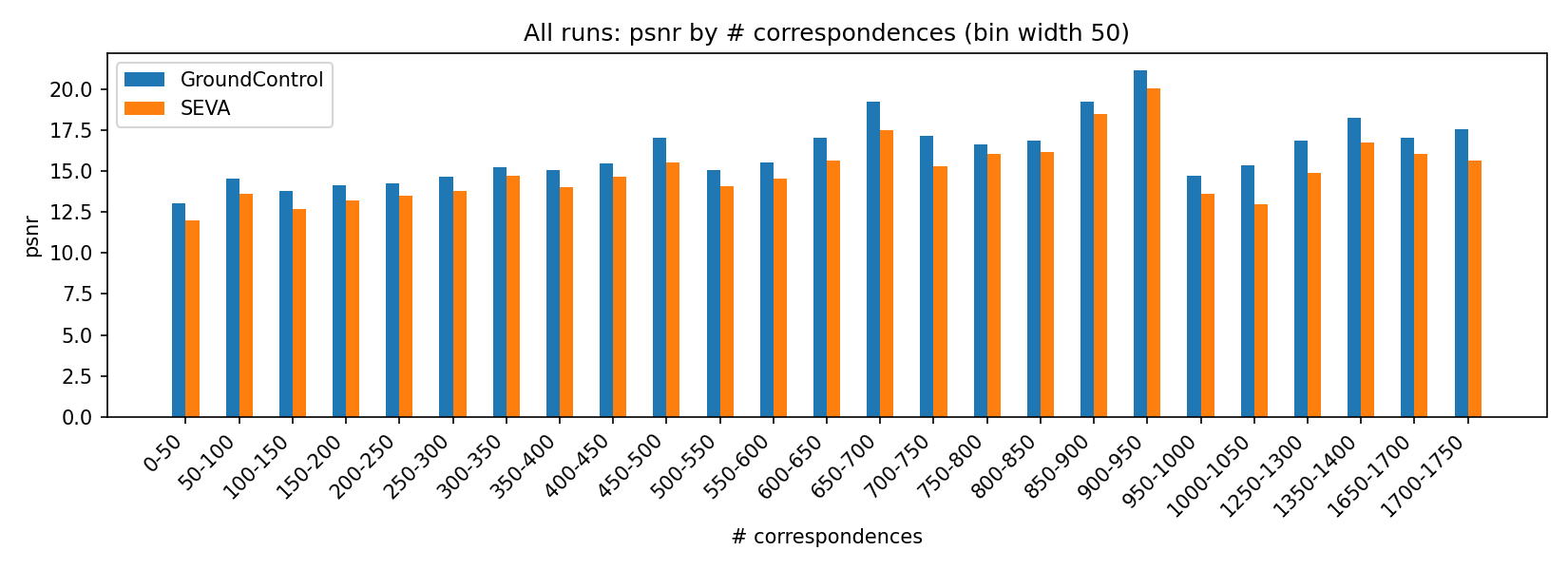}
    \vspace{-5pt}
    \caption{\textbf{PSNR performance vs. image overlap measured by the number of correspondences on Metropolis dataset}. \ourmodel obtains better performance compared to the SEVA baseline across all overlap regimes.}
    \vspace{-5pt}
    \label{fig:psnr_vs_overlap} 
\end{figure*}

In this section we compare the interpolation vs extrapolation capabilities of our model. So, we evaluate how our model's performance varies across different levels of overlap between input and target views. We use 20 scenes from Metropolis, and sample 20 different input-target view configurations for each scene at random. For each configuration, we compute the maximum number of SIFT correspondences for each target view across all input views. This indicates the maximum amount of visual overlap for that target view. We plot the PSNR for the target views bucketed by this indicator of visual overlap in Figure \ref{fig:psnr_vs_overlap}. We find that our model provides improvements across all levels of visual overlap. The improvement in the low-overlap region ($< 200$ correspondences) is noticeably consistent. 

\subsection{Robustness to satellite pose misalignment} 

Our method assumes that the transformation from the reference frame used in the diffusion model to the satellite image is known. In practice, this geo-alignment can be achieved using methods like OrienterNet \cite{sarlin2023orienternet}. To benchmark on this more practical setting, we introduce random pose offsets in our satellite alignment for nuScenes and DL3DV in Table \ref{tab:sat_misalign}. We add random translation and rotation offsets to the alignment (upto 10m translation, $10\degree$ rotation offset). OrienterNet falls within this range (99.5\% recall $<$ 5m, $5\degree$). GroundControl performance only suffers very little  degradation in this range, and remains competitive with SEVA with improvements in PSNR/SSIM. 

\begin{table}[t]
\centering
\scriptsize
\setlength{\tabcolsep}{2.2pt}
\renewcommand{\arraystretch}{1.05}
\resizebox{\columnwidth}{!}{%
\begin{tabular}{l cccc cccc}
\toprule
& \multicolumn{4}{c}{nuScenes} & \multicolumn{4}{c}{DL3DV} \\
\cmidrule(lr){2-5} \cmidrule(lr){6-9}
offset (m,\degree)
& PSNR$\uparrow$ & SSIM$\uparrow$ & DINO$\downarrow$ & Seg$\uparrow$
& PSNR$\uparrow$ & SSIM$\uparrow$ & DINO$\downarrow$ & Seg$\uparrow$ \\
\midrule
SEVA
& 11.334 & 0.308 & \textbf{6.140} & 0.492
& 14.912 & 0.360 & \textbf{3.534} & \textbf{0.674} \\
0,0
& \textbf{11.880} & \textbf{0.331} & 6.145 & \textbf{0.498}
& \textbf{15.170} & \textbf{0.372} & 3.653 & \textbf{0.674} \\
\midrule
5,5
& 11.518 & 0.311 & 6.156 & 0.490
& 15.131 & \textbf{0.372} & 3.679 & 0.670 \\
10,10
& 11.509 & 0.311 & 6.156 & 0.490
& 15.124 & 0.371 & 3.682 & 0.669 \\
\bottomrule
\end{tabular}%
}
\vspace{-1em}
\caption{\textbf{Robustness to pose misalignment:} Performance degradation is mild with alignment offsets.}
\label{tab:sat_misalign}
\end{table}

\subsection{Additional baselines}

In the main paper, we use Sat2Density as our satellite conditioned street view generation baseline. Although it mainly relies on the satellite image, it also uses the input street-level views to compute a global appearance embedding that contains lighting information, which the generation model is conditioned on. For our novel-view synthesis diffusion baseline, we primarily compare to SEVA \cite{zhou2025stable}. Since GroundControl is built on top of SEVA, it is comparable in terms of number of parameters and pretraining data. Here we discuss comparisons to more novel-view synthesis baselines. We also discuss comparisons to video generation models. 

\subsubsection{Video generation models}

Vanilla image-to-video (I2V) diffusion models are ill-suited for novel-view synthesis (NVS) as they cannot generate specific target viewpoints. Even when finetuned for camera conditioning, the target frames are required to form a continuous trajectory, which is incompatible with our task of predicting arbitrary viewpoints. They are also trained with only 1 or 2 input-view conditioning. Our task involves multiple input views. We make a best-effort comparison of our model to some video generation models as described below. 

\paragraph{ViewCrafter on DL3DV} We compared against the ViewCrafter \cite{yu2024viewcrafter} baseline in Table 3 of our paper. It uses a video diffusion model for novel-view synthesis by conditioning it on partial scene point clouds, obtained from a sparse set of input views. It also assumes that the frames to be generated sequentially follow the first frame (input). We therefore provide it an additional view, in the form of the first target view. This makes the comparison unfair (in ViewCrafter's favor). We use the 4 views to construct the point cloud. We then render the point cloud for 25 test views and use the diffusion model to generate the corresponding frames conditioned on these renders. As the model can only generate 25 frames at a time, and our splits have about 40 target views, we repeat this process for subsequent frames. When a set of frames has fewer than 25 test views, we duplicate the last test frame as needed so that we always generate a full set of 25 frames. Despite this, both SEVA and our model significantly outperform ViewCrafter in Table 3 (main paper). 

\paragraph{Comparison on continuous trajectories} To compare against the more recent video models, \eg RealCamI2V \cite{li2025realcam} and Veo 3.1 (proprietary), we created a 1-view (first frame as input) and 2-view (first and last frames as input) sequential input-target split on Metropolis in Tab.~\ref{tab:i2v_models}. This is a specialization of the more general novel-view synthesis task that restricts the target views to a continuous trajectory. We further divide this into straight and turning trajectories to analyze model performances separately in the two cases. RealCamI2V is a camera-controlled video model that can be conditioned on a single input frame only. Even on this video generation task, we are slightly better than RealCamI2V (noticeably better on PSNR, while being slightly worse on DINO and segment-anything feature similarity). We also compare against the proprietary Veo model on a subset of Metropolis with 2 input views (Veo supports a start and end frame conditioning). It is a proprietary model likely trained with more data and model parameters. It is better than our model on the straight trajectories, while being comparable on turns. 

\begin{table}[t]
\centering
\scriptsize
\setlength{\tabcolsep}{2pt}
\resizebox{\columnwidth}{!}{%
\begin{tabular}{clcccccccc}
\toprule
\multirow{2}{*}{Views} & \multirow{2}{*}{Model}
& \multicolumn{4}{c}{Straight} & \multicolumn{4}{c}{Turns} \\
\cmidrule(lr){3-6} \cmidrule(lr){7-10}
& & PSNR$\uparrow$ & LPIPS$\downarrow$ & DINO$\downarrow$ & Seg.$\uparrow$ & PSNR$\uparrow$ & LPIPS$\downarrow$ & DINO$\downarrow$ & Seg.$\uparrow$ \\
\midrule
\multirow{3}{*}{1}
& RealCamI2V & 11.57 & 0.508 & \textbf{4.482} & \textbf{0.627} & 10.28 & \textbf{0.599} & \textbf{5.672} & \textbf{0.557} \\
& SEVA       & 12.35 & \textbf{0.505} & 4.718 & 0.611 & 10.48 & 0.616 & 5.968 & 0.523 \\
& Ours       & \textbf{12.73} & 0.518 & 4.687 & 0.623 & \textbf{11.12} & 0.612 & 5.787 & 0.548 \\
\midrule
\multirow{3}{*}{2}
& \textit{Veo 3.1}    & \textit{\textbf{14.81}} & \textit{\textbf{0.395}} & \textit{\textbf{3.245}} & \textit{\textbf{0.707}} & \textit{\textbf{11.99}} & \textit{\textbf{0.538}} & \textit{5.068} & \textit{0.574} \\

& SEVA       & 12.99 & 0.472 & 4.427 & 0.633 & 11.29 & 0.545 & 5.196 & 0.571 \\
& Ours       & 13.69 & 0.474 & 4.313 & 0.654 & 11.82 & 0.545 & \textbf{5.057} & \textbf{0.593} \\
\bottomrule
\end{tabular}
}
\vspace{-1em}
\caption{Comparison to image-to-video diffusion models on the Metropolis dataset for a video generation task (continuous trajectories). We divide our test set into straight car trajectories vs turns. Our model is slightly better than RealCamI2V in both cases, while being comparable to Veo on turns, with better semantic-feature metrics.}
\label{tab:i2v_models}
\vspace{-1em}
\end{table}

\subsubsection{Feedforward Regression Models}

\begin{figure*}
    \centering
    \includegraphics[width=0.8\textwidth]{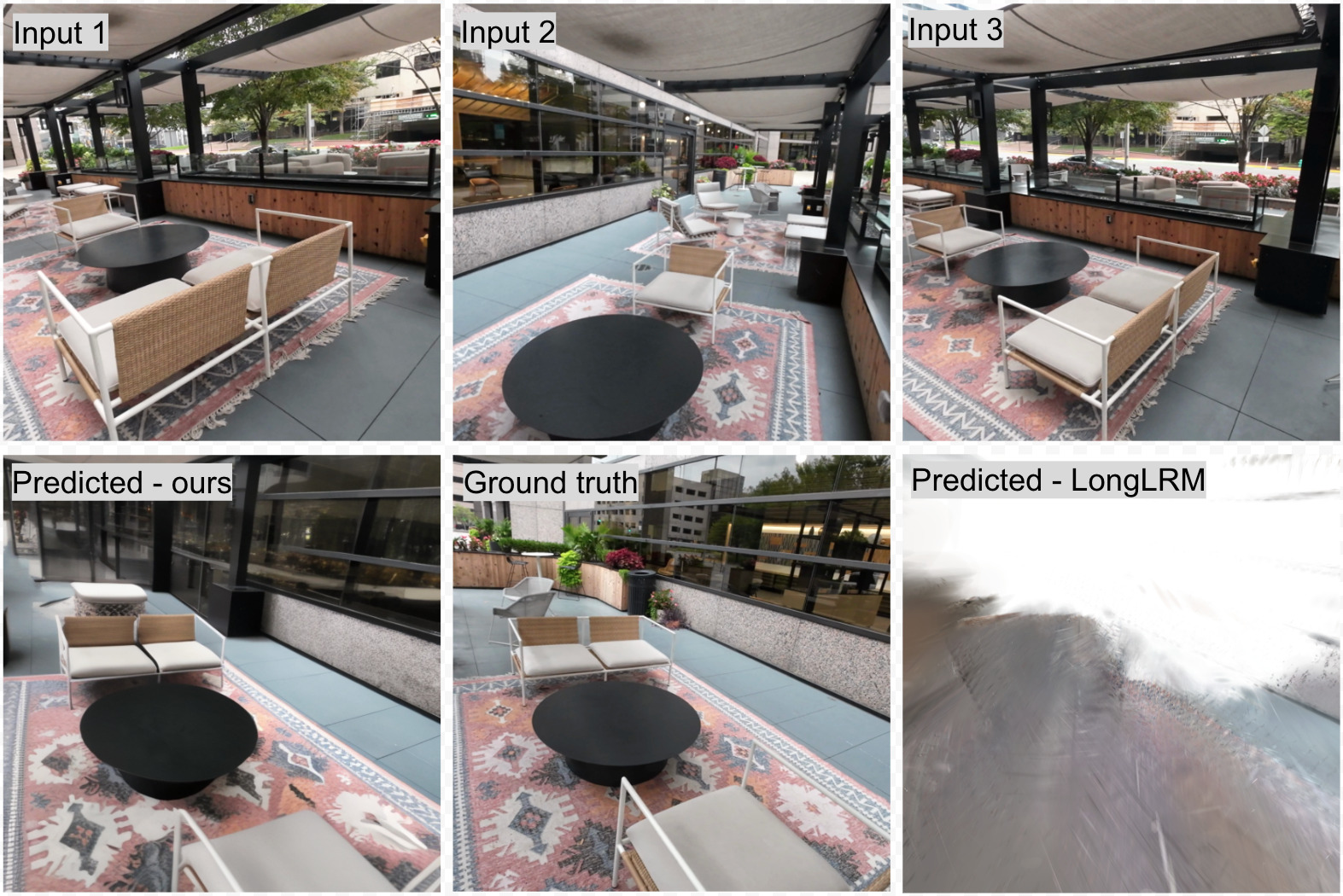}
    \vspace{-5pt}
    \caption{
        Top row: some input views from our 3-view DL3DV split. Bottom row: a test view, showing the ground truth, ours, and LongLRM.
    }
    \vspace{-5pt}
    \label{fig:longlrm}
\end{figure*}

In our evaluation we exclude baselines which are trained for the task of reconstruction, rather than generation. This is because our task focuses on synthesizing accurate content in regions which are not visible in the input views using information from the satellite view. Comparing to models which do not even attempt to generate such content is unfair to those models and uninformative to the performance on our desired task. In Fig.~\ref{fig:longlrm}, we illustrate this with an example output from LongLRM~\cite{ziwen2025llrm} on a scene from our 3-view DL3DV test split. We repeat our 3 input views so that LongLRM gets 32 input views total, and run inference at its native resolution, both of which favor its performance. LongLRM does not perform well because much of the target view is not visible in the input views at all (resulting in much of the target view being blank), and the highly extrapolative nature of the task means that visible areas of the scene also do not look good. For this reason we do not consider feedforward reconstruction models such as LongLRM to be suitable baselines for our task. 

\subsection{Effectiveness of satellite imagery}

\begin{table}[t]
\centering
\scriptsize
\setlength{\tabcolsep}{3.2pt}
\resizebox{\columnwidth}{!}{%
\begin{tabular}{lrrrrrr}
\hline
Dataset & p100 & p75 & p50 & p25 & p10 & p0 \\
\hline
nuScenes & 2.714 & 0.850 & 0.527 & 0.221 & -0.095 & -0.817 \\
Metropolis & 3.444 & 2.396 & 1.962 & 1.449 & 0.941 & 0.258 \\
\hline
\end{tabular}%
}
\vspace{-1em}
\caption{Distribution (percentile values) of per-scene $(\text{PSNR}_{\text{ours}} - \text{PSNR}_{\text{SEVA}})$. These show that satellite conditioning improves generations in most cases on nuScenes and Metropolis.}
\vspace{-1em}
\label{tab:psnr_percentile_deltas}
\end{table}

To better quantify our model's performance compared to SEVA (no satellite conditioning), we further analyze the metrics in Table 2 (main paper). For each scene, we compute $\delta = (\text{PSNR}_{\text{ours}} - \text{PSNR}_{\text{SEVA}})$, the improvement in PSNR for our generation vs SEVA. We tabulate the percentile distribution for this metric in Table \ref{tab:psnr_percentile_deltas}. Satellite helps in all cases on Metropolis (lowest value of $\delta$ is positive). On nuScenes, our model is noticeably worse ($>$ 0.5 dB) for only 3.7 \% scenes. We found \emph{2 common challenges} in these cases on nuScenes: a) A large vehicle (truck / tram) occludes most of the input FOV, or b) the camera is stationary (\eg while waiting at intersections). Both these cases give GroundControl very little ground-level context, which it needs to fuse information from satellite view.  

We also performed another experiment on misaligned satellite images. Instead of the correct satellite image, we provided different images with 2 types of misalignments (i) by flipping the satellite image horizontally, and (ii) by using a mismatched satellite image from another random scene. We performed this analysis on Metropolis and report the results in Table \ref{tab:corrupted_satellite}. These corruptions indeed degrade generation quality. However, the degradation is not too significant. This is likely because many scene layouts in driving datasets are similar (even under the flip transformation). Moreover, the input image condition provides a strong prior that the model does not violate even with a misaligned satellite image (it prefers the street image signals, under a conflicting satellite signal). 

\begin{table}
\centering
\setlength{\tabcolsep}{3pt}
\renewcommand*{\arraystretch}{1}
\resizebox{0.7\linewidth}{!}{%
\begin{tabular}{l ccc}
    \cmidrule(lr){2-4}
     & PSNR $\uparrow$  &   DINO $\downarrow$   &   SEG  $\uparrow$ \\
    \midrule
     Flipped satellite & 12.017 & 5.782 & 0.513 \\
     Mismatched satellite & 12.052 & 5.591 & 0.530 \\
    \midrule
    Ours & \textbf{12.809} & \textbf{5.247} & \textbf{0.605} \\
    \midrule
\end{tabular}%
}
\vspace{-1em}
\caption{Impact of corrupted satellite imagery: using flipped satellite images or mismatched satellite images degrades performance.}
\vspace{-1.5em}
\label{tab:corrupted_satellite}
\end{table}

\section{Additional Qualitative Results}
\label{sec:supp_qual_results}

\begin{figure*}
    \centering
    \vspace{-1em}
    \scalebox{0.95}{
    \includegraphics[width=\textwidth]{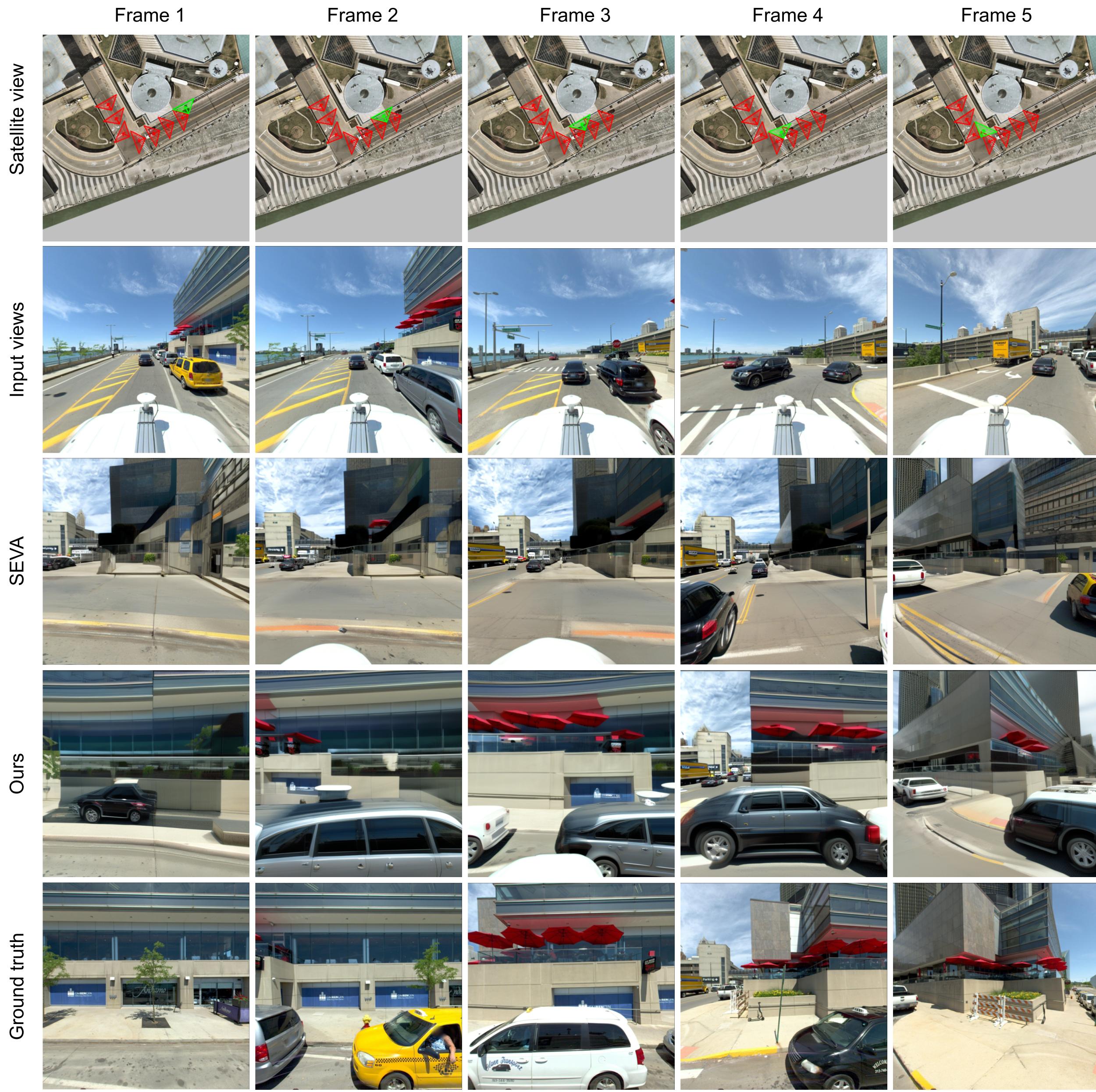}
    }
    \caption{\textbf{Multi-view generations} on the Metropolis dataset: These are multi-frame generations from a single diffusion process with 7 input views (shown in {\color{red}red} on the satellite images). We show target views from 5 consecutive timesteps ({\color{green}green} in the satellite images). Our model's generations are multi-view consistent, and adheres to the spatial layout information in the satellite image.}
    \vspace{-1em}
\label{fig:multi-view-consistency}
\end{figure*}

\begin{figure*}
    \centering
    \newcommand{\imwidthh}{0.195\textwidth}
    
    \setlength{\tabcolsep}{0pt}

    \begin{tabular}{c@{\hskip 2pt}c@{\hskip 2pt}c@{\hskip 2pt}c@{\hskip 2pt}c}
    
        Satellite view ($S$) &
        Example input &
        Ground truth &
        SEVA & 
        Ours
        \\
        
        \includegraphics[width=\imwidthh]{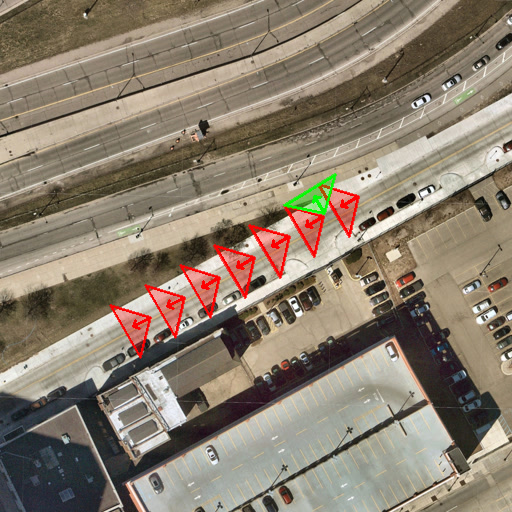} &
        \includegraphics[width=\imwidthh]{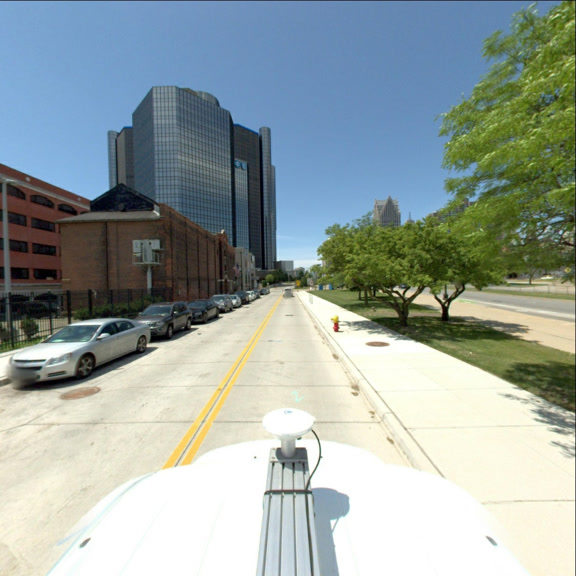} &
        \includegraphics[width=\imwidthh]{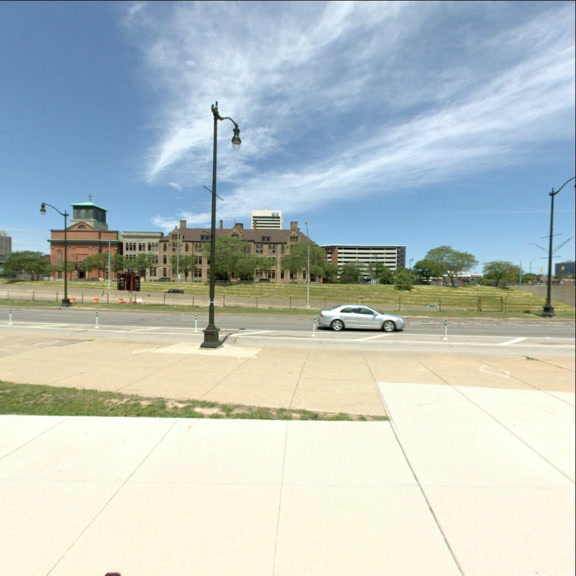} &
        \includegraphics[width=\imwidthh]{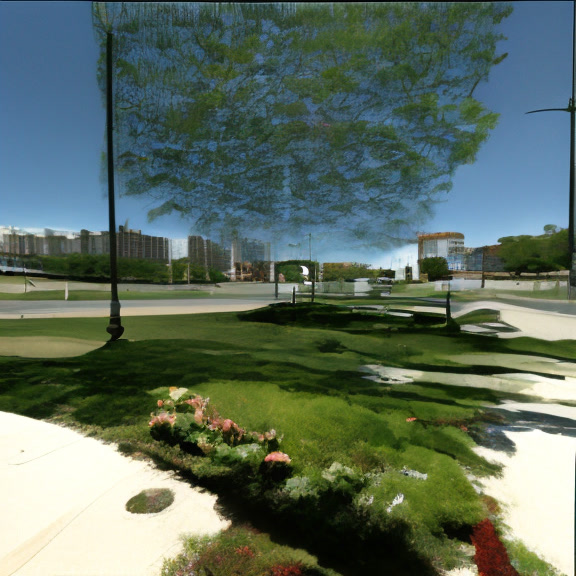} &
        \includegraphics[width=\imwidthh]{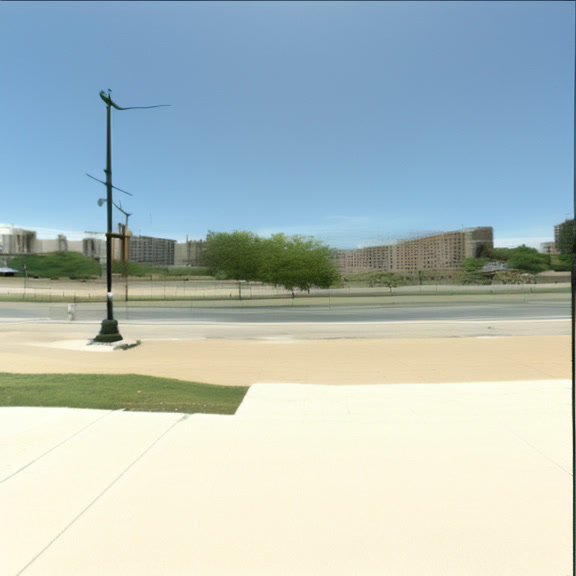} \\

        \includegraphics[width=\imwidthh]{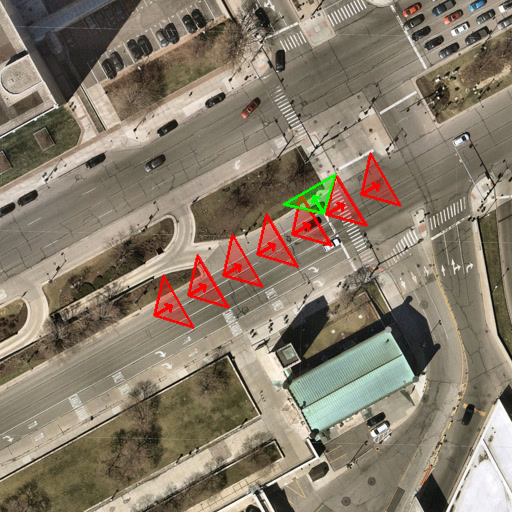} &
        \includegraphics[width=\imwidthh]{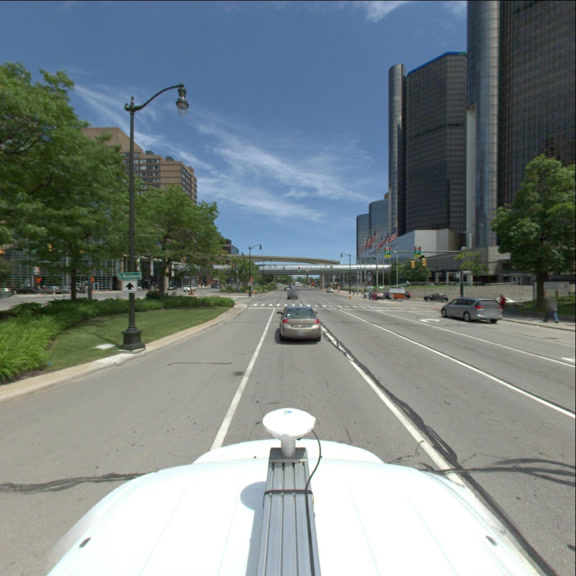} &
        \includegraphics[width=\imwidthh]{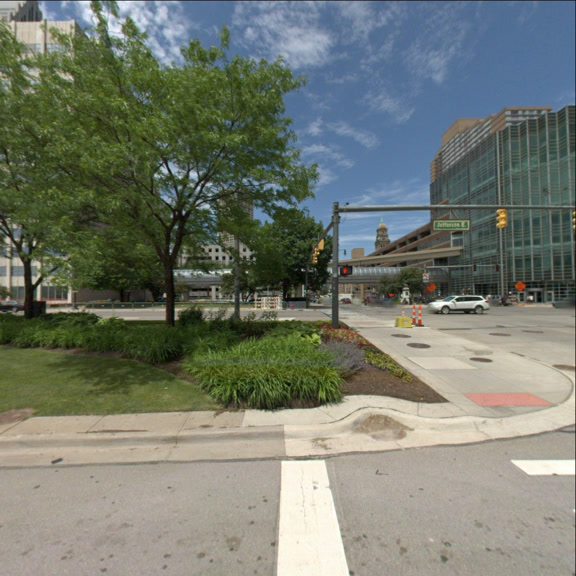} &
        \includegraphics[width=\imwidthh]{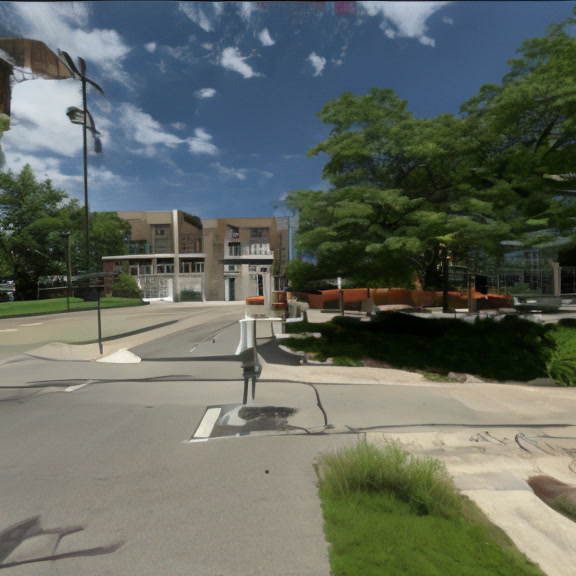} &
        \includegraphics[width=\imwidthh]{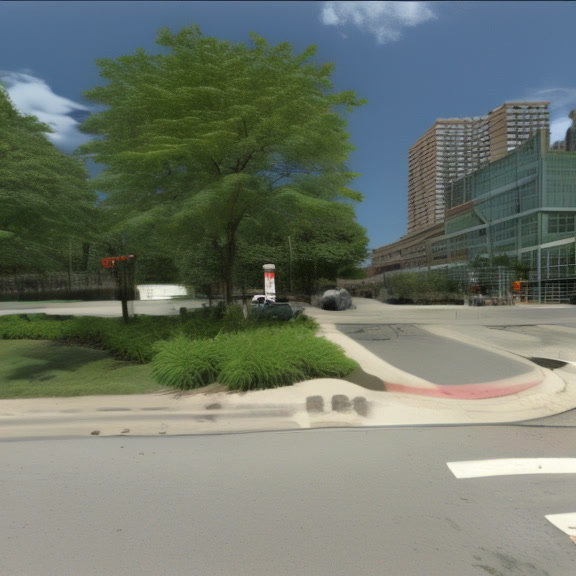} \\

        \includegraphics[width=\imwidthh]{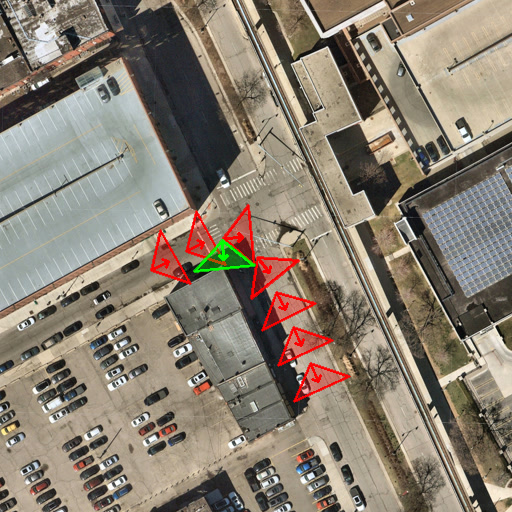} &
        \includegraphics[width=\imwidthh]{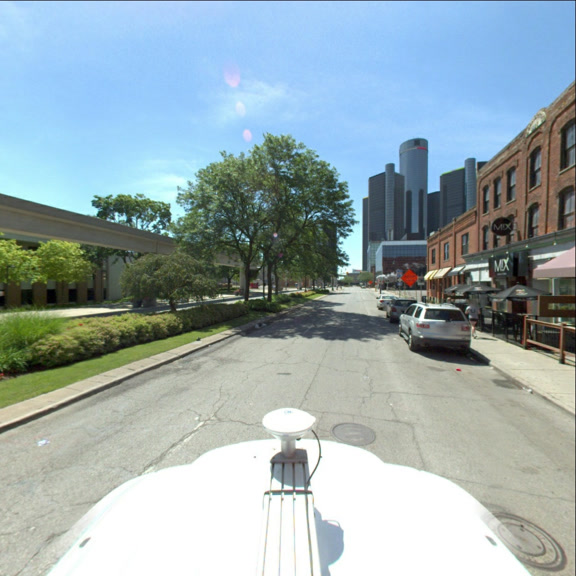} &
        \includegraphics[width=\imwidthh]{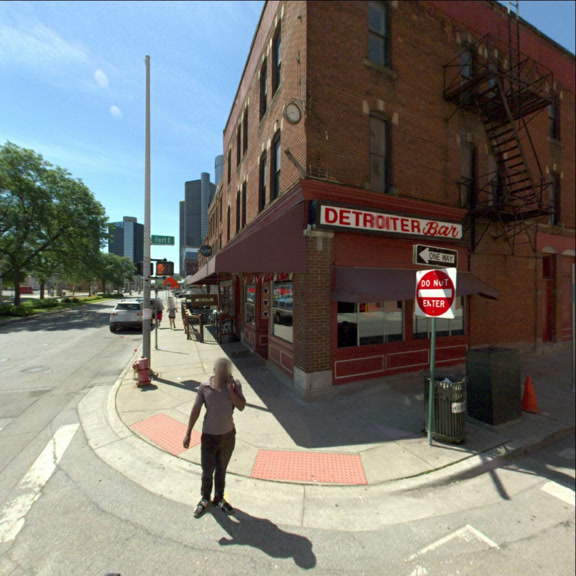} &
        \includegraphics[width=\imwidthh]{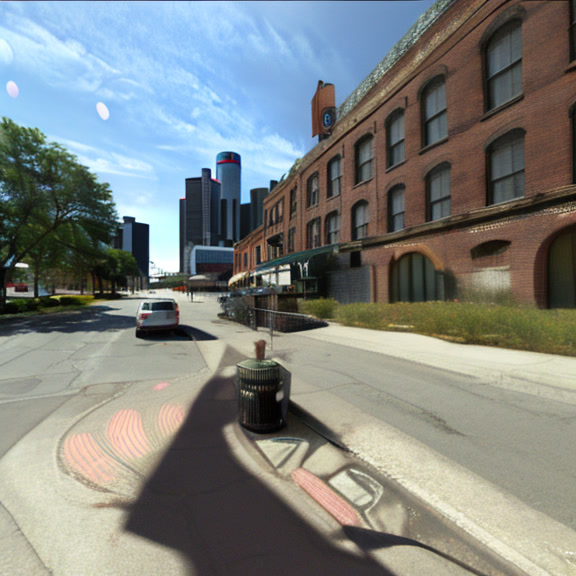} &
        \includegraphics[width=\imwidthh]{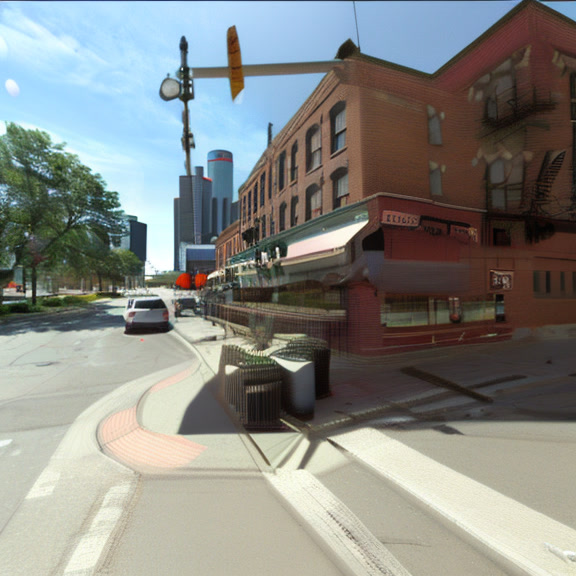} \\

        \includegraphics[width=\imwidthh]{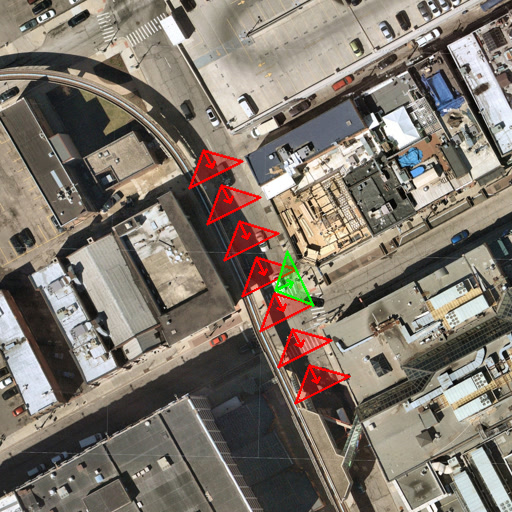} &
        \includegraphics[width=\imwidthh]{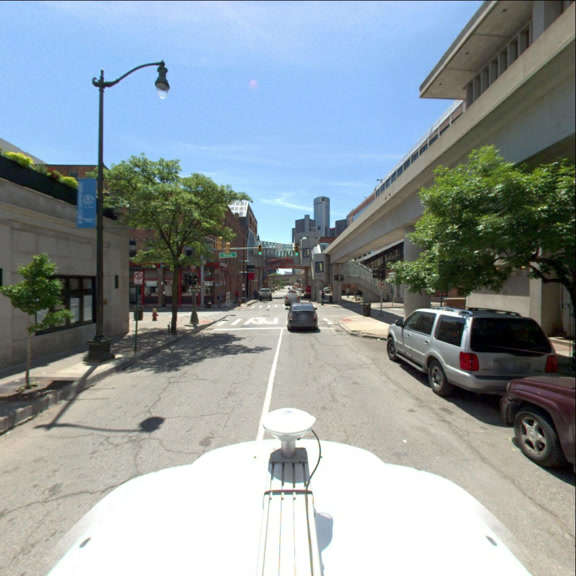} &
        \includegraphics[width=\imwidthh]{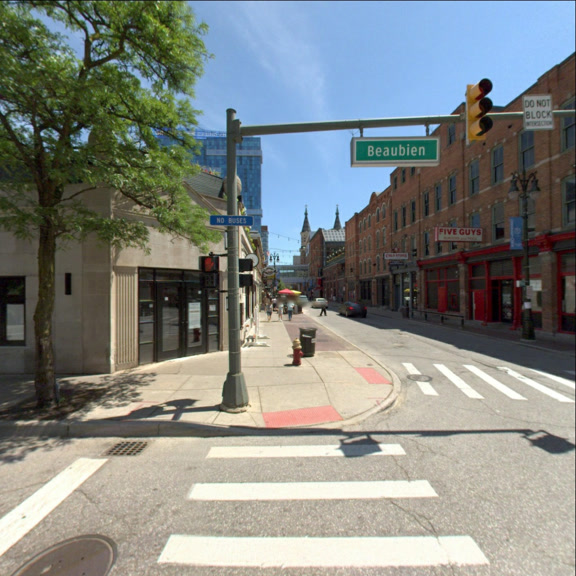} &
        \includegraphics[width=\imwidthh]{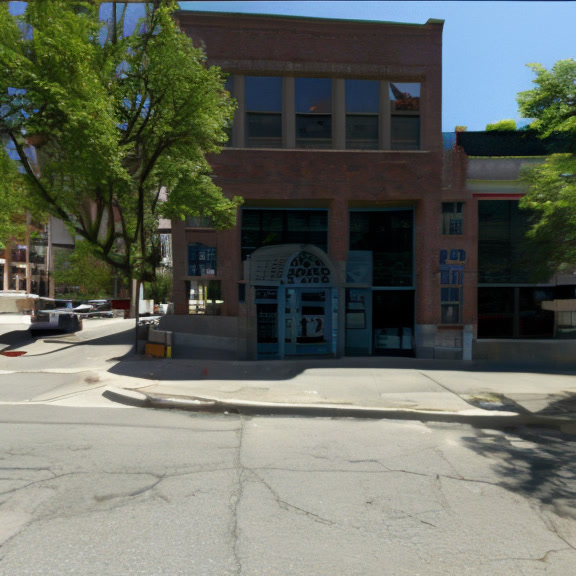} &
        \includegraphics[width=\imwidthh]{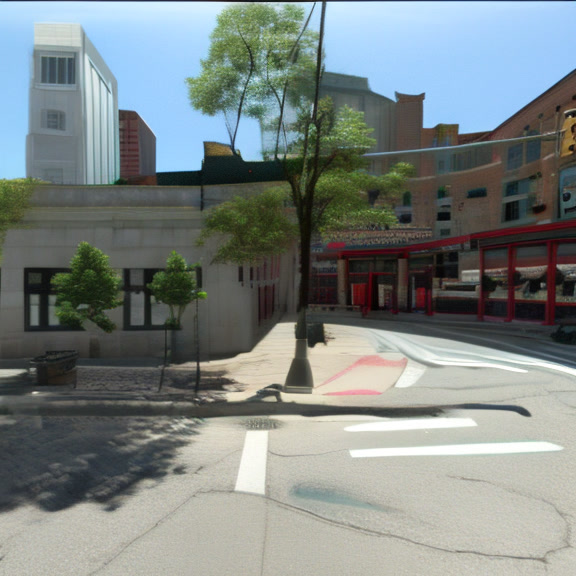} \\

        \includegraphics[width=\imwidthh]{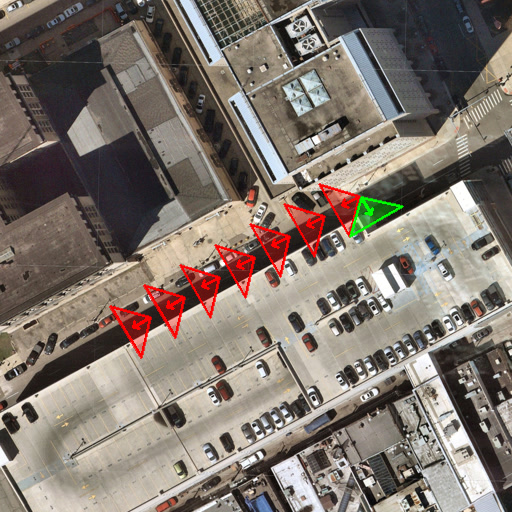} &
        \includegraphics[width=\imwidthh]{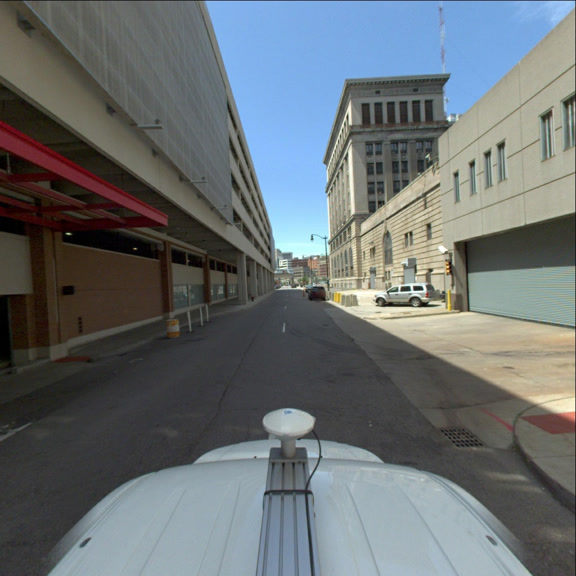} &
        \includegraphics[width=\imwidthh]{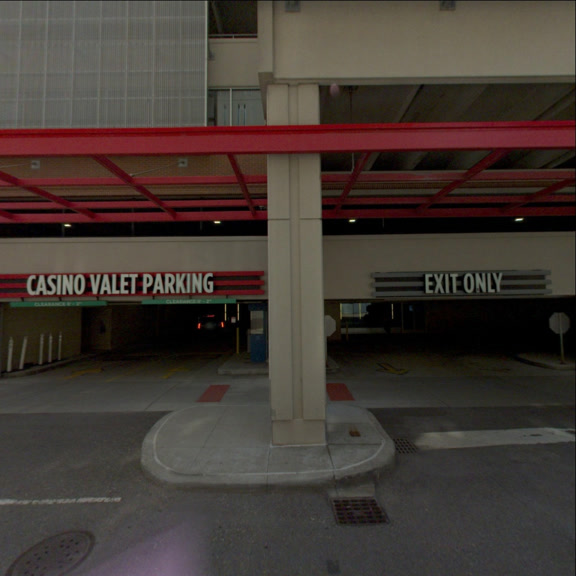} &
        \includegraphics[width=\imwidthh]{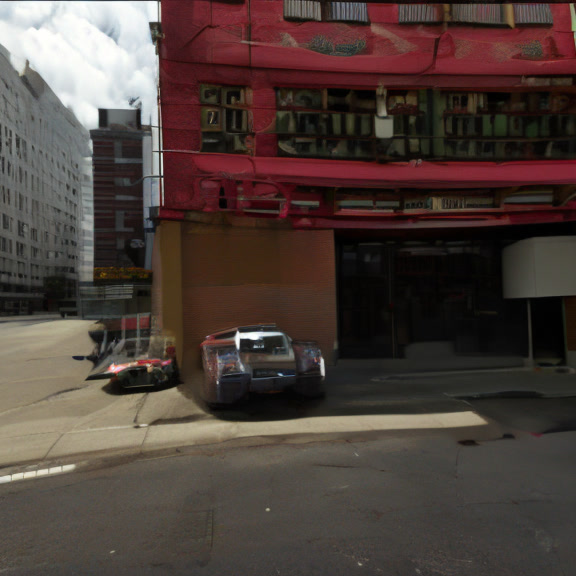} &
        \includegraphics[width=\imwidthh]{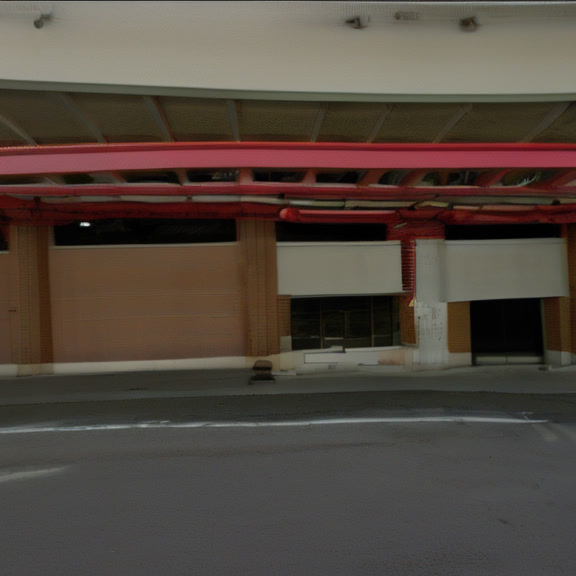} \\

        \includegraphics[width=\imwidthh]{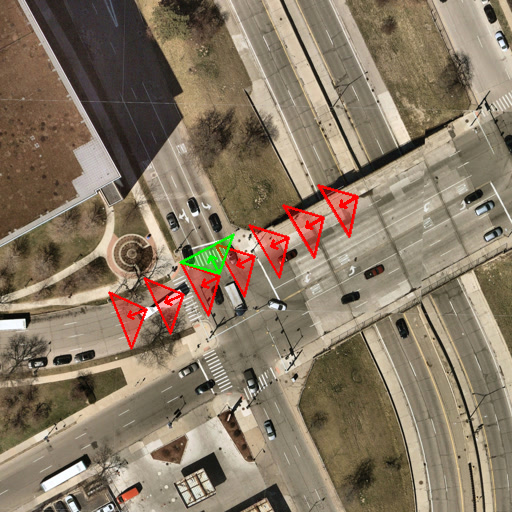} &
        \includegraphics[width=\imwidthh]{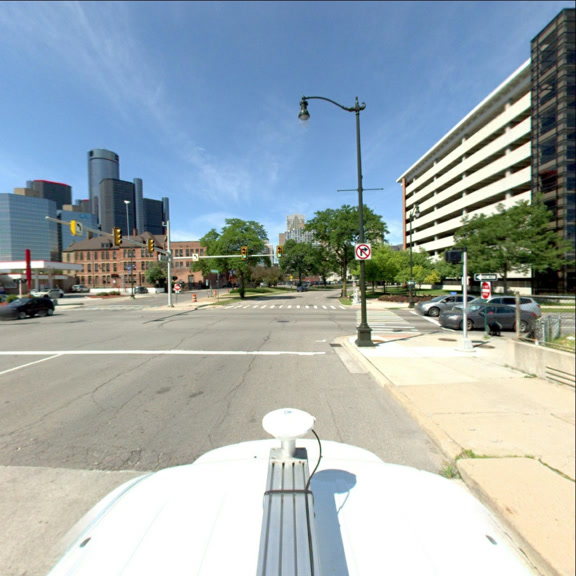} &
        \includegraphics[width=\imwidthh]{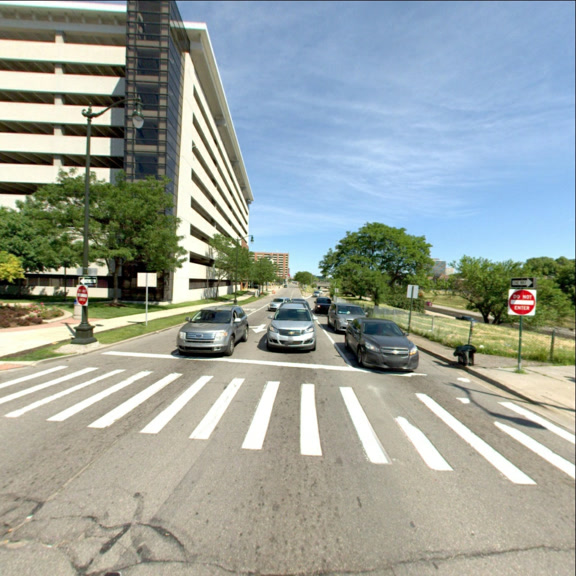} &
        \includegraphics[width=\imwidthh]{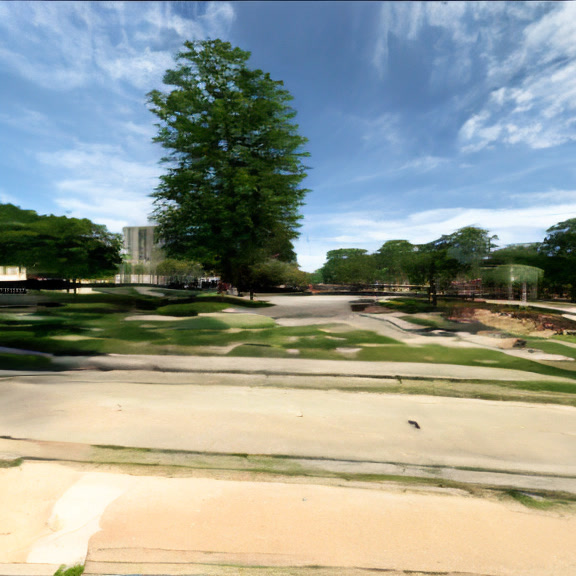} &
        \includegraphics[width=\imwidthh]{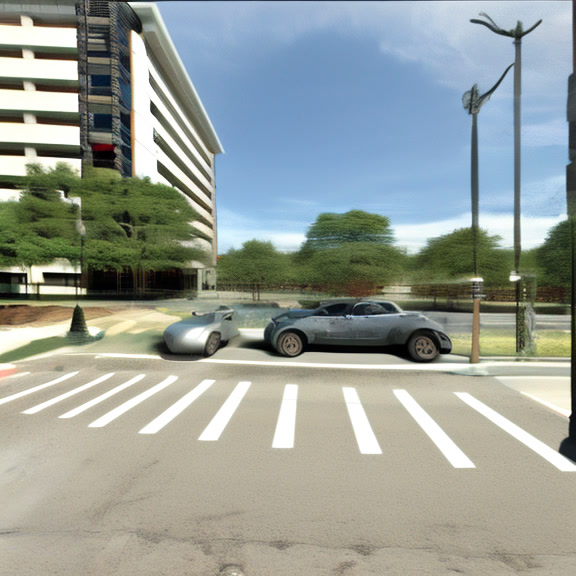} \\        
    \end{tabular}
    \caption{\textbf{Qualitative results on the Metropolis dataset.} We show results of our satellite-conditioned model on sequences of length 21 frames. This includes 7 input frames (from the front camera) and 14 target frames (from the left and right cameras at the corresponding timestamps). The SEVA model without satellite conditioning underperforms in this sparse input-view regime which has little overlap with the target views. }
    \label{fig:supp_metropolis_576}
\end{figure*}

\begin{figure*}
    \centering
    \newcommand{\imwidthh}{0.195\textwidth}
    
    \setlength{\tabcolsep}{0pt}

    \begin{tabular}{c@{\hskip 2pt}c@{\hskip 2pt}c@{\hskip 2pt}c@{\hskip 2pt}c}
    
        Satellite ($S$) &
        Example input &
        Ground truth &
        SEVA & 
        Ours
        \\
        
        \includegraphics[width=\imwidthh]{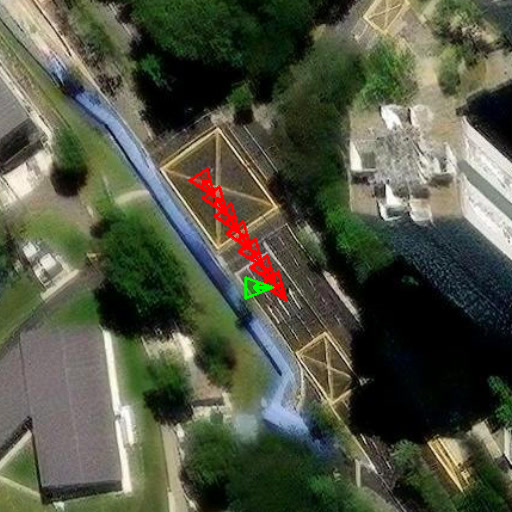} &
        \includegraphics[width=\imwidthh]{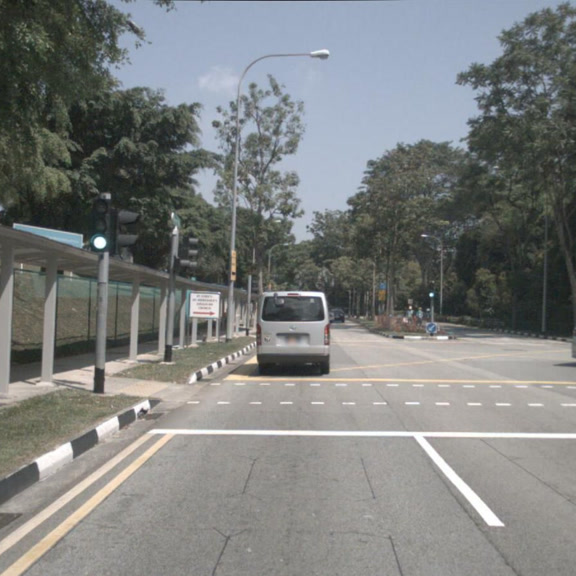} &
        \includegraphics[width=\imwidthh]{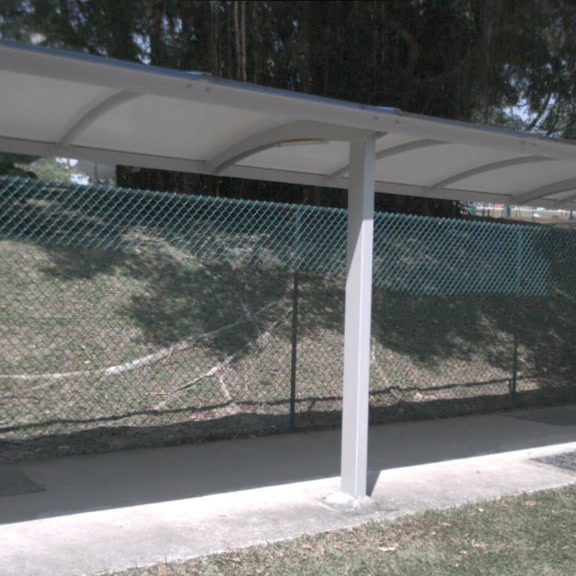} &
        \includegraphics[width=\imwidthh]{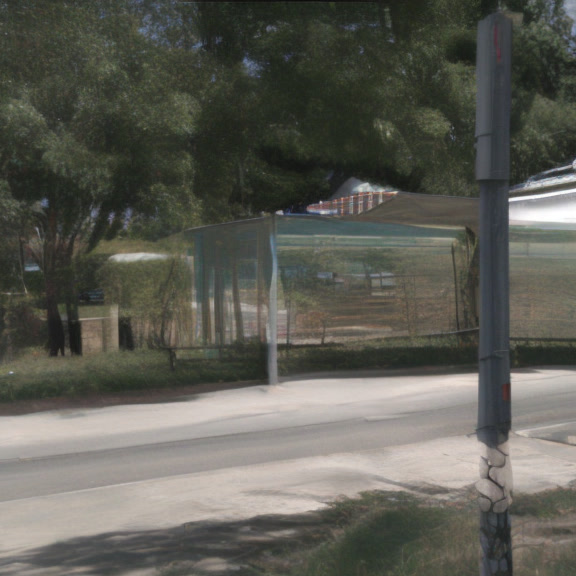} &
        \includegraphics[width=\imwidthh]{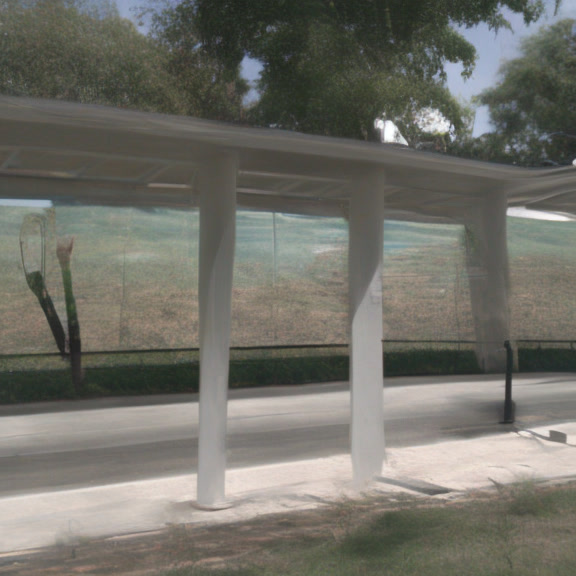} \\
        
        \includegraphics[width=\imwidthh]{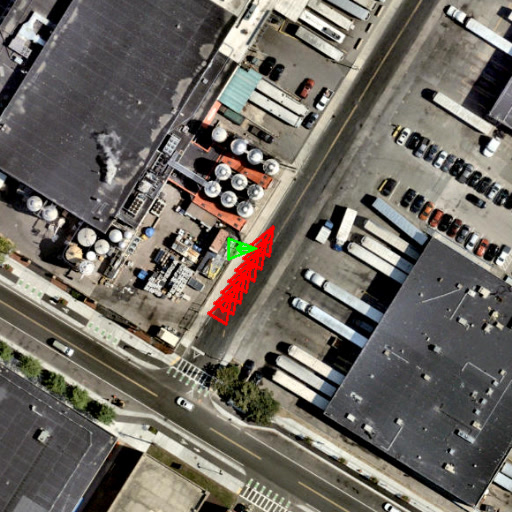} &
        \includegraphics[width=\imwidthh]{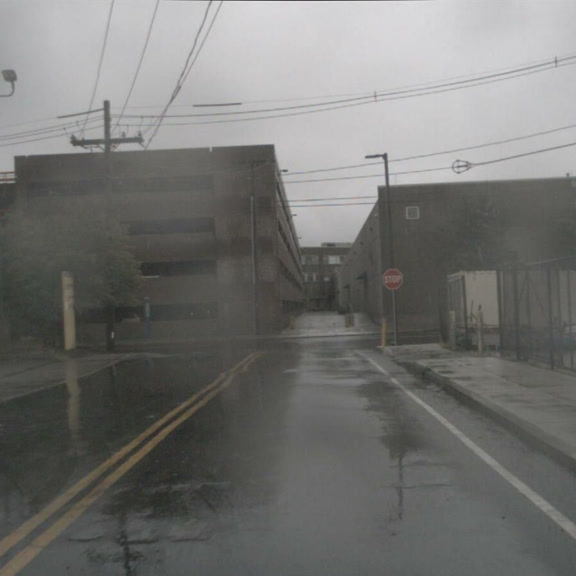} &
        \includegraphics[width=\imwidthh]{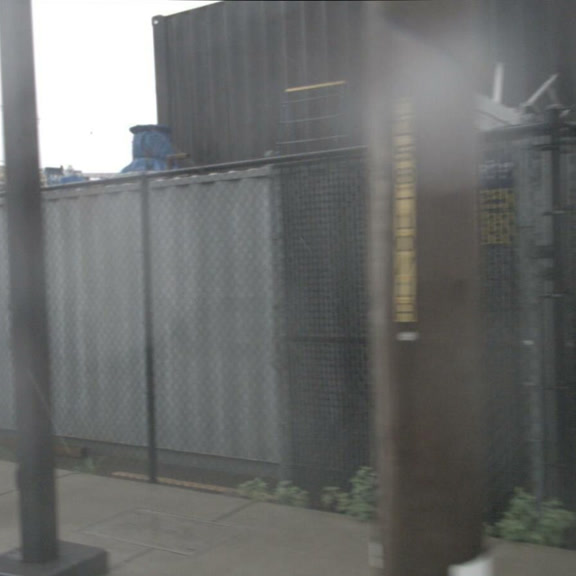} &
        \includegraphics[width=\imwidthh]{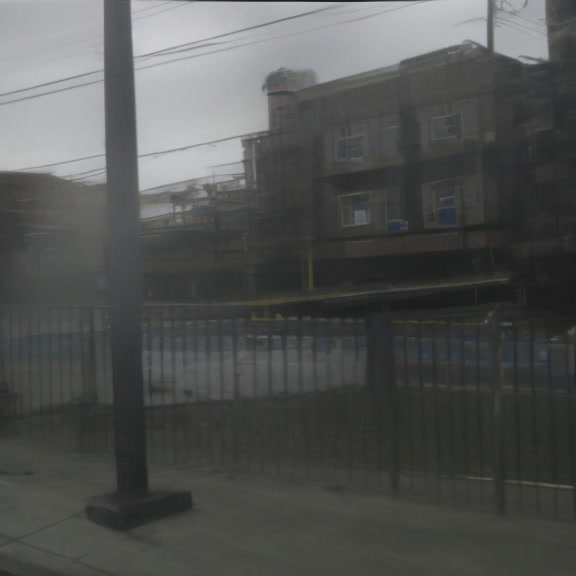} &
        \includegraphics[width=\imwidthh]{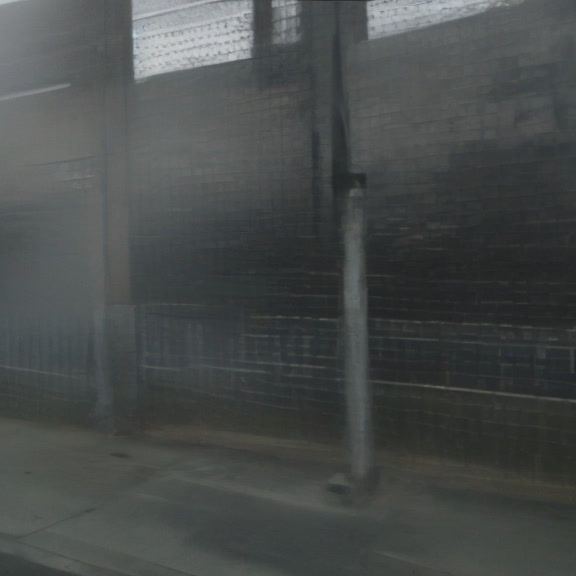} \\

        \includegraphics[width=\imwidthh]{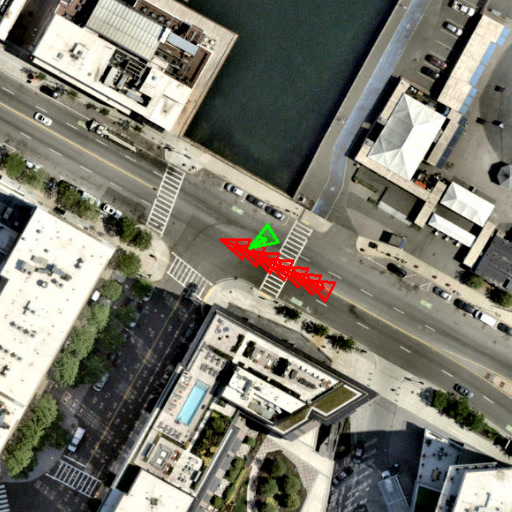} &
        \includegraphics[width=\imwidthh]{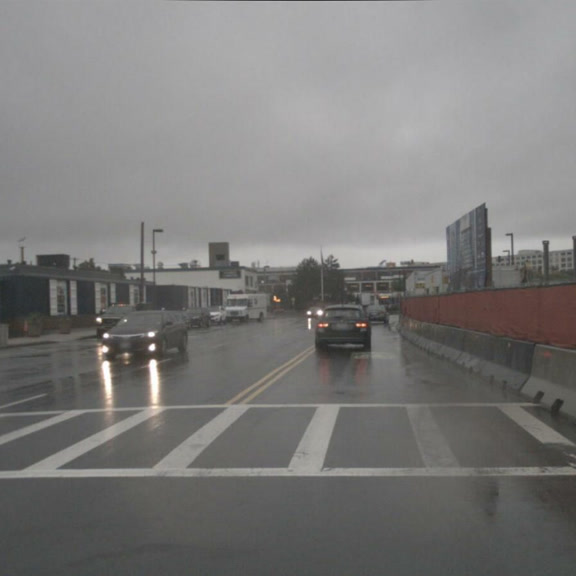} &
        \includegraphics[width=\imwidthh]{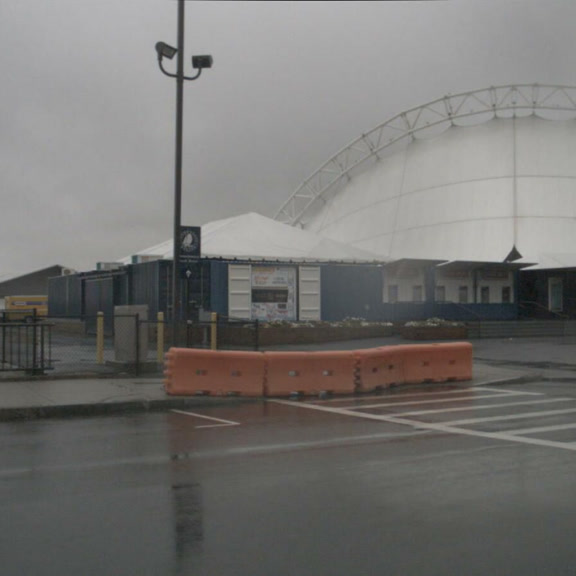} &
        \includegraphics[width=\imwidthh]{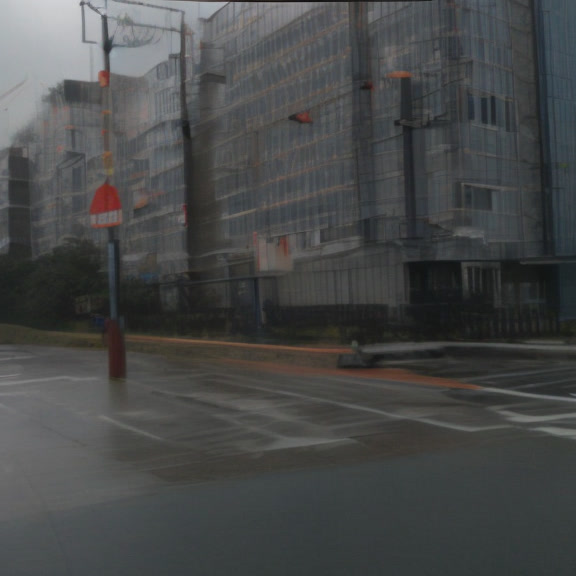} &
        \includegraphics[width=\imwidthh]{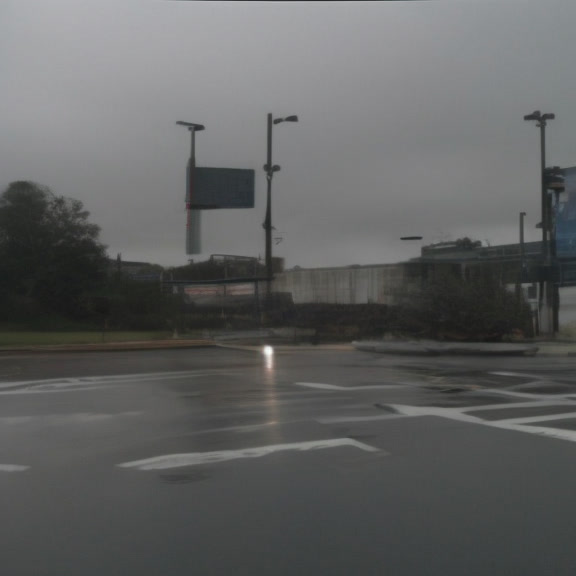} \\

        \includegraphics[width=\imwidthh]{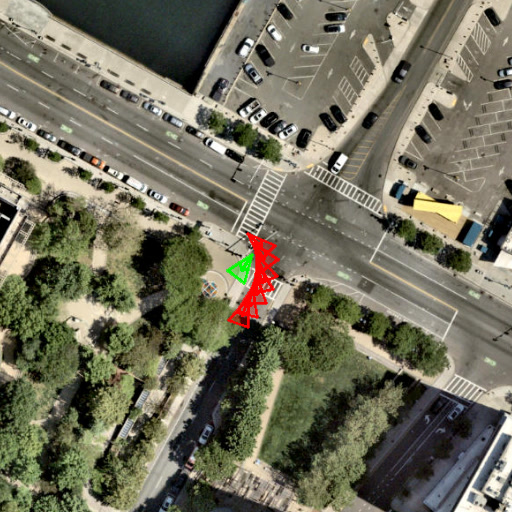} &
        \includegraphics[width=\imwidthh]{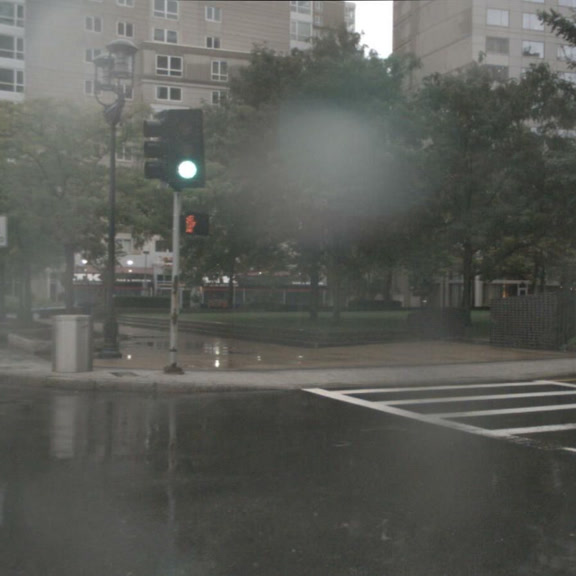} &
        \includegraphics[width=\imwidthh]{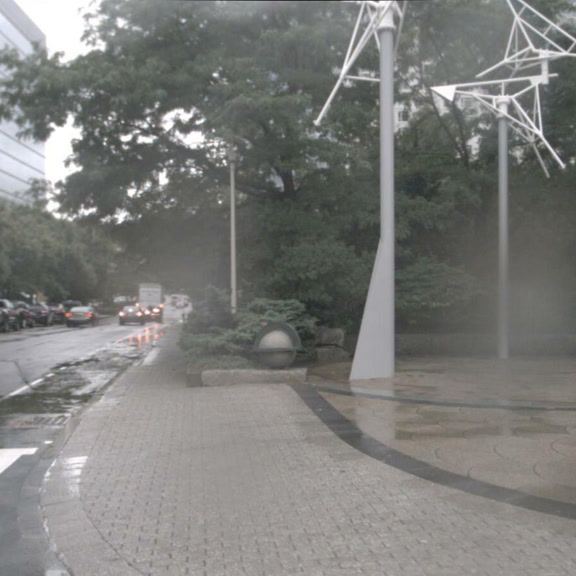} &
        \includegraphics[width=\imwidthh]{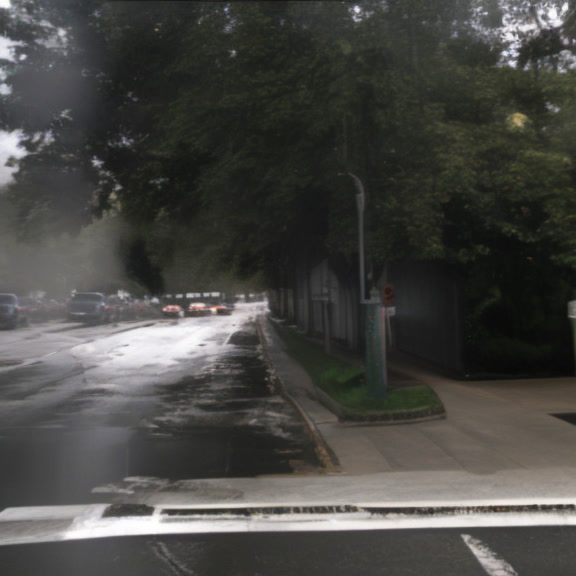} &
        \includegraphics[width=\imwidthh]{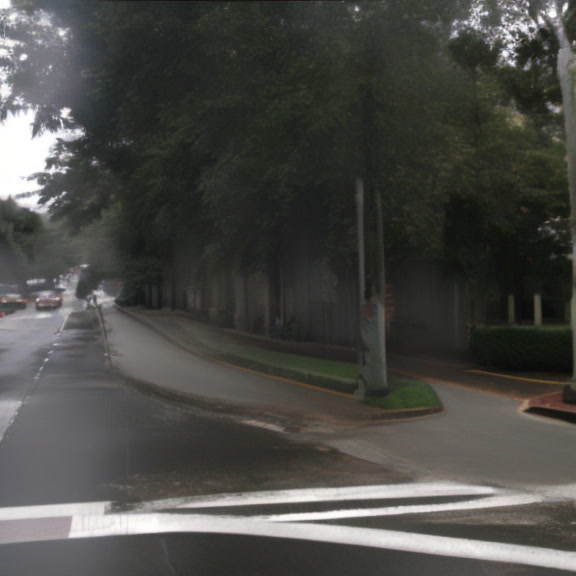} \\

    \end{tabular}
\caption{\textbf{Qualitative results on nuScenes dataset} These show cases with very little overlap between input views (\red{red} in satellite view) and the target views (\green{green} in satellite view). Our model’s generations adhere more closely to the layout of the scene, since these are conditioned on the satellite view.}

    \label{fig:supp_nuscenes}
\end{figure*}

\begin{figure*}[t]
  \centering

  \satpanel{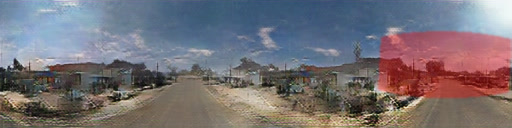}
  \vspace{6pt}

  \satpanel{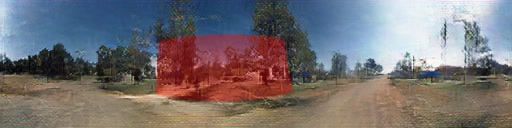}

  \caption{Visualization of Sat2Density results on nuScenes, complementing Fig.~\ref{fig:results_in_distribution}. The predicted panorama is 128$\times$512. Each row shows (left) the panorama prediction and below it a highlight showing the region used to generate the cropped image, and (right) the ground-truth target image.}
  \label{fig:sat2density}
\end{figure*}

\begin{figure*}
    \centering
    \vspace{-1em}
    \scalebox{0.95}{
    \includegraphics[width=0.8\textwidth]{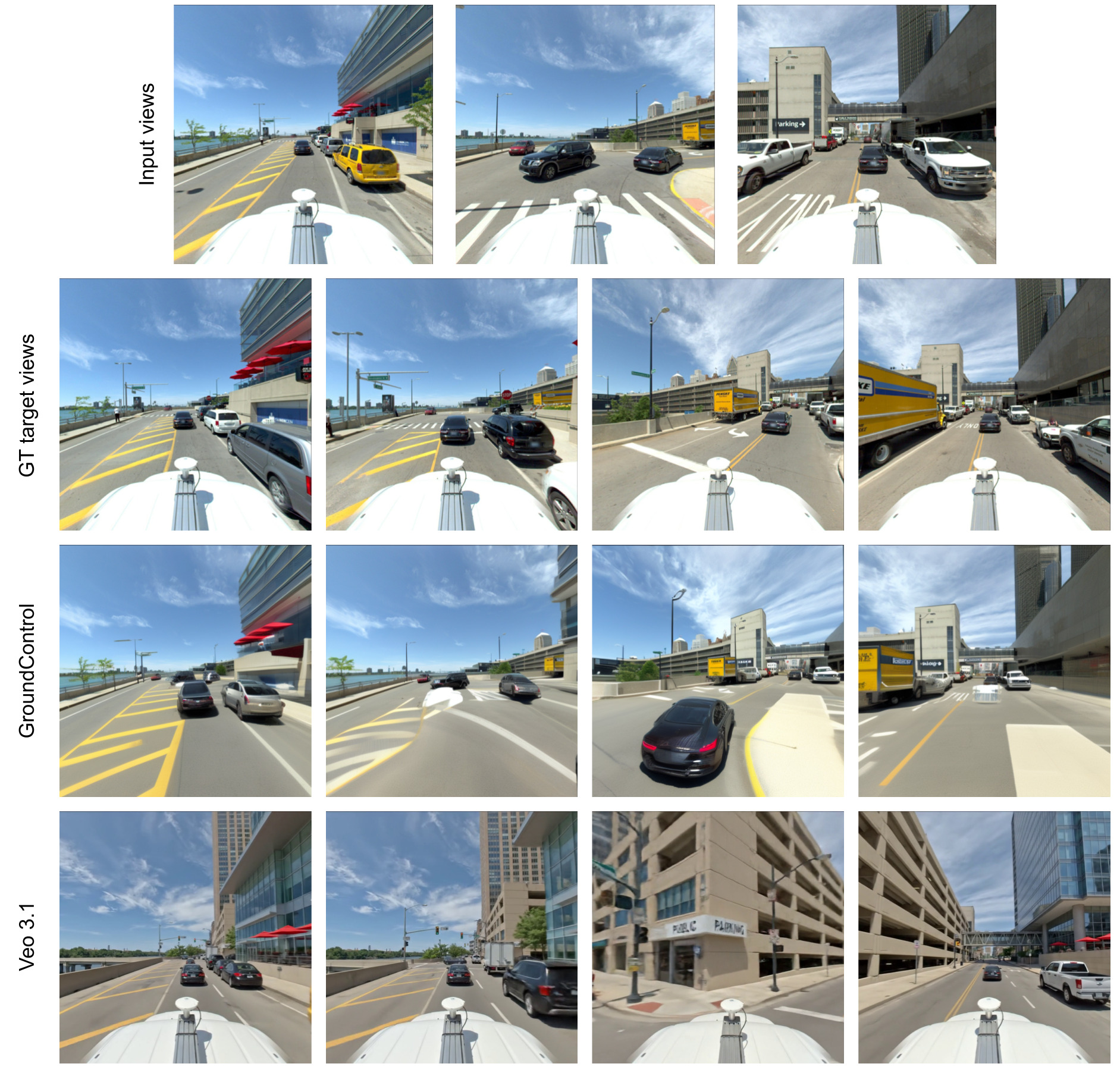}
    }
    \caption{\textbf{Veo 3.1 qualitative results} on the Metropolis dataset with 3 input frames. We generate 2 separate videos with Veo to use the 3 conditioning frames, but GroundControl generates the target views in a single pass. Veo produces highly photorealistic videos, since its model size and training data are larger by orders of magnitude. Unlike our model, it can also handle dynamic objects (including the hood of the car the camera is mounted on). However, its camera trajectory may not necessarily follow the required target poses. It can also frequently hallucinate scene content in incorrect locations (as highlighted in the image). }
    \vspace{-1em}
\label{fig:veo-comparison}
\end{figure*}

Figures \ref{fig:multi-view-consistency} and \ref{fig:supp_metropolis_576} show additional qualitative results on the Metropolis dataset. Figure \ref{fig:multi-view-consistency} particularly highlights the multi-view consistency of our model, and the satellite adherence in comparison to SEVA.  Figure \ref{fig:supp_nuscenes} shows our results on some challenging scenes from nuScenes where the input images have reduced visibility due to weather conditions. For the Sat2Density crops used in the main paper, we include the full generated panoramas in Figure \ref{fig:sat2density}. We also include a qualitative comparison to state-of-the-art proprietary video model Veo in Figure \ref{fig:veo-comparison}. 


\subsection{Multiple Noise Samples}

\begin{figure*}
    \centering
    \newcommand{\imwidthh}{0.16\textwidth}
    
    \setlength{\tabcolsep}{0pt}

    \begin{tabular}{c@{\hskip 4pt}c@{\hskip 2pt}c@{\hskip 2pt}c@{\hskip 2pt}c@{\hskip 2pt}c@{\hskip 2pt}c}
    
        &
        Satellite &
        Input &
        Noise-1 &
        Noise-2 & 
        Noise-3 & 
        GT
        \\

        \rotatebox{90}{Ours} &
        \includegraphics[width=\imwidthh]{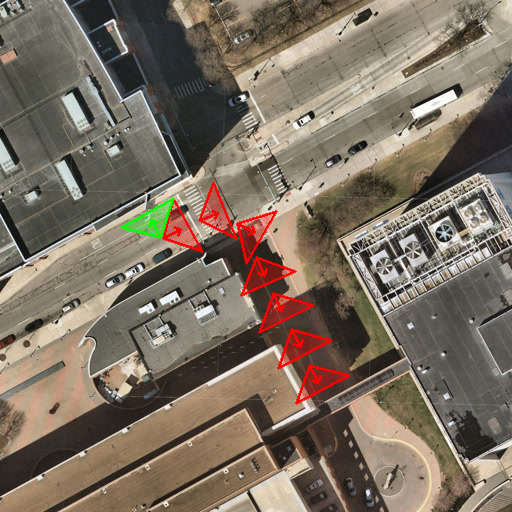} &
        \includegraphics[width=\imwidthh]{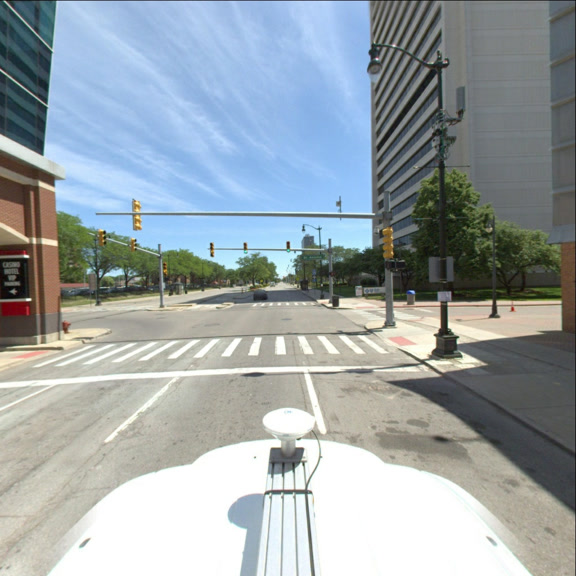} &
        \includegraphics[width=\imwidthh]{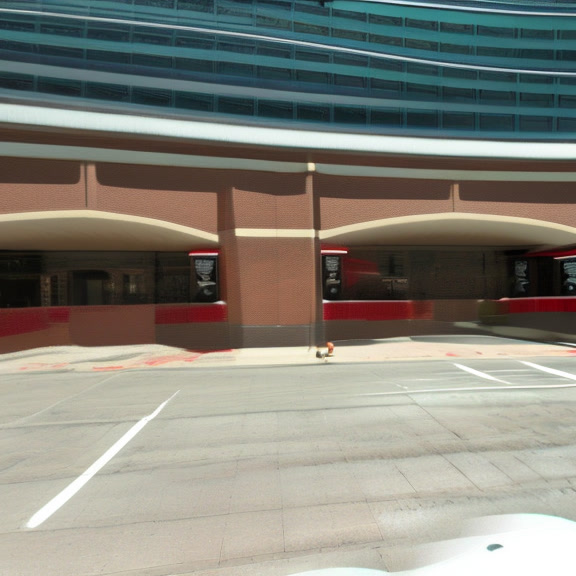} &
        \includegraphics[width=\imwidthh]{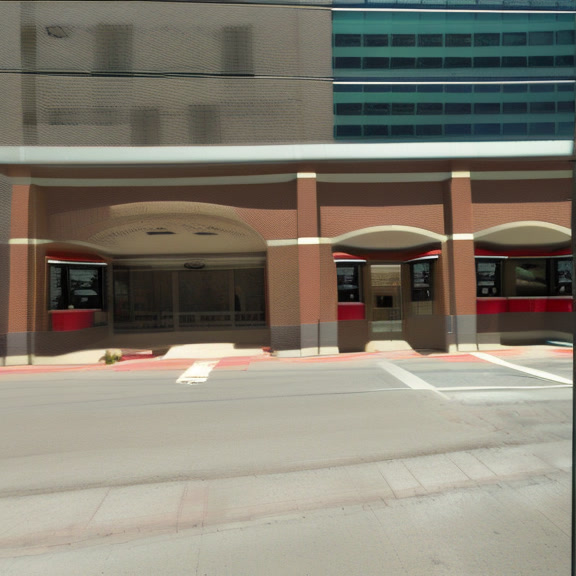} &
        \includegraphics[width=\imwidthh]{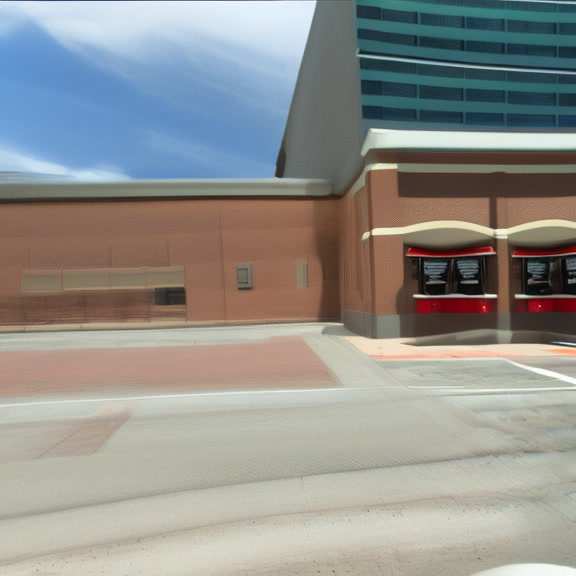} &
        \includegraphics[width=\imwidthh]{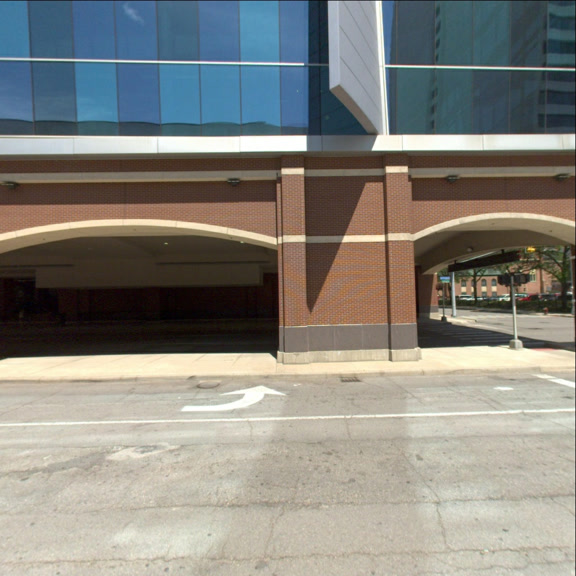} \\

        \rotatebox{90}{SEVA} & 
        & & 
        \includegraphics[width=\imwidthh]{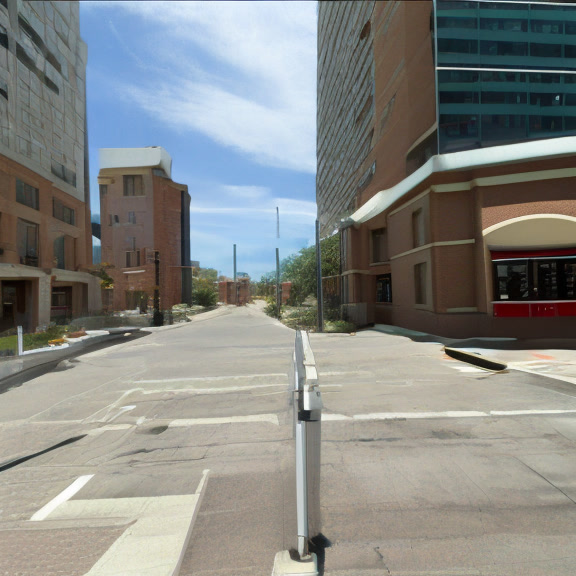} &
        \includegraphics[width=\imwidthh]{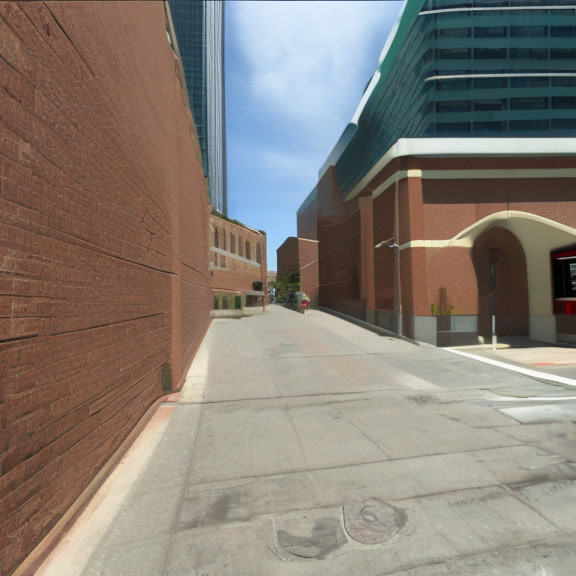} &
        \includegraphics[width=\imwidthh]{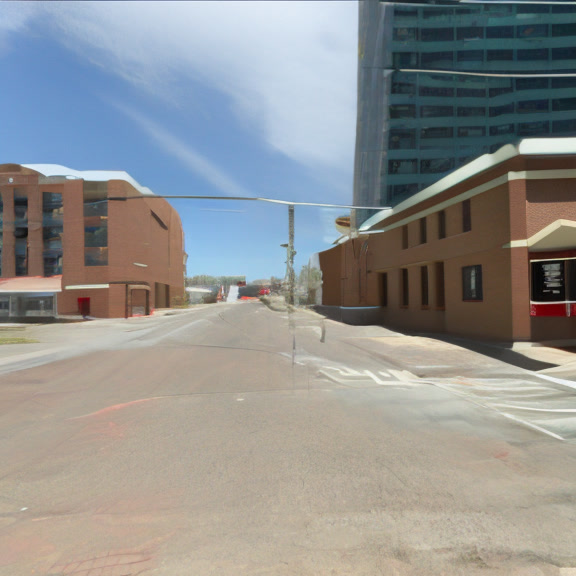} &
        \includegraphics[width=\imwidthh]{figures/multiseed_ie89/gt_00.jpg} \\

        \midrule 

        \rotatebox{90}{Ours} &
        \includegraphics[width=\imwidthh]{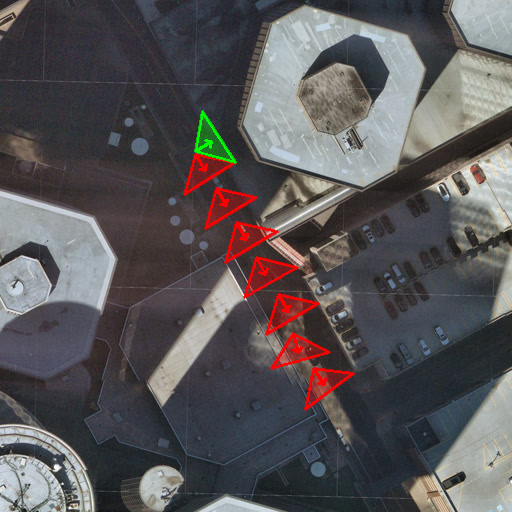} &
        \includegraphics[width=\imwidthh]{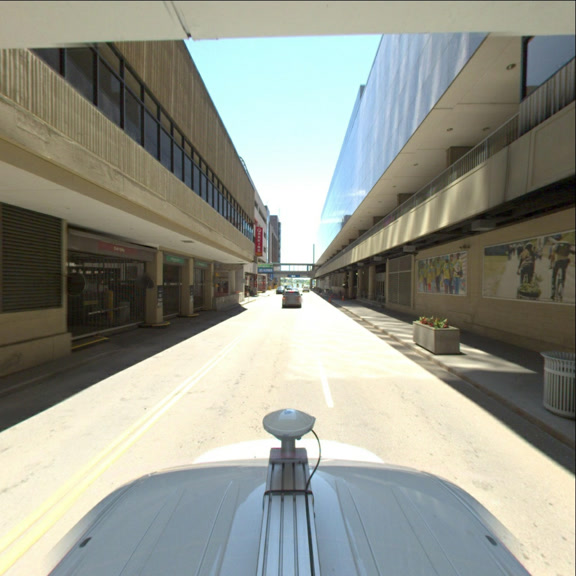} &
        \includegraphics[width=\imwidthh]{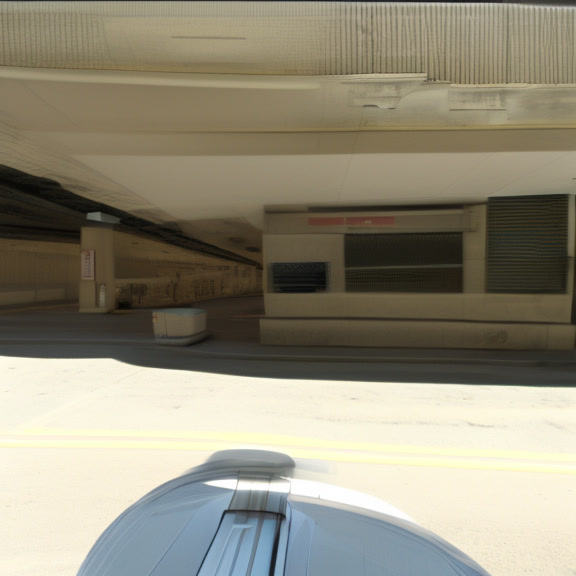} &
        \includegraphics[width=\imwidthh]{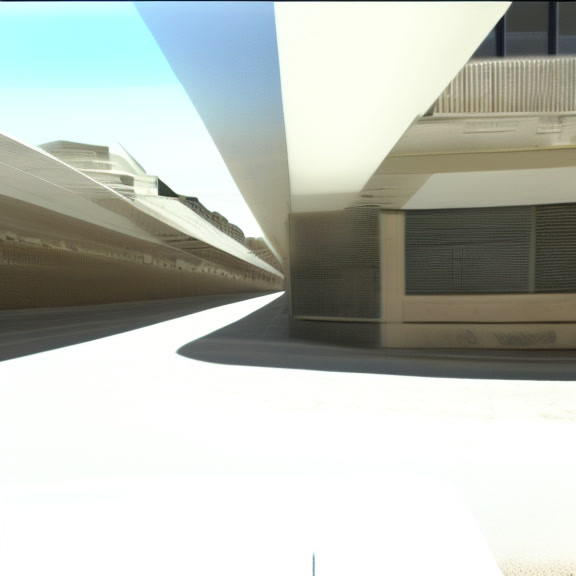} &
        \includegraphics[width=\imwidthh]{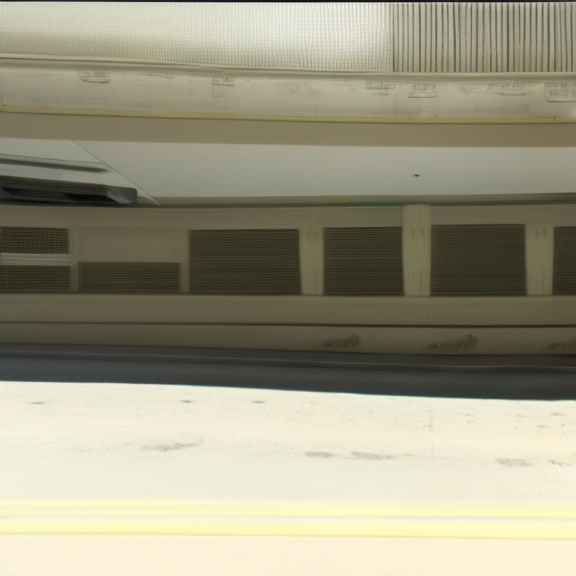} &
        \includegraphics[width=\imwidthh]{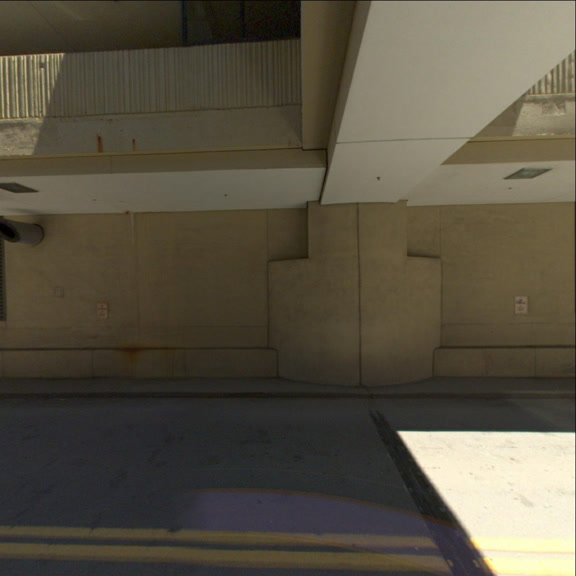} \\

        \rotatebox{90}{SEVA} & 
        & & 
        \includegraphics[width=\imwidthh]{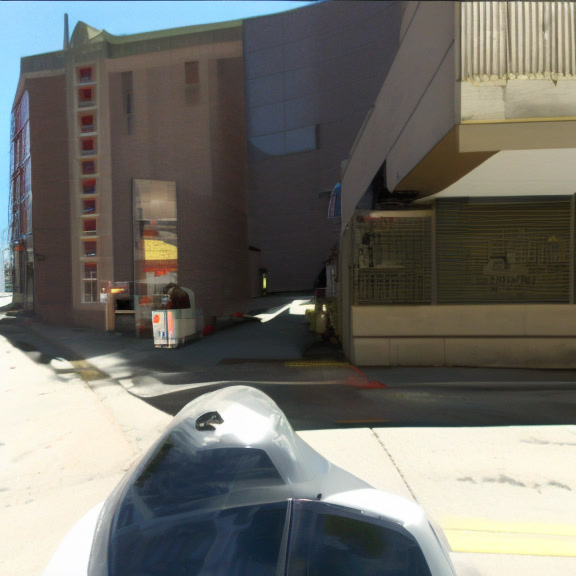} &
        \includegraphics[width=\imwidthh]{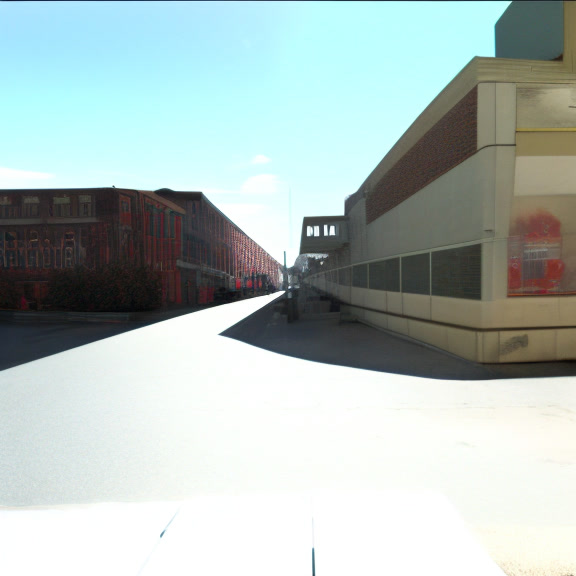} &
        \includegraphics[width=\imwidthh]{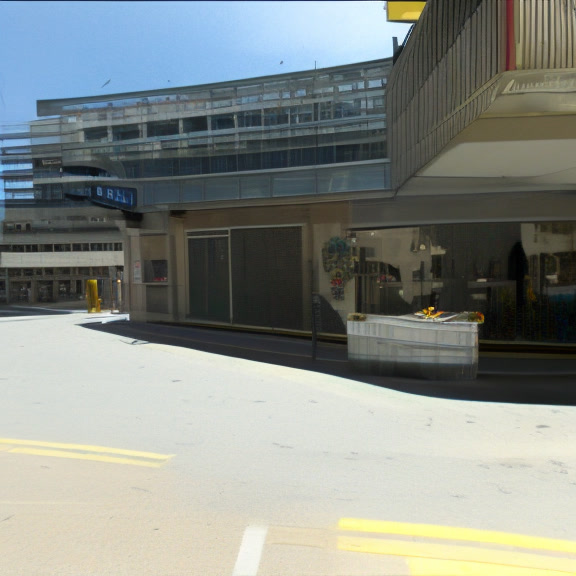} &
        \includegraphics[width=\imwidthh]{figures/multiseed_zun2/gt_00.jpg} \\

    \end{tabular}
    \caption{\textbf{Qualitative results with multiple noise samples}: We compare variations in the generations from SEVA to that of our model, when starting the denoising process with different noise samples. In regions that are non-overlapping with the input view, SEVA's generations vary more, and are inconsistent with the scene layout. Since our generations are conditioned on satellite imagery, they show less variance in the generated layout, while still generating diverse samples.}

    \label{fig:supp_multiseed}
\end{figure*}

We provide qualitative comparisons of multiple generations from SEVA and our model in Figure \ref{fig:supp_multiseed}. This allows us to compare the diversity of samples from both models. We find that the satellite conditioning indeed reduces the variance of layouts generated by our model in comparison to SEVA. Our model's generations still retain diversity in the appearance or height of buildings as this cannot be observed from the satellite view. 

\subsection{Effect of Dual CFG}

\begin{figure*}
    \centering
    \includegraphics[width=0.99\textwidth]{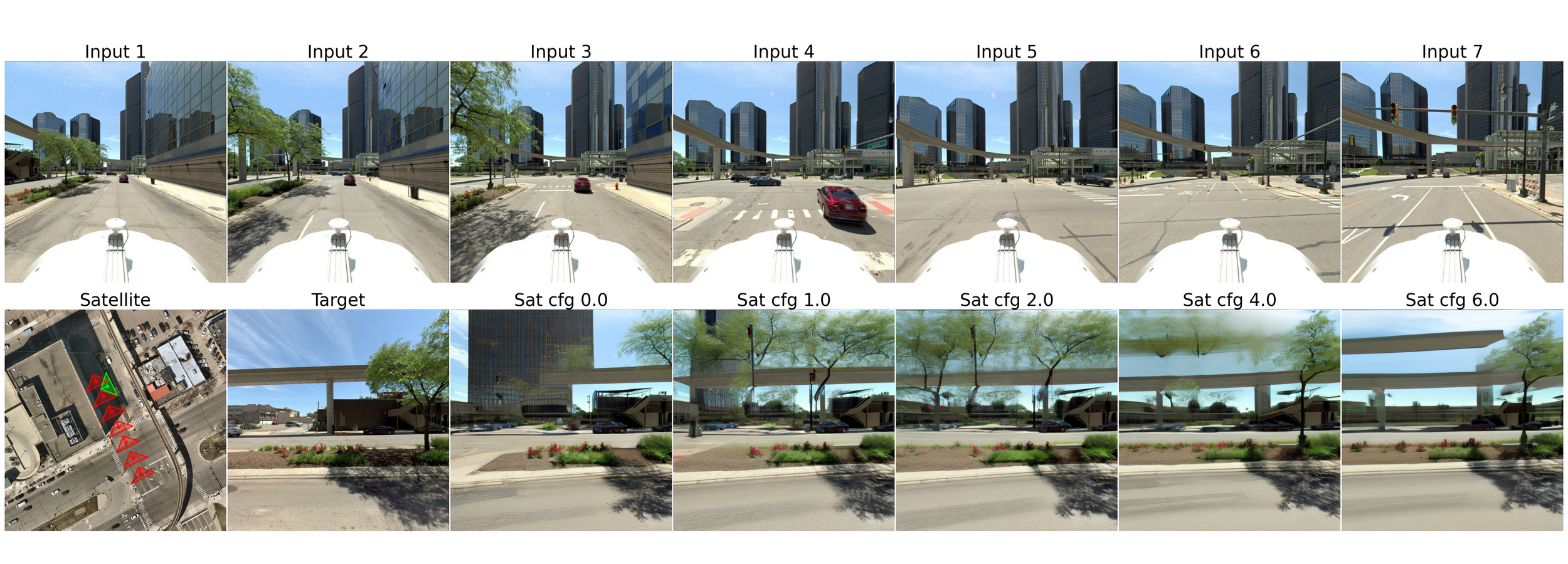}
    \vspace{-5pt}
    \caption{
        \textbf{Varying satellite CFG on a scene from Metropolis} Increasing the satellite CFG strengthens the adherence to the satellite image, but higher values trade off against image quality. The input views on the top row correspond to the red frustums in the satellite image, and the predicted views shown in the second row correspond to the green frustum.
    }
    \vspace{-5pt}
    \label{fig:dual-cfg}
\end{figure*}

As discussed in the main paper, we use a dual-CFG scheme which allows us to separately control the strength of the conditioning on the input views and on the satellite image. In Fig.~\ref{fig:dual-cfg} we show the impact of varying the satellite CFG value while keeping the image CFG value fixed (something made possible by our dual-CFG approach). As we increase the satellite CFG, the model correctly extends the monorail and the lane divider, and removes the building in the background, in accordance with the satellite image. However at very high values, the quality of the generated image begins to decrease. Our dual-CFG approach gives control over this tradeoff. (Note that at satellite CFG 0 the output is independent of the content of the satellite image, because the satellite guidance term from eq. 1 of the main submission gets multiplied by zero.)

\subsection{Ablations}

\renewcommand{\imwidth}{0.16\textwidth}

\begin{figure*}
    \centering
    \begin{tabular}{@{\hskip 2pt}c@{\hskip 2pt}c@{\hskip 2pt}c|c@{\hskip 2pt}c@{\hskip 2pt}c@{\hskip 2pt}c}
        Satellite ($S$) &
        Input &
        GT &
        No sat. & 
        No ray CA & 
        Ours
        \\

        \includegraphics[width=\imwidth]{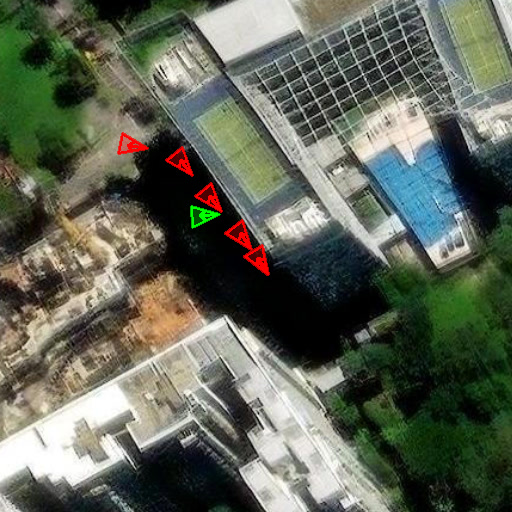} &
        \includegraphics[width=\imwidth]{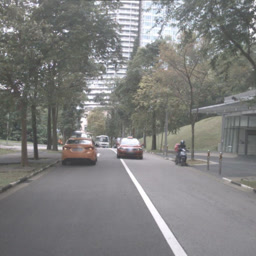} &
        \includegraphics[width=\imwidth]{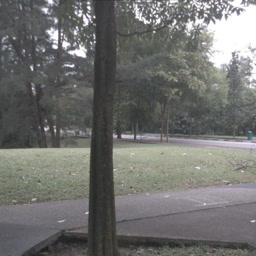} &
        \includegraphics[width=\imwidth]{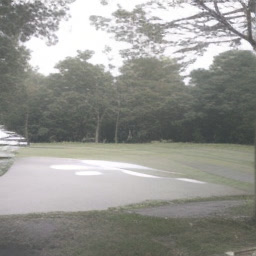} &
        \includegraphics[width=\imwidth]{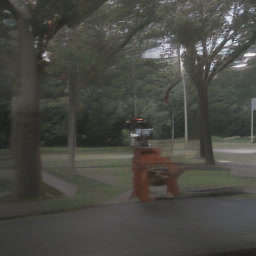} &
        \includegraphics[width=\imwidth]{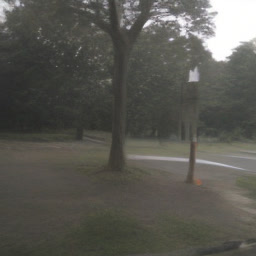} \\

        \includegraphics[width=\imwidth]{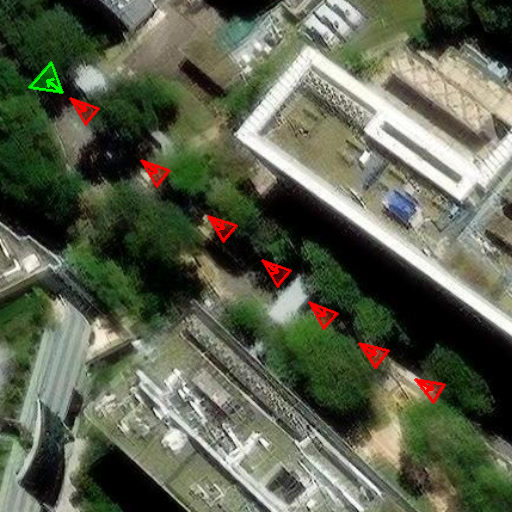} &
        \includegraphics[width=\imwidth]{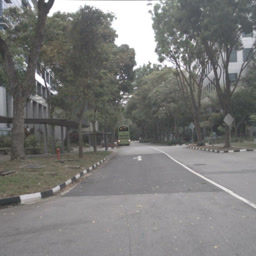} &
        \includegraphics[width=\imwidth]{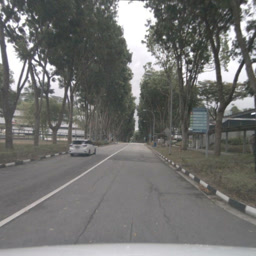} &
        \includegraphics[width=\imwidth]{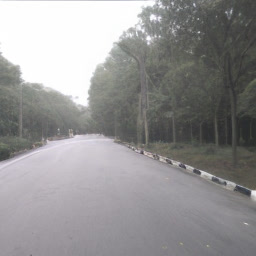} &
        \includegraphics[width=\imwidth]{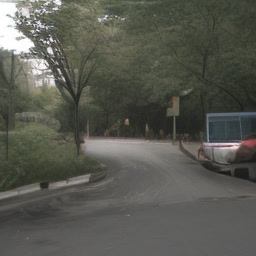} &
        \includegraphics[width=\imwidth]{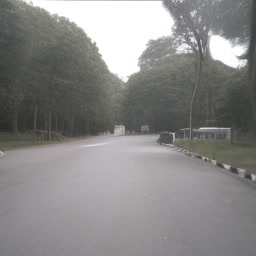} \\
        
        \includegraphics[width=\imwidth]{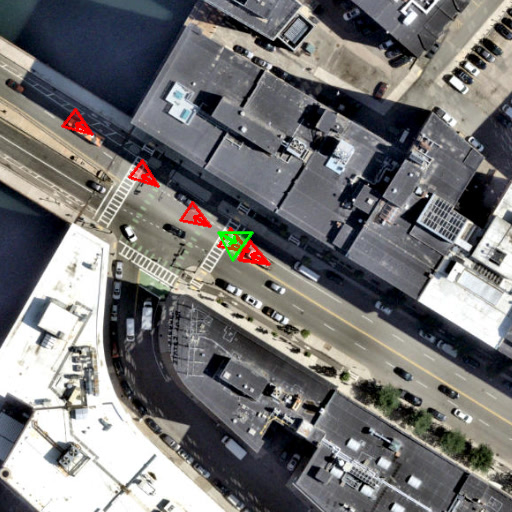} &
        \includegraphics[width=\imwidth]{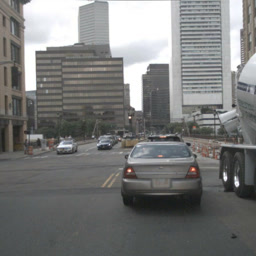} &
        \includegraphics[width=\imwidth]{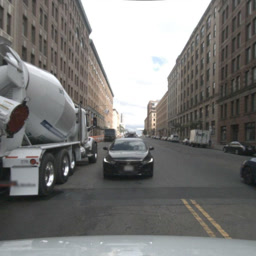} &
        \includegraphics[width=\imwidth]{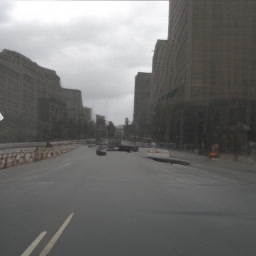} &
        \includegraphics[width=\imwidth]{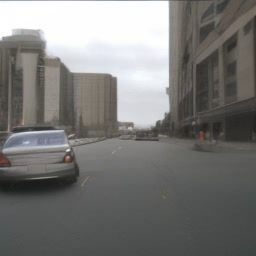} &
        \includegraphics[width=\imwidth]{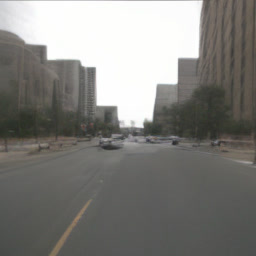} \\
        
        \includegraphics[width=\imwidth]{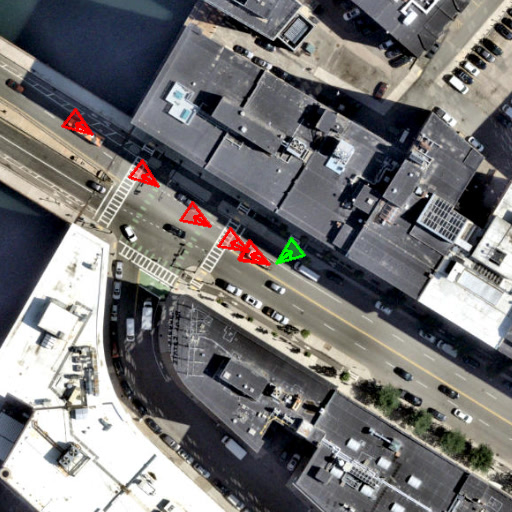} &
        \includegraphics[width=\imwidth]{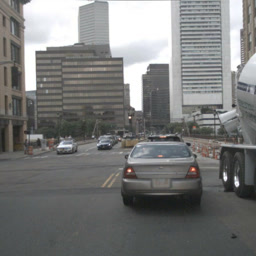} &
        \includegraphics[width=\imwidth]{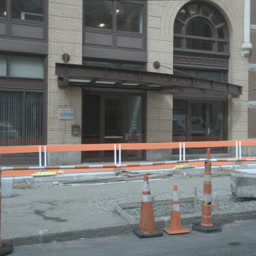} &
        \includegraphics[width=\imwidth]{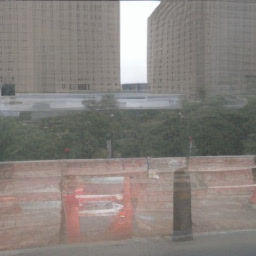} &
        \includegraphics[width=\imwidth]{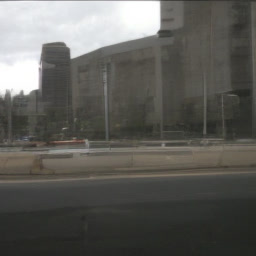} &
        \includegraphics[width=\imwidth]{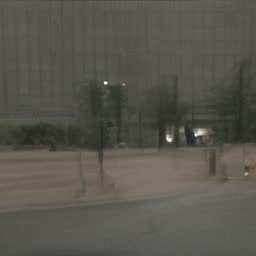} \\
        
    \end{tabular}
    \caption{\textbf{Qualitative ablation results on NuScenes.} Here we compare three alternatives: a model with no satellite input (No sat\.); a model with satellite input but no ray-based cross-attention (no ray CA), and finally our model. Columns on the left show the satellite view (S) with the input views in \red{red} and target view in \green{green}, a representative input view (input) and the ground truth (GT)  corresponding to this target view.  We note that our model tends to have better fidelity to the ground truth layout.}
    \label{fig:ray-ablations}
\end{figure*}

Figure~\ref{fig:ray-ablations} shows our ablations where we compare our model with variants which don't have satellite imagery as input (\textbf{No satellite input}) and a model which uses satellite as input but doesn't include ray-guided cross-attention (\textbf{No ray CA}).
We see that our model has better fidelity to the overall structure of the scene when compared with the ground truth, vs.~both the alternative formulations.

\subsection{Visualizations of Attention Maps for Satellite Rays}
\begin{figure*}
    \centering
    \includegraphics[width=0.12\textwidth]{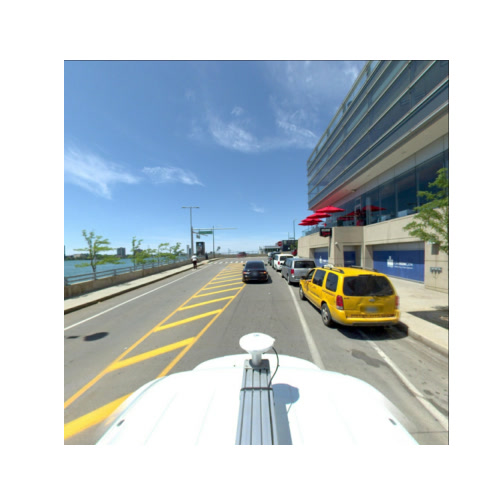}
    \includegraphics[width=0.12\textwidth]{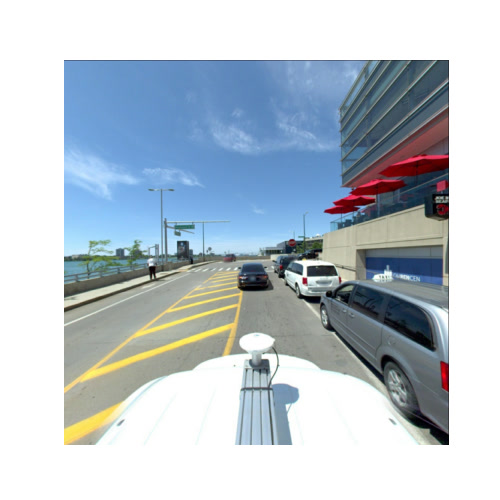}
    \includegraphics[width=0.12\textwidth]{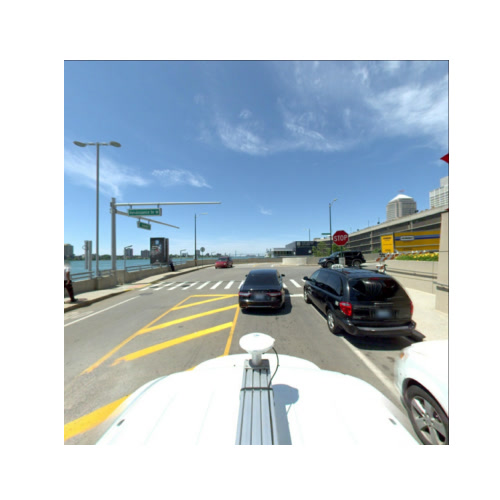}
    \includegraphics[width=0.12\textwidth]{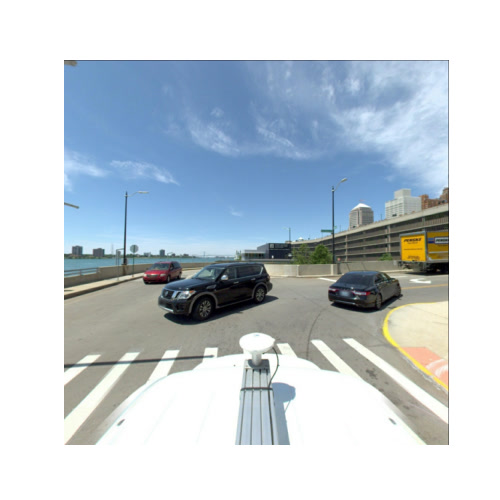}
    \includegraphics[width=0.12\textwidth]{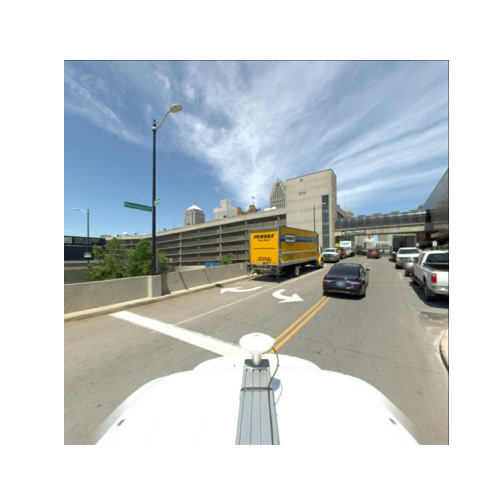}
    \includegraphics[width=0.12\textwidth]{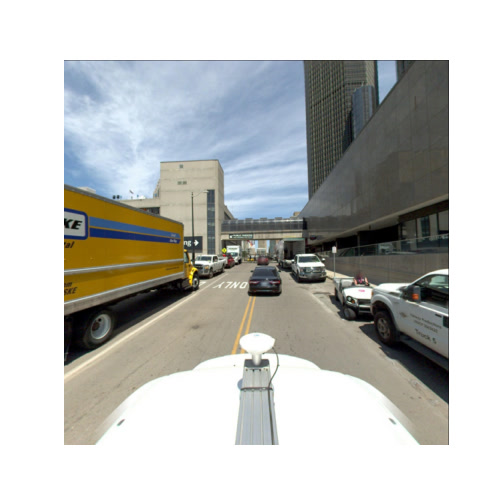}
    \includegraphics[width=0.12\textwidth]{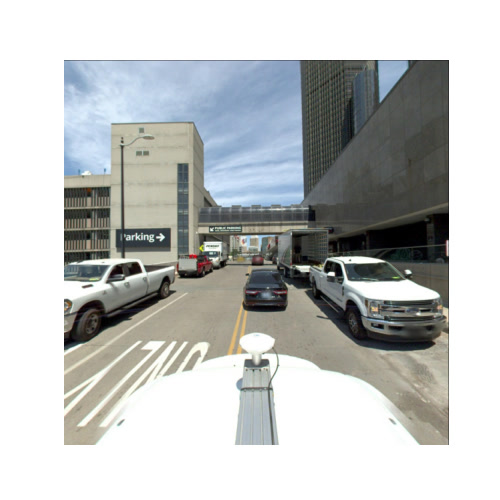}
    \\
    \includegraphics[width=0.2\textwidth]{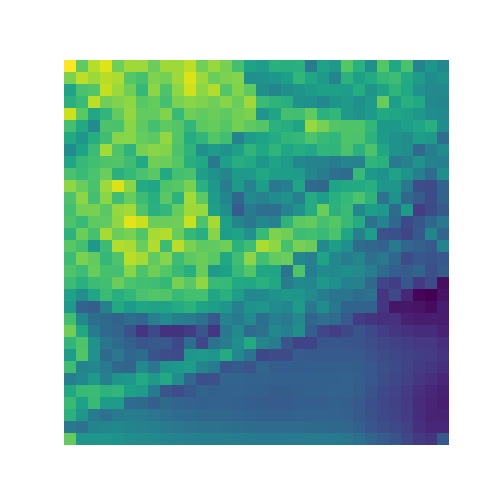}
    \includegraphics[width=0.2\textwidth]{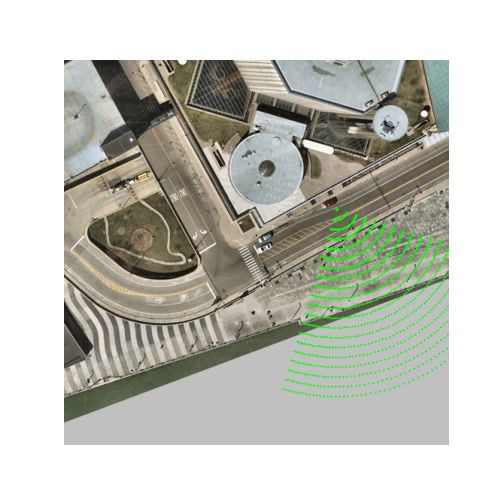}
    \includegraphics[width=0.2\textwidth]{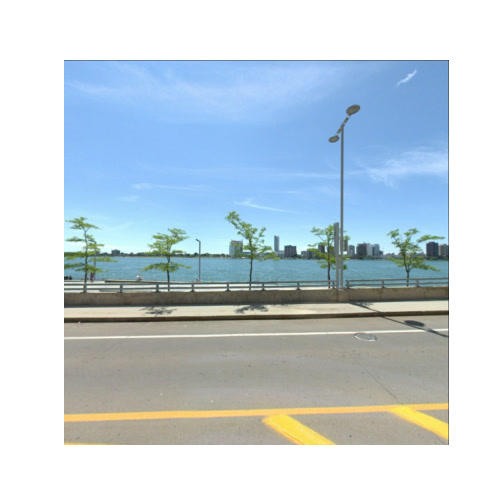}
    \includegraphics[width=0.2\textwidth]{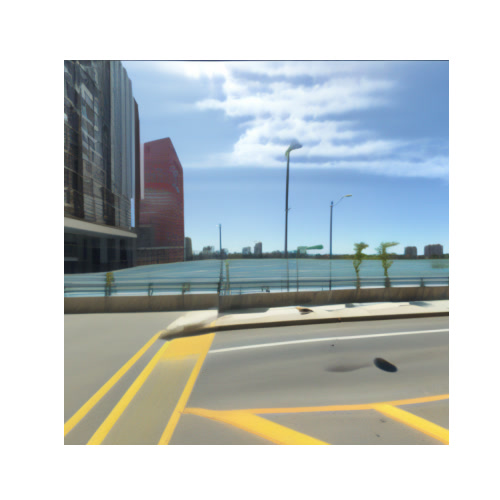}
    \\
    \includegraphics[width=0.99\textwidth]{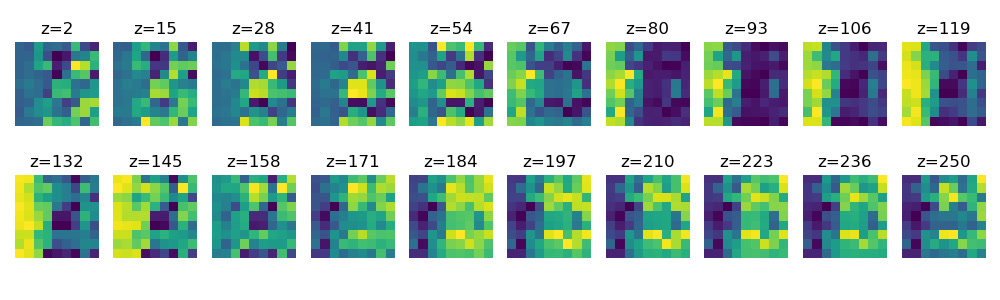}
    \caption{
        Visualization of Satellite attention.
        Top row: input images;
        second row: satellite features, satellite ray samples aligned with target image, target image and generated image;
        third row: attention maps according to the Z value along the ray.
    }
    \vspace{-5pt}
    \label{fig:mapattn}
\end{figure*}

See Figure~\ref{fig:mapattn}
 for visualization of satellite image attention of one of our map conditioning blocks.

\section{Data processing}
\label{sec:supp_training_data}

\subsection{Evaluation datasets}
\label{sec:dataset_clarify}

\noindent \textbf{Metropolis} Each scene contains 7 input views from the front camera on the car, and 14 target views (7 each from the left and right cameras). The front, left, and right cameras each have a $90\degree$ FOV and separated by $90\degree$ in their view directions, leaving little-to-no overlap between input and target views at the same timestep. However, there remains some overlap due to the car's motion over the 7 timesteps. 

\noindent \textbf{nuScenes} We used the front camera on the 6-camera rig as the input view and the remaining 5 cameras as the target views, for 7 timesteps (7 input views, 35 target views total). The images are center-cropped to square, with an FOV of approximately $40\degree$. This setup is similar to that of Metropolis, with very little overlap between target and input views at the same timestep. Any overlap comes from the camera motion across timesteps. 

\subsection{Training datasets}

\begin{table}
\centering
\footnotesize
\resizebox{1.0\columnwidth}{!}{
    {\setlength{\tabcolsep}{3pt}
    \begin{tabular}{lccccc}
        \toprule
        Dataset & MV & Pano &  \#Scenes & Contrib & \shortstack{Satellite \\ source} \\
        \midrule
        Argoverse 2~\cite{Argoverse2} & Yes & No & 700 & 10\% & Google + ArcGIS\\
        nuScenes~\cite{nuscenes} & Yes & No & 700 & 13\% & ArcGIS\\
        Metropolis~\cite{metropolis} & Yes & No & 75 & 7\% &  Original dataset\\
        VIGOR~\cite{zhu2021vigor} & Yes* & Yes & 52105$\dagger$ & 70\% & Azure Maps\\
        \bottomrule
    \end{tabular}}
}
\vspace{-6pt}
\caption{Training datasets.  * We pair nearby panoramas to create a multiview version of the dataset. $\dagger$ Unlike other datasets, the number of scenes in VIGOR corresponds to the number of panoramas. MV = is multi-view? Pano = is panorama? Contrib = training contribution. We train jointly on all datasets.
\vspace{-4pt}
}
\label{tab:training_datasets}
\end{table}

For our training datasets, we augment the existing dataset with
satellite imagery from various sources.
Given GPS latitude and longitude information from the original datasets, we query tiled web map providers such as  Google Maps~\cite{GoogleMaps}, Azure Maps~\cite{AzureMaps}, and Esri World Imagery~\cite{EsriWorldImagery2025}.

The queries are done using the TiledWebMaps package~\cite{tiledwebmaps1,tiledwebmaps2,tiledwebmaps3} at a sampling density of 4 pixels per meter, at a resolution of 128m ${\times}$ 128m.

We will provide the exact GPS coordinates, code, and query parameters used to query satellite images for each scene so that our experiments can be reproduced using any compatible tile provider (Esri, Google, Azure/Microsoft).

\paragraph{VIGOR\cite{zhu2021vigor}} The original VIGOR dataset contains panorama images with non-centered satellite images with large zoom levels. We regenerate the dataset satellite images such that they are centered at the panorama latitude and longitude location.

\paragraph{Metropolis\cite{metropolis}} The Mapillary Metropolis dataset provides high-resolution satellite imagery from which we use a crop that is centered at the centroid of the input and target camera views. 

\paragraph{Argoverse 2\cite{Argoverse2}} We use GPS locations to query Esri World Imagery (ArcGIS) for the satellite images, except Pittsburgh where we use Google Maps imagery because we found that the ArcGIS satellite image quality was poor for this region.

\paragraph{nuScenes\cite{nuscenes}} We use GPS locations to query Esri World Imagery (ArcGIS) for the satellite images.

We found that training our satellite conditioning parameters on relatively small datasets (\eg. Argoverse or VIGOR alone) can cause them to overfit to the camera  trajectory distributions on these datasets. Most of our datasets use a multi-view inside-out facing camera rig with straight-line motion (or no motion in case of panoramas). The model on these alone does not generalize to other scenes with handheld cameras. For this reason, we add data that is closer to the training distribution of SEVA to retain its performance beyond our training dataset. We include scenes from DL3DV \cite{ling2024dl3dv} with the benchmark scenes removed during training. For these scenes, we do not use any satellite encoder features (the conditioning layers get a zero-input). Our training data mix is summarized in Table \ref{tab:training_datasets}. 

\section{Additional Model Details}
\label{sec:supp_model_details}

We provide here more details about our model architecture, in particular - details about the ray-guided cross-attention block that we propose.

\subsection{Ray-guided Cross Attention}

We insert our \textbf{ray-guided satellite cross attention block} after the intra-view self-attention and inter-view self-attention blocks in SEVA. 
The satellite-conditioned cross attention block primarily contains a satellite cross-attention layer and an MLP, both of which are applied in a residual manner \cite{he2016deep} to the diffusion embeddings that are input to the block.

Each such block receives four inputs: 
\begin{itemize}
    \item \textbf{multi-view diffusion embeddings $I$}: 
        These are the intermediate SEVA features at that block. The number of channels and resolution varies throughout the model.
    \item \textbf{Ray embeddings $R$}:
        The Plucker coordinates for the 3D ray direction, which does not include the yaw component \ie is computed using the rotation matrix with the yaw removed. 
        These ray coordinates are passed to an MLP to encode to a fixed dimension.
    \item \textbf{The diffusion timestep embedding $t$}:
        This diffusion timestep $t$ is passed to an MLP to encode to a fixed dimension.
    \item \textbf{Satellite feature samples $S$}:
        20 samples along each street image ray, with 1D position embedding added along each ray.
        These samples are evenly spaced between the camera origin and a maximum depth of 62m.
\end{itemize}
Note that the ray embedding MLP and diffusion timestep MLP are shared across different satellite conditioning blocks. These are added to the UNet's intermediate features to obtain the cross-attention queries. In our ablations, we find that it is important to use these embeddings, as the UNet intermediate features are very noisy in the early diffusion process. Since the satellite image is top-down, all patches along a particular column of the street-image attend to the same ray along the satellite image (assuming images are approximately gravity-aligned). The raymap embeddings in the attention query can therefore, help tokens from the same column attend to different satellite image tokens. 

Given these inputs, the block updates the diffusion embeddings $I$ as follows:

\begin{algorithmic}
    \STATE query $\gets$ MLP( MLP($I$) + $R$ + $t$) \\
    \STATE query $\gets$ layernorm(query) \\
    \STATE key $\gets$ $S$ \\
    \STATE value $\leftarrow$ $S$ \\
    \STATE residual $\leftarrow$ MHCA(query, key, value) \\
    \STATE $I$ $\leftarrow$ $I$ + layernorm(residual) \\
    \STATE $I$ $\leftarrow$ $I$ + MLP(layernorm($I$))
\end{algorithmic}

The output is passed onto the next diffusion model block.

The ray-guided cross-attention is different from a vanilla cross attention in that each image token selectively attends to the satellite tokens along its 3D ray, as opposed to attending to the whole satellite image. This efficient mechanism enables us to condition a large multi-view diffusion on satellite information by training the added satellite-conditioning layers on the relatively small easily accessible paired image-satellite datasets. 

\section{Limitations}

Overall, our approach for improving NVS is only suitable for use in GPS-localized outside locations. We build our satellite-conditioned model on top of an existing multi-view diffusion model (SEVA \cite{zhou2025stable}). SEVA uses an entangled camera pose and appearance representation (by concatenating raymaps and images) which means that any modifications to its embeddings can affect either appearance or poses. When used on datasets with camera trajectories that are completely out-of-distribution, we notice that satellite conditioning can affect the multi-view consistency (pose conditioning) or appearance (saturation or image quality) of the generated images. Our training minimizes this by balancing various different datasets. We also notice that in many cases the satellite conditioning can induce an incorrect generation by being sensitive to shadows of buildings that are visible in the satellite image. Unsurprisingly, our model can also struggle when the images are captured in covered areas, \eg under a bridge, under tree canopies, parking lots, etc. If there have been large changes in the scene since the satellite view was captured, \eg buildings have been built or demolished, then performance may degrade. 

\section{Future work}

Our work can be revisited with emerging DiT-based diffusion models, which are more flexible than UNet-based models. Currently, our model does not use any viewpoint metadata available for satellite images. Incorporating overhead viewpoint information this could allow joint conditioning on satellite and oblique aerial images. Incorporating vectorized semantic map conditioning is also an exciting direction for future work.

\end{document}